\documentclass{article} 

\usepackage{graphicx}
\usepackage{fancyhdr}
\usepackage[final]{dlg-agenticmodeling}

\usepackage{microtype}
\usepackage{hyperref}
\usepackage{url}
\usepackage{booktabs}
\usepackage{amssymb}
\usepackage{algorithm}
\usepackage{algpseudocode}
\usepackage{wrapfig}
\usepackage{pifont}
\usepackage{marvosym}
\usepackage{hyperref}
\usepackage{fontawesome5}
\newcommand{\cmark}{\ding{51}}%
\newcommand{\xmark}{\ding{55}}%

\usepackage[most]{tcolorbox}
\definecolor{SoftBlue}{RGB}{232,242,255} 

\newtcolorbox{abstractbox}{
  colback=SoftBlue,
  colframe=SoftBlue, 
  boxrule=0pt,       
  arc=2mm,           
  left=4mm,right=4mm,top=2mm,bottom=2mm
}

\usepackage{lineno}
\usepackage{multirow}
\usepackage{booktabs}
\usepackage[T1]{fontenc}
\usepackage{xcolor}
\usepackage{tcolorbox}
\usepackage[table]{xcolor}
\tcbuselibrary{listings,skins,breakable}

\renewcommand{\thefootnote}{\fnsymbol{footnote}}
\newcommand\blfootnote[1]{%
  \begingroup
  \renewcommand\thefootnote{}\footnote{#1}%
  \addtocounter{footnote}{-1}%
  \endgroup
}

\definecolor{darkblue}{rgb}{0, 0, 0.5}
\hypersetup{colorlinks=true, citecolor=darkblue, linkcolor=darkblue, urlcolor=darkblue}

\title{Orchard: An Open-Source Agentic Modeling Framework}

\author{%
  \textbf{Baolin Peng}\textsuperscript{$\dagger$1\,\Letter}, 
  \textbf{Wenlin Yao}\textsuperscript{$\dagger$1},
  \textbf{Qianhui Wu}\textsuperscript{$\dagger$1}, 
  \textbf{Hao Cheng}\textsuperscript{$\dagger$1} \\
  \textbf{Xiao Yu}\textsuperscript{$\ddagger$2}, 
  \textbf{Rui Yang}\textsuperscript{$\ddagger$3}, 
  \textbf{Tao Ge}\textsuperscript{1}, 
  \textbf{Alessandro Sordoni}\textsuperscript{1}, 
  \textbf{Xingdi Yuan}\textsuperscript{1}, 
  \textbf{Yelong Shen}\textsuperscript{1} \\
  \textbf{Pengcheng He}\textsuperscript{1}, 
  \textbf{Tong Zhang}\textsuperscript{3}, 
  \textbf{Zhou Yu}\textsuperscript{2}, 
  \textbf{Jianfeng Gao}\textsuperscript{1\,\Letter} \\
  \textsuperscript{1}Microsoft Research \quad
  \textsuperscript{2}Columbia University \quad
  \textsuperscript{3}UIUC \\
  \texttt{\{baolinpeng,jfgao\}@microsoft.com}\\[2mm]
  \faGithub\ \href{https://github.com/microsoft/Orchard}{GitHub}
  \qquad
  \includegraphics[height=0.9em]{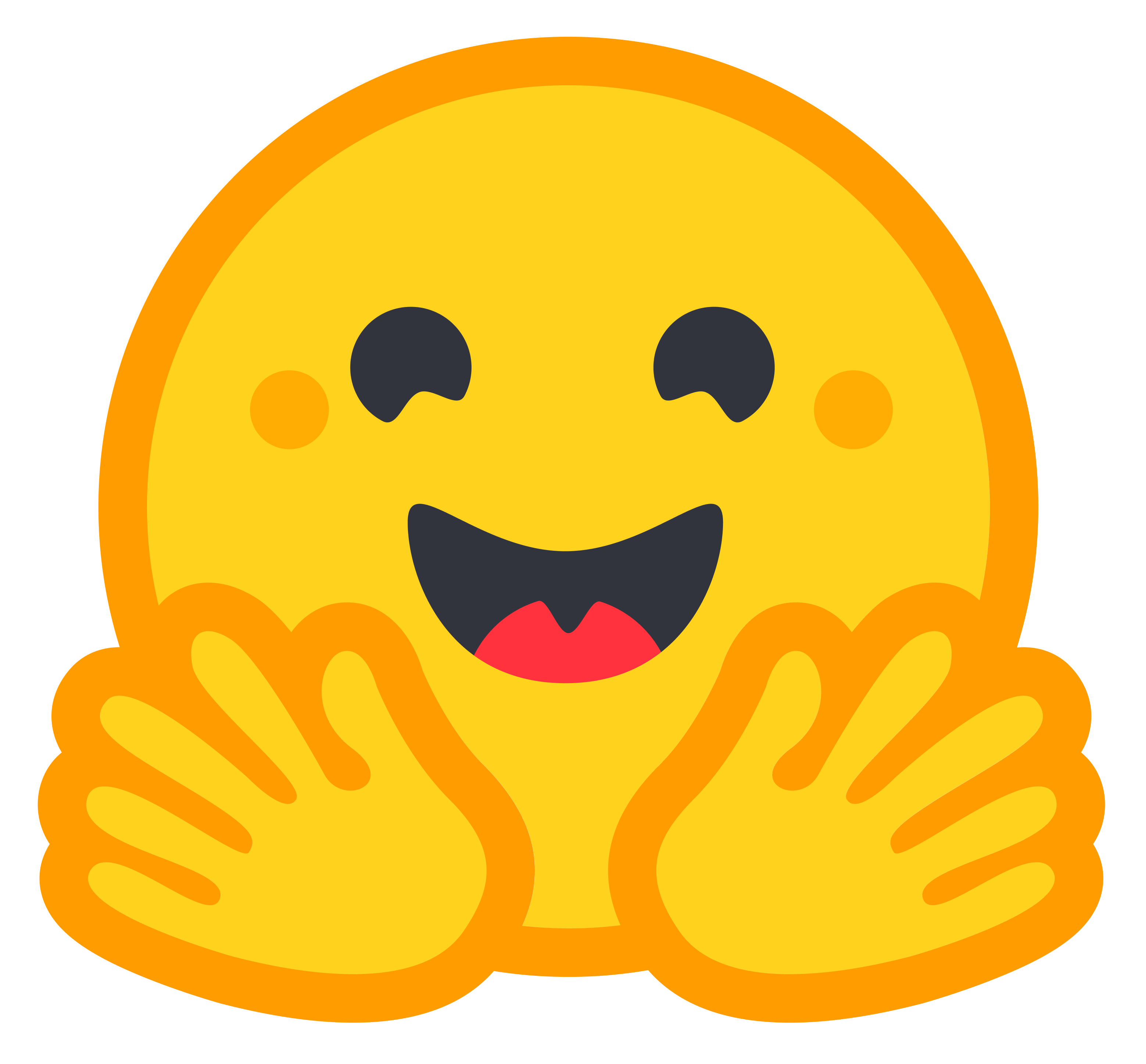}
  \ \href{https://huggingface.co/datasets/microsoft/Orchard}{Hugging Face}
}
\newcommand{\eg}{\textit{e.g.}}
\newcommand{\ie}{\textit{i.e.}}

\newtcblisting{toolfamily}[2][]{%
    enhanced, breakable,
    colback=blue!5!white,
    colframe=blue!55!black,
    colbacktitle=blue!55!black,
    coltitle=white,
    fonttitle=\bfseries,
    title={#2},
    sharp corners=south, arc=3pt,
    boxrule=0.5pt,
    left=8pt, right=8pt, top=4pt, bottom=4pt,
    listing only,
    listing options={
        basicstyle=\ttfamily\footnotesize,
        keywordstyle=\bfseries\color{blue!55!black},
        showstringspaces=false,
        breaklines=true,
        breakatwhitespace=true,
        columns=fullflexible,
        keepspaces=true,
        morekeywords={true,false,null},
    },
    #1
}

\begin{document}

\renewcommand{\thefootnote}{\arabic{footnote}}

\ifcolmsubmission
\linenumbers
\fi

\maketitle
\vspace{-1.0cm} 

\blfootnote{$\dagger$ First authors\ \ $\ddagger$ Second authors}

\begin{abstractbox}
\begin{abstract}
Agentic modeling aims to transform large language models (LLMs) into autonomous agents that can solve complex tasks through planning, reasoning, tool use, and multi-turn interaction with external environments.
We present \textbf{Orchard}, an open-source framework for scalable agentic modeling.
At its core is \textbf{Orchard Env}, a thin, Kubernetes-native environment service that provides reusable primitives for sandbox lifecycle management.
Orchard Env is designed to operate across task domains, agent harnesses, and different training stages.
On top of Orchard Env, we build three agentic modeling recipes.
\textbf{Orchard-SWE} targets software-engineering agents. We 
introduce credit-assignment supervised fine-tuning to learn from productive segments of unresolved trajectories. For RL we go beyond sparse outcome rewards with a progression of signals: Balanced Adaptive Rollout (BAR) for sample-efficient sparse-reward optimization, and two denser rewards—on-policy distillation (OPD) and rubric-based process reward (RPR)—that supervise step-level progress and verification quality. We further introduce \emph{historical experience distillation}, which compresses rollouts from past experiments into a small value model that reranks candidates at inference time. On Qwen3.5-35B-A3B backbone it reaches \textbf{69.7\%} with RPR-based RL and \textbf{73.0\%} with best-of-N value-model reranking
, setting a new state of the art among open-source methods while approaching frontier systems more than $10\times$ larger.
\textbf{Orchard-GUI} trains a 4B vision-language computer-use agent with only 0.4K distilled trajectories and 2.2K open-ended training tasks. It achieves success rates of 74.1\%, 67.0\%, and 64.0\% on WebVoyager, Online-Mind2Web, and DeepShop, respectively (68.4\% average), making it the strongest open-source model while remaining competitive with proprietary systems from OpenAI and Google Gemini.
\textbf{Orchard-Claw} targets personal assistant agents for productivity workflows such as email, calendar, and daily tool-use tasks. Trained with only 0.2K synthetic tasks, it achieves 59.6\% pass@3 on Claw-Eval, and improves to 73.9\% pass@3 when paired with a stronger ZeroClaw harness.
Collectively, these results demonstrate that a thin, open, harness-agnostic environment layer enables the reuse of agentic data, training recipes, and evaluation protocols across domains and harnesses.
\end{abstract}
\end{abstractbox}

\section{Introduction}
\label{sec:intro}

\begin{figure}[t]
    \centering
    \begin{minipage}[c]{0.55\linewidth}
        \centering
        \includegraphics[width=\linewidth,keepaspectratio]{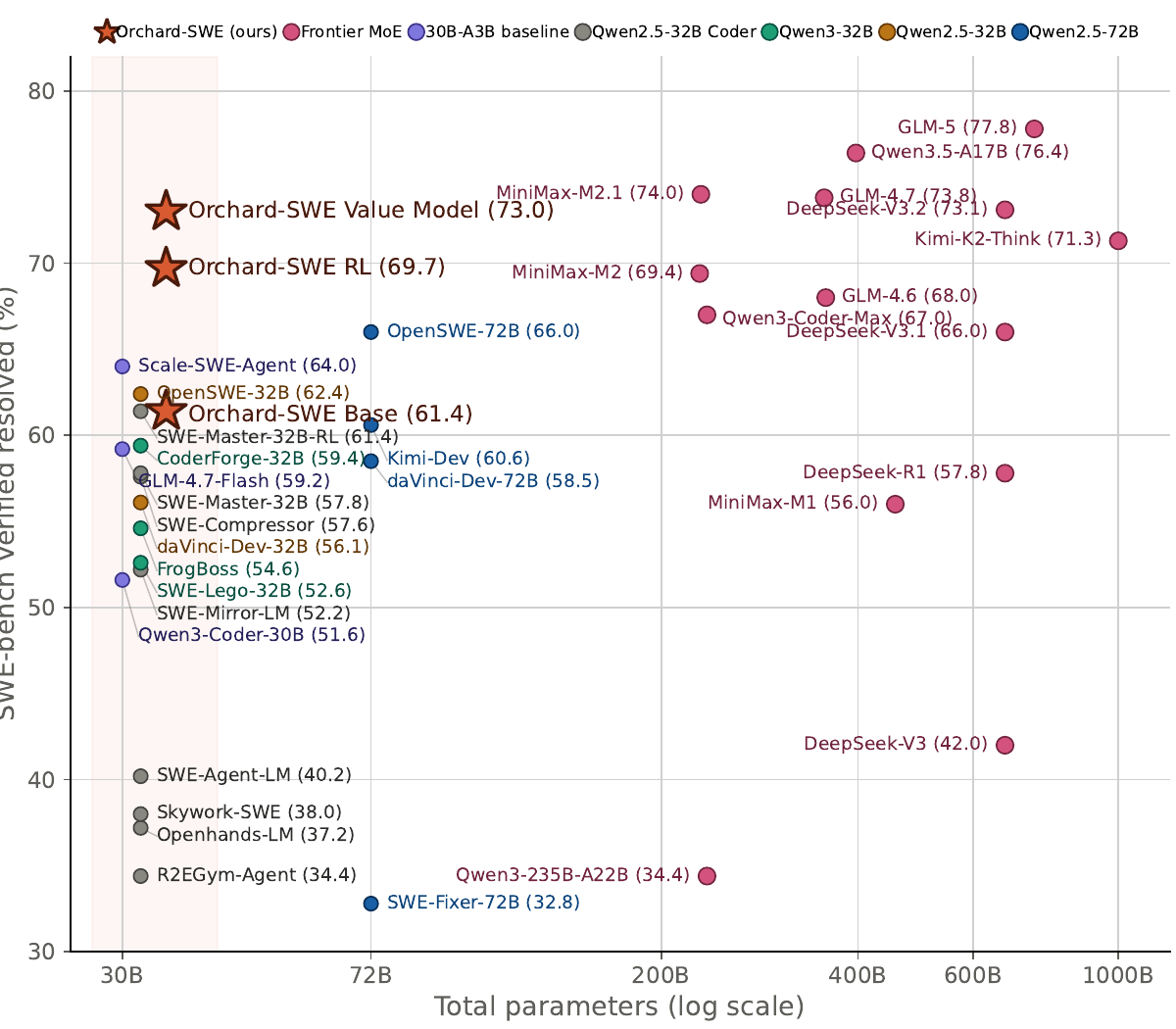}
    \end{minipage}\hfill
    \begin{minipage}[c]{0.44\linewidth}
        \centering
        \includegraphics[width=\linewidth,keepaspectratio]{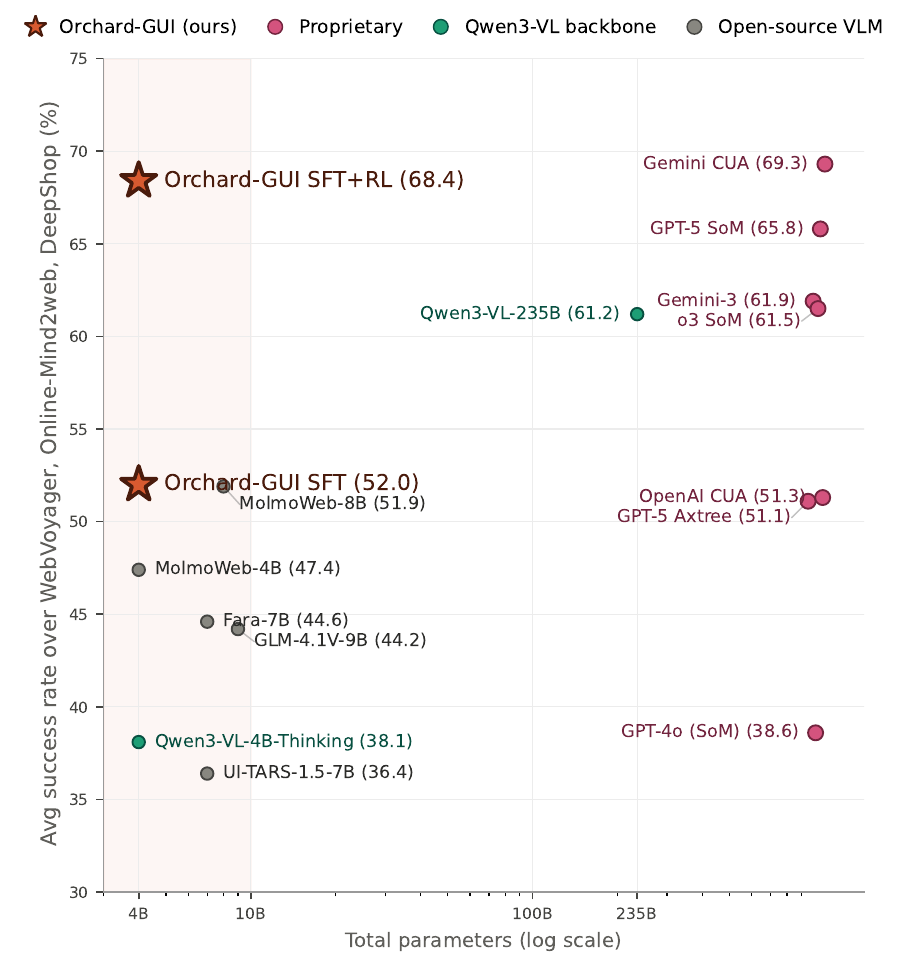}
    \end{minipage}
    \caption{
    \textbf{Performance comparison.}
    \emph{Left:} \textbf{Orchard-SWE} (35B-A3B) reaches 73\% on SWE-bench Verified, approaching frontier MoE systems $10$--$30\times$ larger.
    \emph{Right:} \textbf{Orchard-GUI} (4B) achieves 68.4\% average success across WebVoyager, Online-Mind2web, and DeepShop, making it the strongest open-source GUI agent while staying on par with proprietary systems from OpenAI and Google.
    }
    \label{fig:performance}
\end{figure}

Large language model (LLM) agents that interact with external environments over multiple turns have become a central paradigm for tasks ranging from software engineering~\citep{jimenez2024swebench,yang2024sweagent} and web navigation~\citep{zhou2023webarena,zhang2024large,ning2025survey} to general computer use~\citep{xie2024osworld,hu2025agents}.
Training such agents---through supervised fine-tuning on expert trajectories or reinforcement learning from environment rewards---requires generating large numbers of rollout trajectories, each involving dozens of sequential interactions with a sandboxed execution environment.

As agentic training and evaluation scale to new domains and larger datasets, the need for open, scalable, affordable, and research-friendly infrastructure becomes increasingly acute.
For example, generating a single trajectory for a software engineering task may involve cloning a repository, installing dependencies, applying code edits, and running a test suite---all within an isolated container that must be provisioned, managed, and cleaned up.
At scale, thousands of such environments must run concurrently, each with distinct base images, resource requirements, and network isolation constraints.

We identify the environment layer as the foundational bottleneck. When it is closed or rigidly coupled to a particular training stack, every layer above it---training recipes, evaluation pipelines, trajectory collection---inherits those constraints and cannot be independently reproduced or reused.
Existing systems make different choices about where to place environment management, each with trade-offs.
Managed sandbox platforms such as E2B~\citep{e2b2024}, Daytona~\citep{daytona2024}, and Modal~\citep{modal2025} provide convenient hosted runtimes, but give researchers limited control over infrastructure configuration, cost, and reproducibility.
Vertically integrated training stacks such as ProRL Agent~\citep{prorlagent2026} and MegaFlow~\citep{megaflow2026} include environment management as part of a larger rollout or training system, coupling it with inference scheduling, reward computation, and training-loop orchestration.
Broader environment frameworks such as ROCK~\citep{rock2026} provides rich platform functionality, but do not isolate the environment layer as a minimal service boundary.
As a result, trajectory datasets, training recipes, and evaluation pipelines are often tied to a particular harness or infrastructure implementation, making them difficult to reproduce, compare, or reuse.

We argue that the environment layer should instead be a \emph{thin, standalone service} reusable along three axes: across \emph{(i)}~task domains, \emph{(ii)}~agent harnesses within a domain, and \emph{(iii)}~pipeline stages, including trajectory distillation, on-policy RL rollouts, and evaluation.
When this boundary is clean, the layers above it become reusable as well: data can be collected under one harness and evaluated under another, SFT and RL recipes can share the same execution backend, and new domains can reuse the same infrastructure rather than rebuild it.

\begin{figure}[t]
    \centering
    \includegraphics[width=\linewidth]{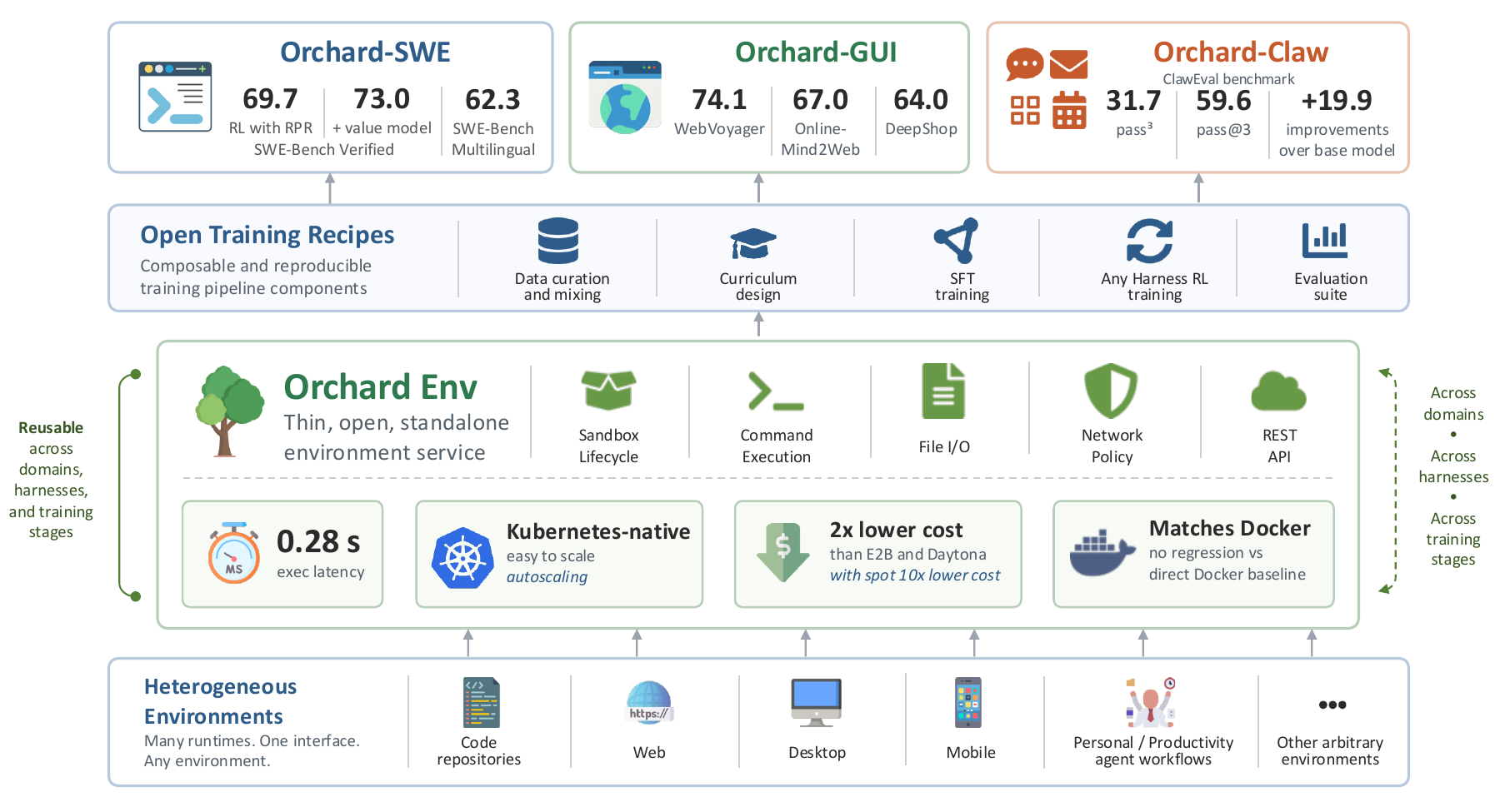}
    \caption{Overview of the Orchard framework. Orchard Env (center) is a thin, Kubernetes-native environment service that exposes generic primitives---sandbox lifecycle, command execution, file I/O, network policy, a REST API, and lightweight agent injection---and supports heterogeneous task environments (bottom row). Open training recipes (second row) compose with this service without coupling to it, and we instantiate the same stack in three domains (top row): \textbf{Orchard-SWE} (software engineering), \textbf{Orchard-GUI} (browser navigation), and \textbf{Orchard-Claw} (AI personal assistant); per-domain headline numbers are summarized inside each domain box.}
    \label{fig:teaser}
\end{figure}

Therefore, we present \textbf{Orchard} (Figure~\ref{fig:teaser}), an open framework for scalable agentic modeling centered on a thin, reusable environment layer. Its core component, \textbf{Orchard Env}, is a Kubernetes-native service that exposes generic primitives—sandbox lifecycle management, command execution, file I/O, network policy, and a REST API—without coupling to any agent harness, trainer, inference backend, or task domain. Orchard Env scales through two key choices: runtime agent injection, which allows arbitrary task-specific Docker images to run separately, and direct routing of execution and file requests to sandbox Pod IPs, avoiding Kubernetes exec/WebSocket overhead. Together with network isolation, asynchronous lifecycle management, heartbeat cleanup, and watch-based readiness tracking, these mechanisms make Orchard Env broadly composable and practical for large-scale environment interaction. Empirically, it achieves 0.28s average command-execution latency, sustains a 1{,}000-sandbox stress test with 100\% success, and substantially lowers estimated sandboxing cost relative to other alternatives.

On top of Orchard Env, we develop three agentic modeling (SFT+RL) recipes that compose with the environment service without tight coupling.
These recipes handle trajectory collection, data curation, reward computation, and policy optimization. We instantiate them with Qwen backbone models for browser agents, software engineering agents and personal assistant agents.
Across three domains, the same environment abstraction supports diverse modalities, tool interfaces, agent harnesses, and reward mechanisms.

\paragraph{Orchard-SWE.}
For software engineering, Orchard-SWE targets two key bottlenecks of open SWE-agent training:
limited supervision and sparse rewards. On the supervision side, we curate 107K trajectories
distilled from MiniMax-M2.5 and Qwen3.5-397B across SWE-rebench~\citep{badertdinov2025swe} and Scale-SWE~\citep{zhao2026immersion}, using both
the OpenHands~\citep{wang2025openhands} and mini-swe-agent~\citep{yang2024sweagent} harnesses;
and—unlike most prior recipes that keep only resolved trajectories—we introduce
\emph{credit-assignment SFT}, which uses retrospective value estimation to extract productive
rise segments from failed trajectories and convert partial progress into supervision. On the
reward side, we develop a progression of increasingly dense RL signals: \emph{Balanced Adaptive
Rollout (BAR)} makes sparse-reward RL sample-efficient by adaptively assembling reward-balanced
trajectory groups; \emph{on-policy distillation (OPD)} supplies a token-level teacher signal
that matches BAR at lower rollout cost; and a \emph{rubric-based process reward (RPR)} scores
verification quality along the whole trajectory, densifying the signal where outcome reward is
silent. Finally, we introduce \emph{historical experience distillation}, which compresses
rollouts accumulated across past RL runs into a small value model that reranks candidates at
inference time. On Qwen3-30B-A3B-Thinking, Orchard-SWE reaches 64.3\% after SFT and 67.5\% after
SFT+RL under mini-swe-agent; on the stronger Qwen3.5-35B-A3B backbone, RPR-based RL reaches
\textbf{69.7\%} and value-model reranking reaches \textbf{73.0\%}—a new state of the art among
open-source models of comparable size, approaching frontier systems more than $10\times$ larger.

\paragraph{Orchard-GUI.}
For browser-based GUI agents, Orchard-GUI shows that the same environment service and recipe transfer beyond text-only computer use tasks.
We train a 4B vision-language backbone with a generic ReAct-style~\citep{yao2023react} browser harness and evaluate on WebVoyager~\citep{he2024webvoyager}, Online-Mind2Web~\citep{deng2023mind2web}, and DeepShop~\citep{lyu2025deepshop}.
After SFT+RL training, Orchard-GUI achieves success rates of 74.1\%, 67.0\%, and 64.0\% on the three benchmarks, averaging 68.4\% overall with the largest gains observed on long-horizon benchmarks, \ie, Online-Mind2Web and DeepShop.
This is a new open-source state of the art while remaining competitive with leading proprietary computer-use systems, despite using a 4B backbone model and only 2.6K training tasks.
Remarkably, Orchard-GUI substantially outperforms both prior open-source agents and its 235B teacher model, 
suggesting that environment-grounded RL can improve model's agentic capabilities beyond those of the teacher.

\paragraph{Orchard-Claw.}
For personal assistant agent, Orchard-Claw studies whether machine learned agent skills can transfer across different harnesses.
We synthesize training tasks from Claw-Eval~\citep{ye2026clawevaltrustworthyevaluationautonomous} seeds and ClawHub~\citep{clawhub2026} workflows, distill successful MiniMax-M2.5 trajectories,  perform agentic training (SFT+RL) on Qwen3-30B-A3B-Thinking, and evaluate across harnesses, including a ReAct-style harness and the ZeroClaw~\citep{zeroclaw2026} harness.
Orchard-Claw achieves 31.7\% $pass^3$ and 59.6\% $pass@3$ on Claw-Eval, significantly outperforming comparable-size open-source baselines despite using only 0.2K synthetic tasks.
When paired with the stronger ZeroClaw harness at inference time, the same model improves further to 41.0\% $pass^3$ and 73.9\% $pass@3$.

Collectively, the results from the three agentic modeling recipes support the central claim of this study: the environment layer is not merely an infrastructural component, but the substrate governing the reusability of agentic modeling artifacts.
A thin, open, harness-agnostic environment service enables trajectory data, SFT recipes, RL rollouts, and evaluation protocols to transfer across domains, agent harnesses, and pipeline stages.
Orchard demonstrates that open-source agentic modeling can be scaled in a manner that is both cost-effective and reproducible, without coupling the environment to any single training stack.
We release the full Orchard framework---environment service, training recipes, and trajectory datasets spanning software engineering, GUI navigation, and personal-assistant tool use---to facilitate open research in scalable agentic modeling.
\section{Orchard Env}
\label{sec:orchard}


Scaling agentic training across domains and tasks places specific demands on the environment layer.
We identify three core requirements for an environment service that can serve as a practical foundation for the research community:

\begin{enumerate}
    \item \textbf{Thin, standalone service boundary.} Environment management should be isolated as a narrow service---decoupled from agent harness, model serving, and training orchestration---so that any combination of trainer, agent design, and task domain can compose with the same service.
    \item \textbf{Low-cost image compatibility.} The service should support heterogeneous task environments and arbitrary Docker images at low adaptation cost.
    \item \textbf{Accessible and cost-practical at scale.} The service should be deployable on any standard cloud infrastructure, making large-scale agentic training affordable and easy to adopt.
\end{enumerate}

This section describes how Orchard Env realizes these requirements, presents its architecture and key design choices, and positions it among existing systems. More details can be found in Appendix \ref{sec:appendix:env_details}.

\subsection{Architecture Overview}
\label{sec:orchard:arch}

\begin{figure}[t]
    \centering
    \includegraphics[width=\linewidth]{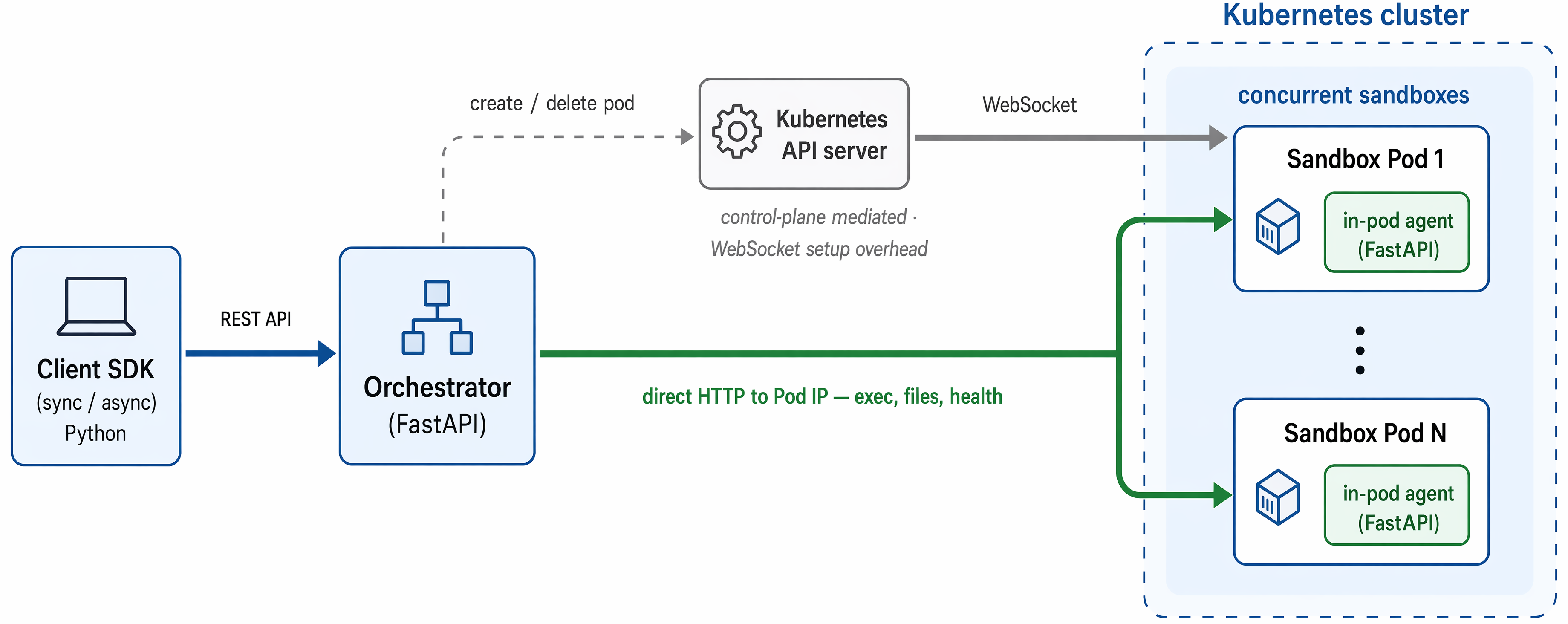}
    \caption{Orchard Env architecture. A Python client SDK (synchronous or asynchronous) issues REST calls to a FastAPI orchestrator, which manages sandbox lifecycle in a Kubernetes cluster. Pod creation and deletion (cold path) go through the Kubernetes API server, while all execution, file, and health requests (hot path) are dispatched directly to each sandbox pod's in-pod agent via Pod IP, bypassing the API server and avoiding \texttt{kubectl exec}/WebSocket setup overhead.}
    \label{fig:orchard-arch}
\end{figure}

Orchard Env follows a three-layer architecture, as illustrated in Figure~\ref{fig:orchard-arch}: a \emph{client SDK} that provides synchronous and asynchronous Python interfaces, an \emph{orchestrator} that manages sandbox lifecycle and scheduling, and a lightweight \emph{in-pod agent} injected into each sandbox container.

This three-layer separation reflects three deliberate choices.
First, the orchestrator and the in-pod agents are deployed and scaled independently: lifecycle decisions (creation, deletion, readiness) flow through the central orchestrator, while per-command execution traffic is dispatched directly to each sandbox's in-pod agent, isolating control-plane operations from the latency-sensitive hot path.
Second, the in-pod agent is \emph{injected} into user-supplied images at runtime rather than baked in at build time, so that arbitrary task images integrate with no per-image modifications.
Third, the entire stack runs on standard Kubernetes primitives (Pods, NetworkPolicy, Watch), inheriting open ecosystem tooling, multi-cloud portability, and cost optimizations such as cluster autoscaling and spot instances.
We describe each layer in turn.

\paragraph{Client SDK.}
Orchard Env provides both synchronous (\texttt{SandboxClient}) and asynchronous (\texttt{AsyncSandboxClient}) Python clients.
Sandboxes are created from user-specified Docker images and expose methods for command execution, file upload/download, and patch application.
Context managers provide automatic cleanup, and the SDK exposes heartbeat utilities for keeping long-lived sandboxes alive when desired.
The SDK also includes configurable retry logic with exponential backoff for transient connection errors and service unavailable errors.

\paragraph{Orchestrator.}

The orchestrator is a FastAPI service deployed as a Kubernetes Deployment with multiple replicas.
It exposes a REST API for sandbox lifecycle management and can delegate sandbox metadata tracking to an optional Redis backend across replicas.
Key responsibilities include:
\textbf{Sandbox provisioning:} Translating \texttt{POST /sandboxes} requests into Kubernetes Pod specifications, including init container configuration, resource limits, network policies, and readiness probes.
\textbf{Readiness tracking:} A \texttt{PodWatcher} component maintains a persistent Kubernetes \texttt{LIST+WATCH} stream, caching pod state transitions and waking blocked clients
when pods become ready.
\textbf{Execution scheduling:} An \texttt{ExecManager} routes execution requests to the target sandbox's in-pod agent via direct HTTP calls to the Pod IP, serializing concurrent requests to the same sandbox via per-sandbox locks.
\textbf{Lifecycle management:} A background reconciliation loop detects and cleans up orphaned sandboxes (those whose heartbeat has expired or whose backing Pod has been evicted).

\paragraph{In-Pod Agent.}
The in-pod agent\footnote{Here, ``agent'' refers to the sandbox-side execution service, not the LLM-based agents studied elsewhere in this paper.} is a lightweight FastAPI server that runs inside each sandbox container.
It exposes endpoints for command execution (\texttt{/exec}), file upload, download, listing, and health checking.
Commands are executed as subprocesses with configurable timeouts; on timeout, the entire process tree is killed via process group signal.
The agent is reachable only through the sandbox pod's internal cluster network endpoint, and its health endpoint serves as the Kubernetes readiness probe.

\subsection{Comparison with Existing Systems}
\label{sec:orchard:comparison}

To position Orchard Env relative to existing systems, Table~\ref{tab:env-comparison} compares environment and training infrastructure along four dimensions derived from the requirements above: whether an open-source server stack exists that researchers can self-host, whether the system is operated primarily as a managed service, whether it exposes a thin standalone environment service, and its relative cost at research scale.
Concretely, we treat a system as a \emph{thin env service} when \emph{(i)}~environment management is the system's primary scope rather than a by-product of agent harness, training orchestration, or LLM serving; \emph{(ii)}~the environment layer presents a stable API---typically a small REST surface for sandbox lifecycle and command execution---that does not require the caller to adopt the system's trainer, scheduler, or rollout abstractions; and \emph{(iii)}~that API is independent of the choice of agent harness, RL trainer, and inference backend, so the same service can back distillation, RL rollouts, and evaluation interchangeably. We highlight three aspects of Orchard Env's positioning\footnote{The comparison is based on public documentation and repositories as of April~2026.}:

\begin{table*}[t]
\centering
\caption{Environment and training systems for agentic training, based on public documentation as of April~2026.
\emph{Scope}: the system's primary design scope.
\emph{Self-host.}: an open-source server/control-plane stack exists that researchers can deploy on their own infrastructure.
\emph{Mgd.\ default}: the system's primary product is offered as a managed/hosted service.
\emph{Thin env service}: environment management is exposed as a narrow, standalone service boundary independent of agent harness, training loop, and inference backend (operational definition above).
\emph{Rel.\ cost}: normalized to Daytona for a 2-vCPU, 8-GiB sandbox; ``---'' indicates no publicly comparable pricing.
See Appendix~\ref{sec:appendix:cost} for methodology.}
\label{tab:env-comparison}
\footnotesize
\setlength{\tabcolsep}{4pt}
\begin{tabular}{@{}llcccr@{}}
\toprule
\textbf{System} & \textbf{Scope} & \textbf{Self-host.} & \textbf{Mgd.\ default} & \textbf{Thin env service} & \textbf{Rel.\ cost} \\
\midrule
E2B           & managed sandbox     & \cmark$^\dagger$ & \cmark & \cmark & 1.0$\times$ \\
Daytona       & managed sandbox     & \cmark$^\dagger$ & \cmark & \cmark & 1.0$\times$ \\
Modal         & compute platform    & \xmark & \cmark & \xmark & 1.5$\times$ \\
SkyPilot      & compute orch.       & \cmark & \xmark & \xmark & --- \\
ROCK          & env.\ framework     & \cmark & \xmark & \xmark & --- \\
ProRL Agent   & rollout infra.      & \cmark & \xmark & \xmark & --- \\
MegaFlow      & training orch.      & \xmark & \xmark & \xmark & --- \\
\midrule
\textbf{Orchard Env} & \textbf{env.\ service} & \cmark & \xmark & \cmark & \textbf{0.47$\times$}$^*$ \\
\bottomrule
\end{tabular}
\\[2pt]
\hfill{\footnotesize $^*$0.10$\times$ with spot instances. $^\dagger$E2B and Daytona ship limited open-source server components, but their primary product is the hosted control plane and the \emph{Rel.\ cost} column reflects that managed offering.\hspace{2em}}
\end{table*}

\paragraph{Thin env service vs.\ integrated and broad systems.}
ProRL Agent~\citep{prorlagent2026} achieves an important decoupling---separating the rollout lifecycle from the RL trainer via an HTTP service---but its environment layer remains coupled with agent harness (via \texttt{AgentHandler} plugins), LLM inference routing, and evaluation logic within the same rollout server.
MegaFlow~\citep{megaflow2026} similarly embeds environment management within a larger training orchestration system.
Modal~\citep{modal2025} is a different category altogether: it is a general serverless compute platform that offers flexible function and container execution, but it is not specialized as a thin environment service for agentic training, and its hosted control plane and per-second pricing are difficult to amortize across long-running RL training campaigns.
ROCK~\citep{rock2026} provides a broader environment framework with multiple protocols and richer platform components, targeting a wider scope than a thin service boundary.
SkyPilot~\citep{kim2025skypilotsandbox} provides open-source multi-cloud compute orchestration and can serve as the underlying infrastructure on which Orchard Env is deployed; the two are complementary rather than competing.
E2B~\citep{e2b2024} and Daytona~\citep{daytona2024}, like Orchard Env, expose environment management as standalone sandbox services, but as managed products with hosted control planes and vendor-determined pricing.
Orchard Env's distinguishing technical choice is \emph{agent injection}: a Kubernetes init container copies a self-contained execution agent into any user-provided Docker image at pod startup, avoiding the need to rebuild task images.
This enables Orchard Env to support hundreds of heterogeneous task environments---such as the diverse images required by SWE-bench---without per-image modifications.

\paragraph{Deployment portability.}
Orchard Env targets researcher-controlled infrastructure: any standard Kubernetes environment---managed (AKS, EKS, GKE) or self-hosted---can run the full stack, with direct control over resource allocation, network policies, and autoscaling.
This contrasts with HPC-oriented systems like ProRL Agent, which require access to institutional Slurm clusters and Singularity runtimes, limiting adoption to researchers at specific institutions.

\paragraph{Cost as a consequence of design choices.}
Table~\ref{tab:cost-comparison} compares estimated costs for 128 parallel sandboxes (2~vCPU, 8~GiB each) over 240 hours---a representative RL training workload.
Because Orchard Env is self-hosted on standard Kubernetes, it naturally benefits from cloud-native cost optimization: ephemeral sandbox nodes can run on spot instances, and cluster autoscaling adjusts capacity to actual demand.
This reduces cost to \$673 with spot instances---10$\times$ lower than managed alternatives like Daytona and E2B.
Even at on-demand rates, Orchard Env (\$3,362) is less than half the cost of Daytona (\$7,078) and E2B (\$7,078).
A detailed breakdown is provided in the Appendix~\ref{sec:appendix:cost}.

\begin{table}[t]
\centering
\caption{Estimated cost for 128 parallel sandboxes over 240 hours (30,720 sandbox-hours). Target: 2~vCPU, 8~GiB RAM per sandbox. Costs normalized to Daytona. Prices from official rate cards as of April~2026; see Appendix~\ref{sec:appendix:cost}.}
\label{tab:cost-comparison}
\small
\begin{tabular}{@{}lrrr@{}}
\toprule
\textbf{System} & \textbf{\$/sandbox-hr} & \textbf{Cost (\$)} & \textbf{vs.\ Daytona} \\
\midrule
Daytona                         & 0.230 &   7,078 & --- \\
E2B                             & 0.230 &   7,078 & 0\% \\
Modal                           & 0.335 &  10,305 & +46\% \\
MegaFlow$^\dagger$              & 0.150 (est.) &  4,608 (est.) & $-$35\% \\
Orchard Env (on-demand)         & 0.109 &   3,362 & $-$53\% \\
\textbf{Orchard Env (spot)}     & \textbf{0.022} & \textbf{673} & \textbf{$-$90\%} \\
\bottomrule
\end{tabular}
\\[2pt]
{\footnotesize $^\dagger$MegaFlow is not publicly priced; cells marked \emph{(est.)} are estimated from reported infrastructure usage in \citet{megaflow2026}.}
\end{table}

\subsection{System Evaluation}
\label{sec:orchard:eval}

For both agentic data generation and RL training, the most critical systems metric is environment interaction latency---it directly determines rollout throughput and GPU utilization.
We evaluate Orchard Env on three axes: \emph{(i)}~execution latency relative to existing services, \emph{(ii)}~reliability under high concurrency, and \emph{(iii)}~functional equivalence to a direct Docker baseline in downstream agent evaluations.
Unless noted otherwise, all measurements use a Kubernetes cluster of 8 nodes (each 32~vCPU, 128~GiB RAM) on commodity cloud VMs, with sandbox images pre-pulled on every node and each sandbox provisioned with 2~vCPU and 8~GiB RAM.

\paragraph{Execution latency.}
We compare average command execution latency across four environment services using the benchmark methodology of SkyPilot Code Sandbox~\citep{kim2025skypilotsandbox}, with the same benchmark setup across all platforms.

\begin{table*}[t]
\centering
\caption{System evaluation of Orchard Env. \textbf{Left:} Average command execution latency across environment services (lower is better; benchmark methodology follows~\citet{kim2025skypilotsandbox}). \textbf{Right:} Agent pass rates (\%) on Terminal-Bench 2.0 using Orchard Env vs.\ a direct Docker baseline, averaged over 3 runs per cell, confirming no regression from the environment service layer.}
\label{tab:sys-eval}
\small
\begin{minipage}[t]{0.45\textwidth}
\centering
\begin{tabular}{@{}lr@{}}
\toprule
\textbf{System} & \textbf{Avg.\ latency (s)} \\
\midrule
\textbf{Orchard Env}      & \textbf{0.280} \\
SkyPilot Code Sandbox     & 0.284 \\
E2B                       & 0.747 \\
Modal                     & 2.046 \\
\bottomrule
\end{tabular}
\end{minipage}
\hfill
\begin{minipage}[t]{0.45\textwidth}
\centering
\begin{tabular}{@{}lcc@{}}
\toprule
\textbf{Model} & \textbf{Docker} & \textbf{Orchard Env} \\
\midrule
GPT-4.1           & 34.1 & \textbf{35.1} \\
MiniMax-M2.5      & 52.6 & \textbf{54.4} \\
Qwen3-8B-thinking &  7.0 & \textbf{8.8} \\
\bottomrule
\end{tabular}
\end{minipage}
\end{table*}

As shown in Table~\ref{tab:sys-eval}, Orchard Env achieves an average execution latency of 0.28\,s, essentially matching SkyPilot Code Sandbox (0.284\,s) and significantly outperforming E2B (0.747\,s, 2.7$\times$ slower) and Modal (2.046\,s, 7.3$\times$ slower).
This validates Orchard Env's direct Pod-IP communication design: by routing execution requests directly to the in-pod agent and bypassing the Kubernetes API server on the hot path, Orchard Env achieves latency comparable to optimized native runtimes while retaining the flexibility of a Kubernetes-based deployment.

\paragraph{Reliability under concurrency.}
To stress-test Orchard Env at scale, we ran 1{,}000 sandboxes in parallel through the full lifecycle: create $\rightarrow$ execute 4 commands $\rightarrow$ delete.

\begin{wraptable}{r}{0.42\linewidth}
\vspace{-1em}
\centering
\caption{Stress test results for 1{,}000 parallel sandboxes through the full lifecycle
(create $\rightarrow$ 4$\times$ exec $\rightarrow$ delete).}
\label{tab:concurrency}
\small
\begin{tabular}{@{}lr@{}}
\toprule
\textbf{Metric} & \textbf{Value} \\
\midrule
Parallel sandboxes   & 1{,}000 \\
Commands per sandbox & 4 \\
Success rate         & \textbf{100\%} \\
End-to-end time      & 26\,s \\
Avg.\ create time    & 11.75\,s \\
Avg.\ exec latency   & 0.28\,s \\
\bottomrule
\end{tabular}
\vspace{-1em}
\end{wraptable}

As shown in Table~\ref{tab:concurrency}, Orchard Env achieved a \textbf{100\%} success rate across all 1{,}000 sessions---no failures on creation, execution, or cleanup---with the entire test completing in 26 seconds end-to-end.
Translating these end-to-end numbers, Orchard Env sustains roughly 154 commands per second across the full create~$\rightarrow$~exec~$\rightarrow$~delete lifecycle (4{,}000 commands across 1{,}000 sandboxes in 26\,s), well above the throughput required by typical agentic distillation and RL workloads at this concurrency.
These results confirm that Orchard Env's architecture---watch-based readiness tracking, per-sandbox locking, and heartbeat-based cleanup---remains reliable at the concurrency levels required for large-scale agentic training.

\paragraph{Functional equivalence to Docker.}
Beyond infrastructure metrics, we verify that Orchard Env introduces no performance regression in downstream agent evaluations.
We compare Orchard Env against a direct Docker baseline on Terminal-Bench 2.0~\citep{tbench_2025}, using three models of varying capability; reported numbers are averaged over 3 independent runs per (model, backend) pair.
As shown in Table~\ref{tab:sys-eval} (right), Orchard Env matches Docker within run-to-run variance across all three models, with a marginal edge in every case (1--2 points).
This confirms that the agent-injection mechanism and Orchard Env's execution path introduce no observable overhead or interference in agent--environment interactions.

\section{Orchard-SWE}
\label{sec:swe}

This section presents Orchard-SWE, our instantiation of the Orchard training recipe for software engineering.
We describe the problem setting, trajectory collection pipeline, two-stage training recipe, main results on SWE-bench Verified, and ablations that isolate key design choices.
Our training recipe follows a two-stage pipeline: Supervised Fine-Tuning (SFT) on teacher-distilled trajectories, followed by Reinforcement Learning (RL) with environment-grounded rewards.
Both stages use Orchard Env as the execution backend.

\subsection{Problem Setting}
\label{sec:swe:setup}

\paragraph{Task and evaluation.}
We target the SWE-bench task formulation~\citep{jimenez2024swebench}: given a GitHub issue description and a snapshot of the repository at the time the issue was filed, the agent must produce a code patch that resolves the issue.
A solution is scored as correct if and only if it passes the full gold test suite associated with the ground-truth pull request.
We use \textbf{SWE-bench Verified}~\citep{openai2024swebenchverified}---a human-validated subset of 500 instances---as our primary evaluation benchmark. We also report auxiliary evaluations on \textbf{SWE-bench Multilingual}~\citep{yang2025swesmith} and \textbf{Terminal-Bench 2.0}~\citep{tbench_2025}.

\paragraph{Agent harness and tool interface.}
The agent operates in a multi-turn ReAct-style loop~\citep{yao2023react}: at each step, it produces a reasoning trace (Thought) and a tool invocation (Action), then observes the environment response before proceeding.
The tool interface includes shell command execution, file viewing and editing, and patch submission.
All environment interactions are routed through the Orchard Env service: each task instance runs in an isolated sandbox (2~vCPU, 8~GiB memory) provisioned from the task-specific Docker image, with Orchard Env's agent-injection mechanism handling image heterogeneity transparently.
A distinctive aspect of Orchard-SWE is that we collect trajectories using \textbf{two different agent harnesses}---the full-featured \emph{OpenHands}~\citep{wang2025openhands} framework and a lightweight \emph{mini-swe-agent}~\citep{yang2024sweagent}---and study how harness design affects both trajectory characteristics and downstream training outcomes (Section~\ref{sec:swe:ablations}).

\subsection{Trajectory Collection and Curation}
\label{sec:swe:data}

We construct the Orchard-SWE dataset through large-scale trajectory distillation from strong teacher models, followed by systematic filtering and curation.
Table~\ref{tab:swe-data} summarizes the composition of the final dataset.

\paragraph{Task sources.}
We draw task instances from three sources:
(1)~\textbf{SWE-rebench}~\citep{badertdinov2025swe}, a large-scale collection of real-world GitHub issues with executable Docker-based test environments.
We use its \emph{filtered} subset, which applies quality and difficulty filters to retain instances that are both solvable and non-trivial, covering over 1{,}400 Python repositories.
(2)~\textbf{SWE-rebench V2}~\citep{badertdinov2026swe}, a language-agnostic extension of SWE-rebench that harvests more software engineering tasks. It provides over 32k containerized executable tasks spanning 20 programming languages\footnote{In our experiments, we primarily use its Python tasks for consistency with the rest of our task pool.} and more than 3.6k repositories, together with pre-built images. We reserve SWE-rebench V2 entirely for RL training.
(3)~\textbf{Scale-SWE}~\citep{zhao2026immersion}, a complementary task source that constructs 100k task instances from real GitHub pull requests across 5.2k repositories.
Each instance is packaged with a Docker image, a gold patch, and automatically generated test scripts, significantly expanding the diversity of repositories and issue types available for trajectory collection.

\paragraph{Multi-teacher trajectory generation.}
We use multiple teacher models to increase the diversity of successful trajectories while keeping the downstream action space fixed.
For each task instance, we sample five rollout trajectories through Orchard Env and retain all trajectories that successfully resolve the task.
Our teacher pool includes \textbf{Qwen3.5-397B}~\citep{qwen2026qwen35} and \textbf{MiniMax-M2.5 230B}~\citep{minimax2026m25}.
On SWE-rebench, we collect trajectories from both teachers under the mini-swe-agent and OpenHands harnesses.
Empirically, MiniMax-M2.5 achieves a higher task pass rate, while Qwen3.5-397B occasionally emits tool calls that are not defined in the OpenHands tool interface.
Based on these observations, we use MiniMax-M2.5 as the sole teacher for Scale-SWE, where rollout efficiency and stability become more important due to the larger number of instances.
In all cases, teachers interact through the same sandboxed tool interface used at evaluation time, ensuring that collected trajectories remain faithful to the downstream action space.

\paragraph{Harness selection.}
We collect trajectories with two agent harnesses: \textbf{OpenHands}~\citep{wang2025openhands} and \textbf{mini-swe-agent}~\citep{yang2024sweagent}.
On SWE-rebench, we use both harnesses so that trajectory collection covers a broader range of interaction styles and tool-use patterns.
For the OpenHands runs, we follow its standard SWE-bench tool configuration.\footnote{\url{https://github.com/OpenHands/benchmarks/tree/main/benchmarks/swebench}}
For Scale-SWE, we use only mini-swe-agent, a lightweight harness with a minimal tool set (bash execution, file editing, submission), since we did not observe a meaningful performance gap relative to OpenHands on this source and the lighter harness is more practical for large-scale rollout collection.
This dual-harness setup also lets us analyze how harness choice affects downstream training outcomes (Section~\ref{sec:swe:ablations}).

\paragraph{Filtering and curation.}
Unlike most prior work, which retains only successful (resolved) trajectories for SFT, we keep \textbf{both resolved and unresolved trajectories} in the training corpus.
Resolved trajectories provide standard imitation signal; unresolved trajectories are curated via temporal-difference credit assignment to extract continuous \emph{rise segments}---spans where the trajectory is making progress---which contribute partial-progress supervision (formalized in Section~\ref{sec:swe:sft}).
We additionally apply the following quality filters:
(1)~trajectories exceeding 64K tokens are pruned to ensure training stability;
(2)~trajectories containing tool calls not defined in the harness's tool interface (primarily observed with Qwen3.5-397B) are discarded;
(3)~trajectories with syntactically invalid or unparsable actions are removed.

\begin{table}[t]
\centering
\caption{Composition of the Orchard-SWE training dataset. The corpus retains both resolved and unresolved trajectories: resolved trajectories provide direct imitation signal, while unresolved trajectories contribute partial-progress signal through credit-assignment SFT.}
\label{tab:swe-data}
\small
\setlength{\tabcolsep}{4pt}
\begin{tabular}{@{}llrrrl@{}}
\toprule
\textbf{Source} & \textbf{Teacher} & \textbf{Resolved} & \textbf{Unresolved} & \textbf{Total} & \textbf{Harness} \\
\midrule
Scale-SWE       & MiniMax-M2.5 & 33{,}527 & 20{,}591 & 54{,}118 & mini-swe-agent \\
SWE-rebench     & MiniMax-M2.5 & 13{,}364 & 10{,}099 & 23{,}463 & mini-swe-agent \\
SWE-rebench     & Qwen3.5-397B & 15{,}545 &  1{,}846 & 17{,}391 & mini-swe-agent \\
SWE-rebench & MiniMax-M2.5 & 12{,}213 &        0 & 12{,}213 & OpenHands \\
\midrule
\textbf{Total}  &              & \textbf{74{,}649} & \textbf{32{,}536} & \textbf{107{,}185} & \\
\bottomrule
\end{tabular}
\end{table}

After filtering, the Orchard-SWE dataset comprises \textbf{107K trajectories} (74.6K resolved, 32.5K unresolved) spanning \textbf{19{,}287 unique task instances}, with an average of 47.5 interaction turns and approximately 21K tokens per trajectory.
We release the full dataset, including both resolved and unresolved trajectories, as an open-source artifact.


\subsection{Off-policy Learning: Supervised Fine-Tuning with credit assignment}
\label{sec:swe:sft}
We initialize from the base backbone and fine-tune on the curated teacher trajectories.
Each training example pairs the issue description and repository context with the full multi-turn interaction trace, serialized as a sequence of observations and actions.
Following standard practice for long-horizon agent training, we apply multi-turn masking so that environment observations are excluded from the loss and the model is trained only to predict its reasoning traces and actions.

A distinguishing feature of our SFT stage is the use of \textbf{credit-assignment SFT}, which incorporates not only the 74.6K resolved trajectories but also a curated subset of unresolved trajectories where partial progress is identifiable.
We instantiate credit assignment as a lightweight LLM-based variant of temporal-difference value estimation, formulated as follows.

\paragraph{Retrospective value estimation.}
For each unresolved trajectory $\tau = (s_0, a_0, s_1, \ldots, s_T)$, we use the trajectory's own teacher model as a zero-shot retrospective value function.
The teacher is shown the full trajectory together with the gold test outcome and is asked to estimate, at each step $t$, the probability that the agent will resolve the issue given the history $h_t = (s_0, a_0, \ldots, s_t)$:
\begin{equation}
V(s_t) \;=\; \mathbb{P}\bigl(\text{resolve} \,\big|\, h_t,\, \text{outcome}\bigr).
\end{equation}
The teacher annotates a sparse set of key steps and the remaining values are interpolated, yielding a per-step value curve $V(s_0), \ldots, V(s_T)$.
Because the judgment is retrospective and outcome-conditioned, the value curve calibrates well to actual progress: across our annotated trajectories, the curve is inverted-U in 98.9\% of cases, peaking during exploration and decaying near the failed submission.
The exact value-estimation prompt is shown in Appendix \ref{app:swe:prompts}.

\paragraph{Rise-segment extraction.}
We define the per-step credit as the temporal-difference shift in estimated success probability,
\begin{equation}
c_t \;=\; V(s_{t+1}) - V(s_t),
\end{equation}
and extract \emph{rise segments}: maximal contiguous subsequences $[t_i, t_j]$ over which the agent makes positive progress, i.e.~$c_t \geq \varepsilon$ for all $t \in [t_i, t_j{-}1]$ (with a small threshold $\varepsilon$ to filter annotation noise).\footnote{We use $\varepsilon = 0.05$; see Appendix for sensitivity analysis.}
Rise segments are typically short (median $\sim$2 steps before merging with surrounding context) but capture the productive parts of an otherwise unsuccessful trajectory---repository navigation, file localization, and partial root-cause analysis.

\paragraph{SFT objective.}
We train with a standard next-token cross-entropy loss on action tokens, masking environment observations as well as action tokens that fall outside any rise segment:
\begin{equation*}
\mathcal{L}_{\text{SFT}} \;=\; -\sum_{t \in \mathcal{S}(\tau)} \log \pi_\theta\bigl(a_t \,\big|\, h_t\bigr),
\end{equation*}
where $\mathcal{S}(\tau)$ is the set of action tokens contributing to the loss for trajectory $\tau$.
For resolved trajectories, $\mathcal{S}(\tau)$ contains all action tokens (equivalent to a single segment spanning the entire trace, since the terminal value is~1).
For unresolved trajectories, $\mathcal{S}(\tau)$ is restricted to action tokens inside the extracted rise segments, with the preceding history retained as context.
After this construction, the 32{,}536 unresolved trajectories yield exploration-focused supervision that complements the full solve-and-submit traces from the resolved set.

We train Qwen3-30B-A3B-Thinking~\citep{qwen3technicalreport} with slime~\citep{slime_github} for five epochs with a global batch size of 128 and a 64K context window, using a cosine-decayed learning rate from $10^{-5}$ to $10^{-6}$.
Although training uses a maximum sequence length of 64K, we extend the context limit to 128K at inference time to accommodate longer repository contexts and interaction histories.

\subsection{On-policy Learning}

Starting from the SFT checkpoint, we improve the policy with Reinforcement
Learning (RL). Our recipe is organized around the two questions that dominate the
behavior of RL on long-horizon, sparse-reward agentic tasks: \emph{how we
sample rollouts} and \emph{how we reward them}. 

\paragraph{(1) Sampling via Balanced Adaptive Rollout (BAR, \S\ref{sec:swe:rl_bar}).}
BAR replaces fixed-$N$ group sampling with an
\emph{online, per-prompt, stride-based} schedule that keeps generating only
until it can assemble a fixed-size group whose positive-reward fraction lands
in a target interval, raising the information density of every gradient batch.
Unlike prior dynamic-sampling or difficulty-filtering methods that
\emph{discard} zero-variance prompts, \emph{pre-filter} on historical success
rates, or \emph{post-hoc down-sample} oversized rollout sets, BAR assembles the
group online and per prompt, and composes with any group-relative estimator.
 
\paragraph{(2) Dense reward functions (\S\ref{sec:swe:opd}--\S\ref{sec:swe:rpr}).}
Beyond the sparse, environment-grounded outcome reward, we add two
\emph{complementary} dense signals, each targeting a weakness the outcome
reward cannot address. \textbf{On-Policy Distillation (OPD, \S\ref{sec:swe:opd})}
adds a \emph{token-level} signal from a stronger teacher, densifying
supervision along the \emph{teacher-distribution-matching} axis; it is,
however, only as reliable as the teacher and is silent about whether the agent
actually verified its own work. \textbf{Rubric-based Process Reward (RPR, \S\ref{sec:swe:rpr})} adds a
\emph{trajectory-level} signal from an LLM-as-a-judge scored against a
process-quality rubric, densifying along the \emph{verification-quality} axis
by rewarding disciplined problem-solving behavior—writing tests that reproduce
the issue, asserting the right behavior, and respecting failures—which the
outcome reward leaves entirely unsupervised.

\subsubsection{Reinforcement Learning with Balanced Adaptive Rollout (BAR)}
\label{sec:swe:rl_bar}

For challenging agentic tasks such as SWE, the standard fixed-$N$ group rollout used by GRPO~\citep{shao2024deepseekmathpushinglimitsmathematical} has two major problems: 1) {\it Wasted compute}. When the policy is competent on a prompt, all $N$ trajectories tend to succeed; when it is not, all $N$ tend to fail. In both regimes the resulting group has zero reward variance, contributes a zero advantage to every token, and is silently discarded — yet we have already paid for $N$ long, environment-bound trajectories.
2) {\it Group-imbalance noise}. When the success rate of a prompt is far from $0.5$, even a "non-degenerate" group is dominated by either positives or negatives, and the resulting GRPO advantages are noisy and biased toward whichever class is over-represented.

\begin{algorithm}[t]
\small
\caption{Balanced Adaptive Rollout (BAR) for a single prompt}
\label{alg:bar}
\begin{algorithmic}[1]
\Require Prompt $x$; policy $\pi_\theta$; reward function $R$; environment factory $\mathcal{E}$
\Require Training group size $N$; max budget $N_{\max}$; stride $s$ (with $s \mid N_{\max}$)
\Require Positive-fraction interval $[\rho_{\min}, \rho_{\max}]$; ideal ratio $\rho^\star = (\rho_{\min}+\rho_{\max})/2$
\Ensure A training group $\mathcal{G} \subset \mathcal{T}$ with $|\mathcal{G}| = N$
\State $\mathcal{T} \gets \emptyset$ \Comment{pool of completed trajectories}
\For{$t = 0, s, 2s, \ldots, N_{\max} - s$}
    \State $\mathcal{B}_t \gets \{\,\tau_i \sim \pi_\theta(\cdot \mid x; \mathcal{E}) \,:\, i = 1, \ldots, s\,\}$ \Comment{generate stride in parallel}
    \State $r_i \gets R(\tau_i)$ for each $\tau_i \in \mathcal{B}_t$
    \State $\mathcal{T} \gets \mathcal{T} \cup \mathcal{B}_t$
    \State $(\mathcal{G}, \texttt{ok}) \gets \textsc{TryAssemble}(\mathcal{T}, N, \rho_{\min}, \rho_{\max}, \rho^\star)$
    \If{\texttt{ok}}
        \State \Return $\mathcal{G}$ \Comment{early-stop: balanced group found}
    \EndIf
\EndFor
\State $(\mathcal{G}, \texttt{ok}) \gets \textsc{TryAssemble}(\mathcal{T}, N, 0, 1, \rho^\star)$ \Comment{relaxed fallback}
\If{\texttt{ok}} \Return $\mathcal{G}$ \EndIf
\State \Return $\textsc{TopRanked}(\mathcal{T}, N)$ \Comment{best-effort fallback}
\Statex
\Procedure{TryAssemble}{$\mathcal{T}, N, \rho_{\min}, \rho_{\max}, \rho^\star$}
    \State Partition $\mathcal{T}$ into $\mathcal{T}_+ = \{\tau : R(\tau) > 0,\ \tau \text{ usable}\}$,
           $\mathcal{T}_- = \{\tau : R(\tau) \le 0,\ \tau \text{ usable}\}$,
           and put the rest (aborted / time-exceeded) into a backfill pile.
    \State Sort $\mathcal{T}_+$ and $\mathcal{T}_-$ by (status, response length) ascending
           \Comment{prefer completed and concise}
    \If{$|\mathcal{T}_+| + |\mathcal{T}_-| < N$} \Return $(\bot,\ \texttt{false})$ \EndIf
    \State $n^\star \gets \mathrm{round}(\rho^\star \cdot N)$;\quad
           $n_{\min} \gets \lceil \rho_{\min} \cdot N \rceil$;\quad
           $n_{\max} \gets \lfloor \rho_{\max} \cdot N \rfloor$
    \For{$n_+ \in \{n_{\min}, \ldots, n_{\max}\}$ sorted by $|n_+ - n^\star|$}
        \State $n_- \gets N - n_+$
        \If{$n_+ \le |\mathcal{T}_+|$ \textbf{and} $n_- \le |\mathcal{T}_-|$}
            \State \Return $(\mathcal{T}_+[1{:}n_+] \cup \mathcal{T}_-[1{:}n_-],\ \texttt{true})$
        \EndIf
    \EndFor
    \State \Return $(\bot,\ \texttt{false})$
\EndProcedure
\end{algorithmic}
\end{algorithm}

We address both issues with {\it Balanced Adaptive Rollout (BAR)}, a progressive, group-aware rollout algorithm. 
Unlike prior dynamic sampling and difficulty-filtering methods that discard zero-variance prompts
\citep{yu2025dapo,le2025nopromptleftbehind}, pre-filter prompts using historical success rates
\citep{bae2026online,zheng2025greso}, or post-hoc down-sample oversized rollout sets
\citep{xu2025pods,shang2025rstar2agent,zhang2026aero}, our method performs online,
per-prompt, stride-based group assembly. It adaptively continues generation only until it can
construct a fixed-size training group whose positive-reward fraction lies in a target interval,
while accounting for trajectory status, truncation, sandbox failures, and length. This makes the
rollout scheduler directly compatible with group-relative estimators in long-horizon agentic
environments.

For each prompt we set three quantities: a {\it training group size} $N$ (the number of trajectories the optimiser will actually consume), a {\it maximum budget} $N_{\max}>N$ (an upper bound on how many trajectories we are willing to generate), and a {\it stride} $s$ (the size of an incremental generation batch). We additionally specify a target positive-reward fraction interval $[\rho_{\min},\rho_{\max}]$, with ideal ratio $\rho^{\star}=(\rho_{\min}+\rho_{\max})/2$.
The algorithm proceeds as follows. We generate $s$ trajectories, score them with the reward model, and partition the pool of completed trajectories into a {\it positive} set (trajectories with reward $>0$) and a {\it negative} set (otherwise), after first moving aborted or truncated trajectories to a backfill pile. We then attempt to assemble a training group of exactly $N$ trajectories whose positive fraction lies in $[\rho_{\min},\rho_{\max}]$ and that is closest to the ideal ratio $\rho^{\star}$. Within each class, trajectories are ranked by terminal status (completed > truncated > aborted)\footnote{Other ranking criteria could also be used here, for example ranking by trajectory length, model likelihood, diversity, or estimated uncertainty, etc.
}, so as to prefer succinct, well-terminated trajectories. If a feasible group exists, we early-stop and return it; otherwise we generate another stride and retry. If after $N_{\max}$ trajectories no balanced group can be built, we fall back to a relaxed selection (any positive fraction in $(0,1)$), padding with the best backfill trajectories as needed.

BAR therefore behaves as an anytime, self-pacing rollout schedule: 1) Easy or already-mastered prompts (where the first stride is overwhelmingly positive) trigger no further generation — the prompt is either filtered out or returned with the minimum positives needed to satisfy the lower bound; 2) Hard prompts (where positives are rare) keep generating until either enough positives are discovered or the budget $N_{\max}$ is exhausted; 3) Well-balanced prompts finish near the first stride and yield maximally informative gradients.
The result increases the average information density of every gradient batch. Importantly, BAR composes cleanly with GRPO, GSPO~\citep{zheng2025group}, and any other group-relative advantage estimator, because the contract it must satisfy is simply ``return a list of $N$ trajectories per prompt''.

\paragraph{Final Group Filtering.}
Because rewards are produced by a noisy, real environment (sandbox creation can time out, containers can be evicted, the LLM can hit its token budget mid-step), some trajectories carry no usable learning signal even within an otherwise valid group. We therefore plug a group-level filter into the rollout loop, evaluated after reward computation and before the group is admitted into the training batch. A dropped group is simply replenished from over-sampled prompts, which decouples the training batch size from the generation batch size and preserves gradient quality.

Filtering and the Balanced Adaptive Rollout (BAR) are designed to work jointly: BAR maximises the probability that a generated group satisfies the filter on the first try, and the filter provides a hard correctness guarantee on whatever BAR returns. Together they implement a form of reward-aware curriculum that is performed online, at every gradient step.
Together these components form a fault-tolerant, throughput-optimised pipeline for end-to-end RL on long-horizon, sandbox-grounded agentic tasks, with Balanced Adaptive Rollout turning a fixed-batch rollout into a self-pacing, information-dense one. The final estimated advantage for rollout trajectory $o_i$ within a group of $N$ samples is normalized using the group rewards 
$
A_{i,t} =
\frac{
r_i - \operatorname{mean}(\{r_1, r_2, \ldots, r_N\})
}{
\operatorname{std}(\{r_1, r_2, \ldots, r_N\})
}
$
, where $r_i$ denotes the reward assigned to rollout trajectory $o_i$. RL train for at most 150 steps with a global batch size of 128 and a 64K context window, using a cosine-decayed learning rate from $10^{-6}$ to $10^{-7}$. We apply the rollout batch size 16 with a training group size of $N=8$, maximum budget $N_{max}=16$ and stride $s=16$ to encourage parallel rollout. The target positive-reward fraction interval used is $[\rho_{\min},\rho_{\max}]=[0.375, 0.625]$.

\paragraph{Data Selection.} 
\begin{table*}[tb]
\centering
\caption{Performance on SWE-rebench v2. Baseline results with $^\dagger$ are from \cite{badertdinov2026swe}, where models were evaluated on a 60-task Python subset. Orchard-SWE is evaluated on the full Python subset. All models are evaluated using the mini-swe-agent harness.}
\label{tab:orchard-swe-swerebench-v2}
\small
\setlength{\tabcolsep}{6pt}
\begin{tabular}{@{}lcc@{}}
\toprule
\textbf{Model} & \textbf{pass@1} & \textbf{pass@3} \\
\midrule
Claude Opus-4.5$^\dagger$ & 36.11\% & 36.67\% \\
GLM-4.7$^\dagger$         & 27.22\% & 31.67\% \\
MiniMax-M2.1$^\dagger$    & 26.11\% & 31.67\% \\
Gemini$^\dagger$          & 25.56\% & 33.33\% \\
DeepSeek-V3.2$^\dagger$   & 23.33\% & 31.67\% \\
GPT-5.2$^\dagger$         & 20.56\% & 23.33\% \\
gpt-oss-120b$^\dagger$    & 8.89\%  & 16.67\% \\
\midrule
Orchard-SWE               & 22.36\% & 27.94\% \\
\bottomrule
\end{tabular}
\end{table*}
For RL training, we construct a task pool using all Python subset data of SWE-rebench V2 and Scale-SWE data that were not used during SFT. We first run the initial SFT model on each candidate task with 8 rollouts to get its initial pass rate. Table~\ref{tab:orchard-swe-swerebench-v2} reports the performance of initial Orchard-SWE SFT checkpoint on SWE-rebench V2 alongside baseline results. Orchard-SWE SFT achieves 22.36\% pass@1 and 27.94\% pass@3 on the full Python subset.  We then retain only tasks with pass rate $0 < \hat{p} \leq 0.5$, filtering out tasks that are either too difficult to provide a reliable learning signal or already too easy for the SFT model. This selection is particularly important for SWE-rebench V2, which is highly challenging: as shown in the table, even state-of-the-art models achieve 20\%-40\% pass rates on this dataset. After filtering, the final RL training set contains approximately 2k instances. We apply mini-swe-agent as the harness for RL training.

\subsubsection{Denser Reward: On Policy Distillation (OPD)}
\label{sec:swe:opd}
Unlike reinforcement learning based on sparse outcome rewards, on-policy distillation (OPD) \citep{agarwal2024onpolicy,lu2025onpolicy} has emerged as an attractive post-training technique by providing a dense reward signal.
Specifically, rather than evaluating an entire student rollout solely on final task completion feedback, a teacher model scores individual actions (tokens) to provide fine-grained, step-by-step guidance.

For OPD, per-token loss functions are used to evaluate student trajectories. A standard choice is the reverse KL divergence:
\begin{equation}
D(\pi_\theta || \pi_\text{teacher}) = \mathbb{E}_{\pi_\theta} \left[ \log \pi_\theta(x_{t+1}|x_{1\ldots t}) - \log \pi_\text{teacher}(x_{t+1}|x_{1\ldots t}) \right].
\end{equation}
Because computing the exact KL divergence over the full vocabulary distribution at every token is computationally and memory intensive, point-wise sample estimators are commonly used for optimization, such as the $K_1, K_2$, and $K_3$ variants studied by \cite{schulman2020kl}. Due to the high variance of $K_1$, we restrict our focus to $K_2$ and a stabilized $K_3$ variant ($K_3^*$):
\begin{align}
\label{eqn:k2}
& K_2   = \frac{1}{2}\left(\log \pi_\theta - \log \pi_\text{teacher}\right)^2,\\
\label{eqn:k3}
& K_3^* = \frac{\pi_\theta}{\texttt{sg}(\pi_\theta)} \underbrace{\left[\frac{\pi_\text{teacher}}{\pi_\theta} - 1 + \log \pi_\theta - \log \pi_\text{teacher}\right]}_{K_3},
\end{align}
where $\texttt{sg}(\cdot)$ denotes the stop-gradient operator, ensuring that the gradient of $K_3^*$ remains consistent in expectation with the exact reverse KL.

Since both forward and reverse KL divergences are instances of the broader $f$-divergence family \citep{csiszar2004information}, it is natural to extend OPD to general $f$-divergences:
\begin{equation}
D_f(\pi_\text{teacher} || \pi_\theta) = \mathbb{E}_{\pi_\theta} \left[ f\left(\frac{\pi_\text{teacher}}{\pi_\theta}\right) \right],
\end{equation}
where $f(\cdot)$ is a convex, lower-semicontinuous generator function satisfying $f(1)=0$. The gradient of the $f$-divergence with respect to the student policy parameters can be derived as:
\begin{align}
\nabla_\theta D_f(\pi_\text{teacher} || \pi_\theta)
&= \mathbb{E}_{\pi_\theta} \left[
\underbrace{\left(f(r) - r f'(r)\right)}_{\text{$f$-induced reward}} \nabla_\theta \log \pi_\theta
\right],
\end{align}
where $r = \frac{\pi_\text{teacher}}{\pi_\theta}$ and $f'$ denotes the derivative of $f$. Notice that the term $f(r) - r f'(r)$ plays the exact role of an intrinsic advantage/reward signal $R$ in a REINFORCE update, enabling any choice of $f$-divergence to naturally induce a score-function gradient estimator.

A particularly compelling choice within this family is the Jensen–Shannon (JS) divergence, which yields the following REINFORCE-style gradient:
\begin{equation}
\label{eqn:js_grad}
\mathbb{E}_{\pi_\theta} \left[ \log \left( \frac{\pi_\theta}{\lambda \pi_\theta + (1-\lambda) \pi_\text{teacher}} \right) \nabla_\theta \log \pi_\theta \right],
\end{equation}
with $\lambda = 0.5$. Unlike the reverse KL gradient, which can suffer from extreme values due to severe probability mismatches between student and teacher, JS divergence pulls the student policy toward an equal mixture point. By tuning $\lambda$, this reward can be interpreted as a trust-region mechanism that promotes more stable training in OPD \citep{xie2026trustregionpolicydistillation}.

\begin{table}[t]
\centering
\caption{\textbf{On-Policy Distillation (OPD) on SWE-bench Verified.} All models are trained with Qwen3.5-35B-A3B as the student and Qwen3.5-397B-A17B as the teacher. Results report the resolved ratio (\%), averaged across three evaluation runs based on Mini-SWE agent.}
\label{tab:opd_loss_comparison}
\small
\begin{tabular}{lc}
\toprule
\textbf{Divergence / Loss} & \textbf{SWE-bench Verified ($\uparrow$)} \\
\midrule
Base Model & 61.4 \\
\midrule
Reverse KL ($K_2$, Eq.~\ref{eqn:k2}) & 65.2 \\
Reverse KL ($K_3^*$, Eq.~\ref{eqn:k3}) & 65.6 \\
\midrule
JS Surrogate (Eq.~\ref{eqn:js_loss}) & 66.1 \\
\midrule
GRPO-Outcome + OPD-JS & 69.3 \\
\bottomrule
\end{tabular}
\end{table}

\paragraph{Experiments:} Here, we first empirically evaluate the reverse KL variants ($K_2, K_3^*$) and the JS-induced surrogate loss
\begin{equation}
\label{eqn:js_loss}
\texttt{sg}\left(\log \left( \frac{\pi_\theta}{\lambda \pi_\theta + (1-\lambda) \pi_\text{teacher}}\right) \right) \log \pi_\theta.
\end{equation}
We use \texttt{Qwen3.5-35B-A3B} and \texttt{Qwen3.5-397B-A17B} as the student and teacher model, respectively.

\autoref{tab:opd_loss_comparison} summarizes the resolved ratios on SWE-bench Verified using \texttt{Mini-SWE} agent, averaged over three inference runs.
As shown in \autoref{tab:opd_loss_comparison}, OPD training with all three losses consistently improve performance over the base student model, with JS-induced loss slightly outperforming the reverse KL variants.
Beyond task performance, JS leads to noticeably more stable training, exhibiting fewer gradient norm spikes and maintaining healthy policy entropy throughout the training.
Furthermore, because the OPD JS loss provides denser supervision signal, we experiment with augmenting sparse outcome-reward GRPO with it.
To isolate the impact of this dense reward, we omit the BAR technique. 
As shown in \autoref{tab:swe-results}, combining GRPO with OPD JS yields further improvements, matching GRPO+BAR performance while reducing rollout costs.
This confirms that OPD provides an effective dense reward signal that significantly accelerates policy learning.

\subsubsection{Denser Reward: Rubric-based Process Reward (RPR)}
\label{sec:swe:rpr}

\begin{figure*}[t]
  \centering
  \includegraphics[width=\textwidth]{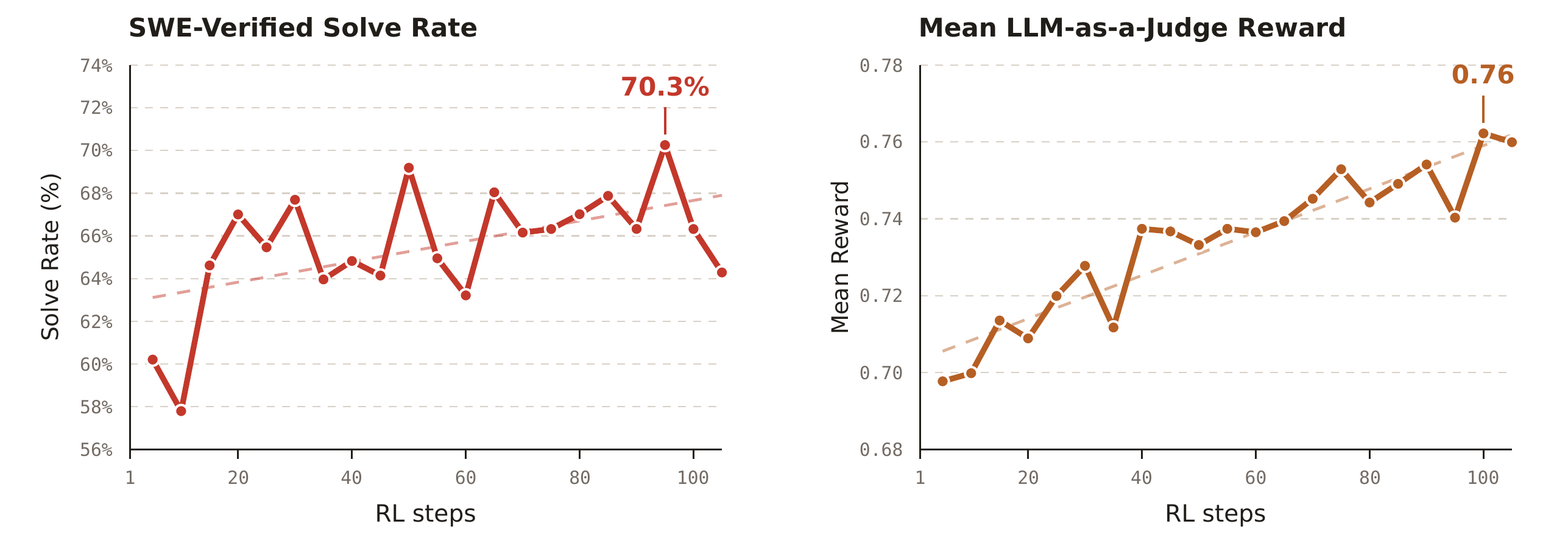}
  \caption{Outcome and process rewards over RL training. \textbf{Left:}
  SWE-Verified solve rate, rising from about 60\% to a peak of 70.3\%.
  \textbf{Right:} mean LLM-as-a-Judge reward, climbing steadily to 0.76.
  Dashed lines show the linear trend in each panel.}
  \label{fig:swe-rpr-curves}
\end{figure*}


On-Policy Distillation densifies the RL signal by matching the student's
per-token distribution to a teacher's; it is therefore only as reliable as the teacher and says nothing
about whether the agent verified its own work. We now densify along a \emph{complementary} axis—the
quality of the agent's problem-solving process. The starting point is again the binary, outcome-grounded
reward: a trajectory scores $+1$ only if its final patch passes the hidden
gold test suite. This signal is faithful but under-specified. It gives no gradient when an entire rollout
group fails, discarding the useful \emph{process}—localization, reproduction, partial root-cause
analysis—a failed trajectory may still contain; and it is indifferent to \emph{how} a patch passed, scoring
one that ignores a failing check and submits anyway identically to one that reproduces the bug, verifies
the fix, and guards against regressions. Outcome reward thus cannot tell a robust fix from a lucky one, and
leaves the agent's self-verification behavior unsupervised—the gap the process reward (RPR) fills.
 
To densify this signal, we introduce a \emph{Rubric-based Process Reward (RPR)}: an
LLM-as-a-judge that scores an entire trajectory on \emph{process quality}, as
independently as possible of whether the tests ultimately passed. Rather than
grading only the final action, the RPR reads the whole interaction trace and
produces a scalar reward $r_{\mathrm{proc}}(\tau)\in[0,1]$ that rewards the
verification discipline an agent exhibits---writing tests that reproduce the issue,
asserting the right behavior, and respecting failures---which the sparse outcome
reward cannot capture.
 
\paragraph{Process-quality rubric.}
Given a trajectory $\tau=(s_0,a_0,s_1,\dots,s_T)$ and its final patch, the judge
scores seven interpretable dimensions (R1--R7), each on a score
$s_d\in[0,1]$. One dimension grades the \emph{fix} itself (R1),
while the remaining six grade the agent's \emph{self-verification} (R2--R7)---the
tests, reproduction scripts, and checks the agent wrote or ran to convince
\emph{itself} that the fix is correct. Table~\ref{tab:rpr-rubric} summarizes the
dimensions, what each measures, and its aggregation weight.

\paragraph{Reward aggregation.}
The seven dimensions are combined into a single scalar by a fixed weighted average.
The weights (Appendix Table~\ref{tab:rpr-rubric}) place the majority of the mass on self-verification
(R2--R7 sum to $0.85$), with the heaviest single weight on self-test relevance,
reflecting that a test which reproduces the issue behavior is the most trustworthy
evidence of a correct fix. Because the dimensions are scored independently and then
averaged, the judge is instructed to reason about each on its own merits and to
cite concrete step evidence (commands and outputs) rather than a holistic
impression.
 
\paragraph{Reference-guided grading.}
Grading self-verification quality in the abstract is unreliable, so when ground
truth is available we condition the judge on a held-out \textsc{[reference]} block
containing the official \texttt{FAIL\_TO\_PASS} test identifiers and the gold
patch. Crucially, this reference is used \emph{only} to ground the relevance,
coverage, and discriminativeness of the agent's own self-tests (R2, R3, R5): the
judge is explicitly instructed \emph{not} to grade exact-match against the gold
edit, and \emph{not} to penalize the agent for not running the hidden official tests.
 
\paragraph{Outcome-decoupled scoring.}
Because the corpus is outcome-resolved, most tracebacks and error strings in a
trajectory are benign noise from normal exploration. A naive judge that lowers a
score whenever an error string appears would therefore be badly miscalibrated. Our
core scoring principle instead ties every penalty to the agent's \emph{response} to
evidence---ignoring a failure, mislabeling test output, or submitting anyway---
rather than to the mere presence of an error in a tool output. This keeps the
process score decoupled from the terminal outcome and focused on verification
behavior.
 
\paragraph{Trajectory serialization.}
The judge sees a single rendered prompt per trajectory: the issue description, the
optional reference block, the step-by-step trace, and the final submitted patch.
Each step is serialized as its reasoning, the issued command, and a compressed
observation; long tool outputs are truncated head-and-tail so that test summaries
and final assertions survive, and large file or directory dumps are collapsed to a
size marker (the fact that a file was read matters more than its contents). The
judge returns a short evidence-citing rationale followed by the seven per-dimension
scores as a JSON object, which are then aggregated as above. The exact process-reward grading prompt is shown in Appendix \ref{app:swe:prompts}.

\subsection{Distilling Historical Rollouts into a Value Model}
\label{sec:historical_experience}
Agentic RL produces substantially more information than is ultimately absorbed by a single policy checkpoint.
During training, agents explore alternative solutions, encounter diverse failure modes, execute diagnostic commands and tests, and receive environment-grounded outcomes.
Once an experiment finishes, however, these trajectories are typically discarded even though generating and verifying them accounts for a substantial fraction of the cost of agentic RL.

We ask whether historical rollouts can instead become a persistent asset for future policies.
Specifically, we study whether trajectories collected from heterogeneous previous policies contain a transferable notion of solution quality that can be compressed into a small value model and applied to trajectories from a new policy.
We refer to this paradigm as \textbf{historical experience distillation}:
transforming environment-grounded trajectories accumulated by previous
policies into reusable learning signals that improve future policies.
Historical experience can be distilled directly into policy parameters
through continued training or SFT, as in the consolidation of rollouts
from specialist RL climbs~\citep{MAI-1-thinking}, or into auxiliary models
such as value model that support selection, exploration, and credit
assignment for RL training. In this work, we study the latter as a controlled first instantiation.

\paragraph{Problem formulation.}
Suppose $M$ previous RL experiments produce a historical experience pool
\[
\mathcal{D}_{\mathrm{hist}}
=
\bigcup_{j=1}^{M}
\left\{
\left(\tau_i^{(j)},r_i^{(j)}\right)
\right\}_{i=1}^{n_j},
\qquad
\tau_i^{(j)}\sim\pi_j,
\]
where $\pi_j$ is the policy checkpoint used in experiment $j$, $\tau_i^{(j)}$ is a complete agent trajectory, and $r_i^{(j)}\in\{-1,+1\}$ is its terminal environment reward. We use $\mathcal{D}_{\mathrm{hist}}$ to train a value model
\[
V_\phi(\tau)
\approx
P(r=+1\mid\tau).
\]
The value model is then evaluated on trajectories generated by a new policy
$\pi_{\mathrm{new}}$ whose rollouts were not included in value-model training.
Given $N$ candidates
\[
\tau_1,\ldots,\tau_N\sim\pi_{\mathrm{new}},
\]
we select
\[
\tau^*
=
\arg\max_{\tau_i}V_\phi(\tau_i).
\]

Unlike an on-policy critic trained jointly with its actor, the historical value model must transfer across the policy distribution shift between
$\{\pi_j\}_{j=1}^{M}$ and $\pi_{\mathrm{new}}$.
Its performance therefore measures whether historical trajectories contain reusable decision-quality structure rather than information specific to their source policies.

\paragraph{Historical experience pool.}
We construct $\mathcal{D}_{\mathrm{hist}}$ from trajectories accumulated across 20 prior Orchard-SWE RL experiments.
The experiments span different experimental configurations, producing a heterogeneous mixture of successful and unsuccessful behaviors.
Each trajectory contains the complete multi-turn interaction among the policy, agent harness, and execution environment, together with a terminal verifier outcome indicating whether the submitted patch resolves the task.
We retain only completed or failed trajectories with valid verifier outcomes.
To prevent a small number of frequently sampled tasks from dominating training, we cap the number of trajectories contributed by each instance.
We additionally downsample the majority outcome class to maintain a balanced mixture of successful and unsuccessful trajectories.

\paragraph{Value-model training.}
We instantiate the value model from a Qwen3-4B backbone by replacing its language-modeling head with a scalar value head. Training is entirely offline and requires no new environment interaction.
Each contiguous assistant response is treated as one decision step.
The value for turn $t$ is read from the state immediately preceding the corresponding action, while environment observations and tool-result tokens are excluded from the loss. The terminal verifier reward is assigned to the final turn and propagated backward through the trajectory using generalized advantage estimation with
$\gamma=0.98$ and $\lambda=0.95$. We regress the value model onto the resulting targets.
This procedure compresses the outcomes and behavioral patterns distributed across the historical experience pool into a standalone value model that is independent of any individual source policy.

\paragraph{Relation to the PPO critic.}
In PPO, the critic (or value model) is co-trained with the policy on states the \emph{current}
policy visits, so it chases a moving target, perpetually re-fitting to a thin,
non-stationary slice of on-policy states and suffering training instability, and the critic is discarded once training ends.
We instead pre-train the value model \emph{offline} on $\mathcal{D}_{\mathrm{hist}}$,
a large, outcome-verified corpus pooled from heterogeneous prior checkpoints
$\{\pi_j\}_{j=1}^{M}$, with no new environment interaction. 
Thus, compared to a co-trained critic, the pre-trained value model is more general and can be used to help train any new policies using policy optimization methods without introducing instability.
In this study, we show that the pre-trained value model improves a policy model via inference-time best-of-$N$ reranking. 

\begin{table*}[t]
\centering
\caption{\textbf{SWE-bench Verified resolve rates (\%).}
Methods are grouped by base-model family;
proprietary/frontier results are included for reference.
Orchard-SWE achieves the best performance among the open-source methods
compared, reaching \textbf{73.0\%}. Despite using a 35B-A3B model, it
approaches the performance of some frontier models that are over 10x larger.}
\label{tab:swe-results}
\scriptsize
\setlength{\tabcolsep}{2pt}
\begin{tabular}{@{}llccr@{}}
\toprule
\textbf{System} & \textbf{Base Model} & \textbf{Size} & \textbf{Harness} & \textbf{Resolved (\%)} \\
\midrule
\multicolumn{5}{@{}l}{\textit{Open-Source Methods: Qwen 2.5 Series}} \\
\midrule
Openhands-LM~\citep{wang2025openhands}       & Qwen2.5-Coder-32B-Inst.  & 32B-dense & OpenHands   & 37.2 \\
Skywork-SWE~\citep{zeng2025skyworkswe}        & Qwen2.5-Coder-32B-Inst.  & 32B-dense & OpenHands   & 38.0 \\
SWE-Agent-LM~\citep{yang2025swesmith}         & Qwen2.5-Coder-32B-Inst.  & 32B-dense & SWE-Agent   & 40.2 \\
SWE-Mirror-LM~\citep{wang2025swe}             & Qwen2.5-Coder-32B-Inst.  & 32B-dense & OpenHands  & 52.2 \\
daVinci-Dev-32B~\citep{zeng2026davincidev}    & Qwen2.5-32B-Base         & 32B-dense & SWE-Agent   & 56.1 \\
SWE-Compressor~\citep{liu2025context}         & Qwen2.5-32B-Base         & 32B-dense & OpenHands   & 57.6 \\
SWE-Master-32B-RL~\citep{song2026swemaster}   & Qwen2.5-Coder-32B-Inst.  & 32B-dense & R2E-Gym     & 61.4 \\
OpenSWE-32B~\citep{fu2026davincienvopensweenvironment} & Qwen2.5-32B-Base & 32B-dense & SWE-Agent   & 62.4 \\
SWE-Fixer-72B~\citep{xie2025swefixer}         & Qwen2.5-72B-Base         & 72B-dense & Agentless   & 32.8 \\
daVinci-Dev-72B~\citep{zeng2026davincidev}    & Qwen2.5-72B-Base         & 72B-dense & SWE-Agent   & 58.5 \\
Kimi-Dev~\citep{yang2025kimidev}              & Qwen2.5-72B-Base         & 72B-dense & Agentless   & 60.6 \\
OpenSWE-72B~\citep{fu2026davincienvopensweenvironment} & Qwen2.5-72B-Base & 72B-dense & OpenHands    & 65.0 \\
OpenSWE-72B~\citep{fu2026davincienvopensweenvironment} & Qwen2.5-72B-Base & 72B-dense & SWE-Agent    & 66.0 \\
\midrule
\multicolumn{5}{@{}l}{\textit{Open-Source Methods: Qwen 3 Series}} \\
\midrule
SWE-Lego-Qwen3-32B~\citep{tao2026swelego}     & Qwen3-32B                & 32B-dense & OpenHands       & 52.6 \\
FrogBoss~\citep{sonwane2025bugpilot}          & Qwen3-32B                & 32B-dense & R2E-Gym         & 54.6 \\
CoderForge-32B~\citep{ariyak2026coderforge}   & Qwen3-32B                & 32B-dense & OpenHands       & 59.4 \\
Qwen3-Coder-30B-A3B-Instruct                  & ---                      & 30B-A3B   & OpenHands       & 51.6 \\
GLM-4.7-Flash-30A3B~\citep{5team2025glm45agenticreasoningcoding} & --- & 30B-A3B & ---            & 59.2 \\
Scale-SWE-Agent~\citep{zhao2026immersion}     & Qwen3-30B-A3B-Instruct   & 30B-A3B   & OpenHands       & 64.0 \\
Qwen3-30B-A3B-Thinking (baseline)                   & ---                      & 30B-A3B   & OpenHands       & 22.0 \\
\textbf{Orchard-SWE-Qwen3 (SFT)~}                            & Qwen3-30B-A3B-Thinking   & 30B-A3B   & OpenHands       & 62.1 \\
\textbf{Orchard-SWE-Qwen3 (SFT)~}                            & Qwen3-30B-A3B-Thinking   & 30B-A3B   & mini-swe-agent  & 64.3 \\
\textbf{Orchard-SWE-Qwen3 (SFT+RL)}                 & Qwen3-30B-A3B-Thinking   & 30B-A3B   & mini-swe-agent  & 67.5 \\
\midrule
\multicolumn{5}{@{}l}{\textit{Proprietary / Frontier Models}} \\
\midrule
GLM-5.1~\citep{zai_glm51_2026}                 & ---  & 744B-A40B  & SWE-Agent & 75.0 \\
Seed-2.1 pro~\citep{bytedance_seed21_2026}     & ---  & ---        & SWE-Agent & 76.0 \\
Hy3~\citep{tencent_hy3_2026}                   & ---  & 295B-A21B  & SWE-Agent & 78.0 \\
DeepSeek-V4 flash~\citep{deepseek_v4_2026}     & ---  & 284B-A13B  & SWE-Agent & 79.0 \\
Qwen-3.7 Max~\citep{qwen37max_2026}            & ---  & ---        & SWE-Agent & 80.4 \\
DeepSeek-V4 pro~\citep{deepseek_v4_2026}       & ---  & 1.6T-A49B  & SWE-Agent & 80.6 \\
Gemini-3.1-pro-preview~\citep{google_gemini31pro_2026}
                                                   & ---  & ---        & SWE-Agent & 80.6 \\
GLM-5.2~\citep{zai_glm52_2026}                 & ---  & 744B-A40B  & SWE-Agent & 84.2 \\
GPT-5.5~\citep{openai_gpt55_2026}              & ---  & ---        & SWE-Agent & 84.4 \\
Claude-opus-4-8~\citep{anthropic_opus48_2026}  & ---  & ---        & SWE-Agent & 88.6 \\
\midrule
\multicolumn{5}{@{}l}{\textit{Our Method: Qwen 3.5 Series (3-run average)}} \\
\midrule
Qwen3.5-35B-A3B (baseline)                         & ---                      & 35B-A3B & mini-swe-agent & 61.4 \\
\textbf{Orchard-SWE-Qwen3.5 (RL with BAR)}                         & Qwen3.5-35B-A3B & 35B-A3B & mini-swe-agent & 69.1 \\
\textbf{Orchard-SWE-Qwen3.5 (RL with OPD)}                         & Qwen3.5-35B-A3B & 35B-A3B & mini-swe-agent & 69.3 \\
\textbf{Orchard-SWE-Qwen3.5 (RL with RPR)}                         & Qwen3.5-35B-A3B & 35B-A3B & mini-swe-agent & 69.7 \\
\textbf{\quad + reranking with value model}      & Qwen3.5-35B-A3B & 35B-A3B & mini-swe-agent & \textbf{73.0} \\
\bottomrule
\end{tabular}
\end{table*}

\subsection{Main Results}
\label{sec:swe:results}
 
Table~\ref{tab:swe-results} compares Orchard-SWE with open-source SWE-agent recipes
on SWE-bench Verified, organized by base-model family. We report results on two model families - Qwen3 and Qwen3.5. On \textbf{Qwen3-30B-A3B} we develop and validate the core
recipe—multi-teacher distillation, credit-assignment SFT, and BAR RL. On the stronger
\textbf{Qwen3.5-35B-A3B} we study how far RL can be pushed with denser reward signals
(OPD, RPR) and inference-time selection via a historical value model. Both backbones
activate only ${\sim}3$B parameters at inference.
 
\paragraph{Same-size family lift (Qwen3-30B-A3B).}
Orchard-SWE lifts
Qwen3-30B-A3B-Thinking by \textbf{45.5 absolute points} on SWE-bench Verified
($22.0\% \rightarrow 64.3\%$ after SFT, $\rightarrow 67.5\%$ after SFT+RL under
mini-swe-agent), and exceeds the code-specialized Qwen3-Coder-30B-A3B-Instruct
($51.6\%$) and the broader-distillation GLM-4.7-Flash-30A3B ($59.2\%$) by wide margins.
At this active-parameter budget it is competitive with or exceeds dense open baselines an
order of magnitude larger: it surpasses every Qwen2.5-32B and Qwen3-32B recipe in the
table
and dense 72B systems (Kimi-Dev $60.6\%$,
OpenSWE-72B $65.0$--$66.0\%$). 
The closest competitor at comparable scale is Scale-SWE-Agent ($64.0\%$), built on the same backbone family: Orchard-SWE matches it under SFT and outperforms it under SFT+RL. 
 
\paragraph{Denser rewards drive a second lift (Qwen3.5-35B-A3B).}
On the stronger Qwen3.5-35B-A3B backbone ($61.4\%$ base), we ask whether the sparse,
outcome-grounded RL reward can be densified for
additional gains. RL with BAR raises the base by $7.7$ points to
$69.1\%$. Both densification strategies improve further: On-Policy Distillation
(OPD) reaches $69.3\%$, \emph{matching} BAR while dispensing
with the outcome-verification rollouts and thereby lowering rollout cost, and the process
reward model (RPR) reaches \textbf{$69.7\%$}, the best
single-model result. The RPR gain is the more informative of the two: because it rewards
verification discipline rather than only terminal success (Figure~\ref{fig:swe-rpr-curves}), it improves solve rate and the agent's self-verification
behavior \emph{jointly}, addressing a failure mode that outcome reward alone leaves
unsupervised. Together these results establish a discovery that motivates the two new
methods: the environment-grounded outcome reward is not the only usable RL signal for
SWE agents—teacher-distribution matching (OPD) and process-quality scoring (RPR) each add gains on top of an already strong outcome-reward baseline.
 
\paragraph{Historical experience compounds at inference time.}
Reranking the RL-with-RPR policy with the historical value model of
Section~\ref{sec:historical_experience} (best-of-$N$, $N=6$) raises the resolve rate from $69.7\%$ to
\textbf{$73.0\%$}, a $+3.3$-point gain from a small 4B value model trained entirely
offline on rollouts accumulated in prior experiments. An oracle that selects a passing
candidate whenever one exists reaches $82.4\%$, leaving $12.7$ points of inference-time
selection headroom. This is our headline result: at $73.0\%$, Orchard-SWE is the strongest
open-source SWE agent in Table~\ref{tab:swe-results} and approaches frontier systems—GLM-5.1
($75.0\%$) and Seed-2.1~pro ($76.0\%$)—that are more than $10\times$ larger. The key
discovery is that trajectories accumulated across heterogeneous prior policies carry a
\emph{transferable} notion of solution quality: a value model distilled from them improves
a new, unseen policy with no additional environment interaction, turning otherwise-discarded
RL exhaust into a reusable inference-time asset.
 
\paragraph{Harness sensitivity.}
The same Orchard-SWE~(SFT) checkpoint reaches $64.3\%$ under mini-swe-agent but
$62.1\%$ under OpenHands, indicating that single-condition leaderboard numbers are sensitive
to harness choice. This sensitivity becomes the central empirical question of
Section~\ref{sec:swe:generalization}: evaluated alongside the closest open recipes
(Scale-SWE, OpenSWE-32B) across multiple harnesses and tasks, Orchard-SWE retains capability
on unseen harnesses and out-of-distribution tasks while other models collapse.

\subsection{Generalization to Unseen Harnesses and Tasks}
\label{sec:swe:generalization}

On SWE-bench Verified, Scale-SWE (64.0\%) and OpenSWE-32B (62.4\%) report similar resolve rates. 
Single-benchmark scores can mask large differences in how well an agent generalizes. We therefore evaluate each model across three harnesses---OpenHands, mini-swe-agent, and Kimi-CLI~\citep{moonshotai_kimi_cli_2026}---and three different tasks: SWE-bench Verified, SWE-bench Multilingual~\citep{yang2025swesmith}, and Terminal-Bench 2.0~\citep{tbench_2025}.
Kimi-CLI was not used during training data collection by any of the three systems, making it an unseen harness in this study.

\begin{table*}[t]
\centering
\caption{Generalization across harnesses and task distributions. Resolve rate (\%) is reported for each system under matched conditions. SWE-V = SWE-bench Verified; SWE-M = SWE-bench Multilingual; T-Bench 2.0 = Terminal-Bench 2.0. \emph{*} denotes numbers reported in the original paper; unmarked entries are our own evaluations under matched conditions; \xmark{} indicates the system produced malformed tool calls under that harness, yielding no valid resolve rate.}
\label{tab:swe-generalization}
\small
\setlength{\tabcolsep}{6pt}
\begin{tabular}{@{}lccccc@{}}
\toprule
                & \textbf{OpenHands} & \multicolumn{2}{c}{\textbf{mini-swe-agent}} & \multicolumn{2}{c}{\textbf{Kimi-CLI (unseen)}} \\
\cmidrule(lr){2-2}\cmidrule(lr){3-4}\cmidrule(lr){5-6}
\textbf{System} & SWE-V              & SWE-V & SWE-M                              & SWE-V & T-Bench 2.0                              \\
\midrule
Scale-SWE       & 64.0$^{*}$         & \xmark        & \xmark                     & \xmark        & \xmark                          \\
OpenSWE-32B     & 62.4$^{*}$         & 54.9         & 28.7                       & 3.6           & 0.0                             \\
\textbf{Orchard-SWE-Qwen3} & 62.1          & \textbf{64.3} & \textbf{51.0}              & \textbf{45.0} & \textbf{20.1}                   \\
\bottomrule
\end{tabular}
\end{table*}

\paragraph{Harness lock-in is severe in single-harness training.}
We find Scale-SWE~\citep{zhao2026immersion} produces invalid outputs under any harness other than its native one, yielding no measurable resolve rate.
OpenSWE-32B~\citep{fu2026davincienvopensweenvironment} remains structurally valid but degrades sharply: from 62.4\% on its native OpenHands to 54.9\% on mini-swe-agent ($-7.5$ pt) and 3.6\% on Kimi-CLI ($-58.8$ pt).
Orchard-SWE, in contrast, holds within a narrow band of 45.0--64.3\% across all three harnesses, with the worst-case drop bounded at 19.3 points relative to its own best.
The two failure modes observed in Scale-SWE and OpenSWE-32B (catastrophic format failure and degraded resolve rate) have the same root cause - a model trained under a single harness has not learned harness-agnostic SWE skills.
This pattern is exactly what our cross-harness ablation predicted in a controlled experiment (Section~\ref{sec:swe:ablations}, Table~\ref{tab:swe-harness}); here we see the same failure mode play out in two independently developed open recipes.

\paragraph{Cross-distribution generalization.}
On SWE-bench Multilingual under the mini-swe-agent harness, Orchard-SWE drops from 64.3\% (Verified) to 51.0\% ($-13.3$ absolute, $-20.7\%$ relative).
OpenSWE-32B drops from 54.9\% to 28.7\% ($-26.2$ absolute, $-47.7\%$ relative).
Orchard's relative drop is roughly half, indicating that multi-teacher distillation across SWE-rebench and Scale-SWE provides broader exposure to repositories and issue types than any single source alone.

\paragraph{Cross-domain transfer to Terminal-Bench 2.0.}
Terminal-Bench 2.0 evaluates a broader family of terminal interaction tasks beyond GitHub-issue resolution.
Under the Kimi-CLI harness, Orchard-SWE retains a 20.1\% resolve rate, while OpenSWE-32B drops to 0.0\%.
Both systems degrade substantially relative to their SWE-bench Verified scores, but only Orchard-SWE retains a non-trivial level of capability on this out-of-domain benchmark.
We hypothesize that broader trajectory diversity during training---multiple teachers, multiple harnesses, and multiple task sources---provides indirect exposure to more varied tool-use and terminal-interaction patterns than narrower training corpora.

\paragraph{Discussion.}
Two largely independent generalization improvements are visible in Table~\ref{tab:swe-generalization}: robustness across harnesses and across tasks. Both trace to diversity choices at three layers of the Orchard-SWE recipe. \textbf{Data design:} trajectories span two harnesses (mini-swe-agent, OpenHands), two teachers (MiniMax-M2.5, Qwen3.5-397B), and two complementary task sources (SWE-rebench, Scale-SWE), yielding 107K trajectories that vary along harness, repository structure, and issue type. \textbf{Orchard Env:} it makes data-design choices practical at scale and exposes only sandbox lifecycle, command execution, and file I/O---and imposes no assumptions about the harness or tool schema sitting above it---any harness can compose with the same env layer at zero adaptation cost. \textbf{Learning design:} credit-assignment SFT extracts partial-progress supervision from the unresolved trajectories that resolved-only recipes discard, broadening the exploration patterns the student sees, while Balanced Adaptive Rollout (BAR) keeps RL gradients informative by enforcing a balanced positive/negative mix within each prompt group across the full difficulty spectrum.
However, Orchard-SWE improves but does not solve generalization: the drop to Kimi-CLI Verified is substantial, and entirely-unseen harness or domain conditions still remain a meaningful challenge.

\subsection{Ablations and Analysis}
\label{sec:swe:ablations}

We ablate key design choices in Orchard-SWE-Qwen3 to understand their contribution to the final result. Four questions guide our analysis: \emph{(i)}~how does data scale interact with selection strategy? \emph{(ii)}~how harness-coupled is the trained model? \emph{(iii)}~what does credit-assignment SFT contribute over resolved-only training? \emph{(iv)}~what does reinforcement learning add on top of large-scale SFT?

\paragraph{Data scale vs.~selection strategy.}
We first study how SFT performance depends on training data scale and selection strategy.
We hold the recipe fixed---same base model, same SFT hyperparameters, mini-swe-agent harness, no RL---and vary only the number of training trajectories ($N \in \{512, 1024, 2048\}$) and the strategy used to choose those trajectories from the resolved-trajectory pool.
The selection strategies fall into two families:

1) \textbf{Heuristic baselines} (no use of gold-patch information): \emph{Random}: uniform random sampling from the resolved pool.
\emph{Diverse repo}: maximize repository diversity by capping per-repo trajectory count.
\emph{Concentrated repo}: concentrate samples on the largest-resolved ``core'' repositories.
2) \textbf{Property-based selectors} (use gold-patch characteristics as a signal of issue complexity):
\emph{Multi-file}: prefer instances whose gold patch modifies multiple files.
\emph{Large diff}: prefer instances with larger gold-patch diffs.
\emph{Composite}: composite scoring over multiple gold-patch properties.

\begin{table*}[t]
\centering
\caption{Effect of data scale and selection strategy on SFT-only resolve rate (\%) on SWE-bench Verified, evaluated under the mini-swe-agent harness. All cells use identical hyperparameters; only $N$ (trajectories) and selection strategy vary. Per-row maximum is bolded.}
\label{tab:swe-scaling}
\small
\setlength{\tabcolsep}{6pt}
\begin{tabular}{@{}lcccccc@{}}
\toprule
 & \multicolumn{3}{c}{\textbf{Heuristic baselines}} & \multicolumn{3}{c}{\textbf{Property-based selectors}} \\
\cmidrule(lr){2-4}\cmidrule(lr){5-7}
\textbf{$N$} & Random & Diverse repo & Concentrated repo & Multi-file & Large diff & Composite \\
\midrule
512   & 45.8 & 44.0 & 47.9 & 47.5 & \textbf{49.5} & 48.1 \\
1024  & 50.3 & 49.0 & 51.3 & 51.6 & \textbf{52.2} & 52.1 \\
2048  & 52.8 & 52.2 & 53.4 & \textbf{54.2} & 52.9 & 53.7 \\
\bottomrule
\end{tabular}
\end{table*}

Table~\ref{tab:swe-scaling} reports SFT-only resolve rates on SWE-bench Verified.
Two patterns dominate the table.
First, \textbf{data scale dominates selection strategy} at every regime tested.
Doubling data twice (512 $\to$ 2048) on the worst-performing method (\emph{Diverse repo}, 44.0 $\to$ 52.2) yields a +8.2-point gain, larger than the entire 5.5-point spread across all selection strategies at $N=512$ and far larger than the 2.0-point spread at $N=2048$.
Second, \textbf{the spread across selection strategies shrinks monotonically with $N$}: 5.5 pt at $N=512$, 3.2 pt at $N=1024$, 2.0 pt at $N=2048$.
At sufficient data scale, the choice of selection strategy matters much less than scale itself.

A few specific behaviors are worth noting.
\emph{Large diff} attains the strongest small-$N$ result (49.5 at $N=512$) but saturates earliest, gaining only +0.7 pt from $N=1024$ to $N=2048$, plausibly because the pool of large-diff gold patches is finite and additional samples come from a distribution closer to the overall mean.
Counterintuitively, \emph{Concentrated repo} beats \emph{Diverse repo} at small $N$ by 3.9 points, with the gap shrinking to 1.2 points at $N=2048$: at small data scales, deeper exposure to a few repositories produces more transferable behaviors than thin coverage of many.
Property-based selectors edge out heuristic baselines at $N=512$ but converge to the baselines by $N=2048$, suggesting that gold-patch heuristics function as a sample-efficient prior that random sampling matches given enough data.

Even the worst (method, scale) cell in Table~\ref{tab:swe-scaling} (44.0\% at $N=512$) lifts the resolve rate by 22 absolute points over the underlying base model (22.0\%, Table~\ref{tab:swe-results}), confirming that even a small dose of high-quality SFT trajectories provides most of the structural lift over the base.
However, the entire ablation grid plateaus around 54\% under SFT-only at $N=2048$, while the full Orchard-SWE recipe---using the full 107K-trajectory corpus and adding RL---reaches 67.5\% on SWE-bench Verified.

What makes this kind of scaling practical is Orchard Env itself.
Its \emph{thin, harness-agnostic service boundary} lets the same env layer serve any harness, enabling multi-harness data collection and training at no additional infrastructure cost.
\emph{Image-agnostic agent injection} allows arbitrary task images to be added to the corpus without per-image rebuilds.
\emph{Low command-execution latency} (0.28\,s; Section~\ref{sec:orchard:eval}) keeps rollout throughput high.
And \emph{affordable cost} (an order of magnitude cheaper than managed alternatives; Table~\ref{tab:cost-comparison}) makes large scale data collection and RL rollout feasible for academic research groups.

\paragraph{Cross-harness generalization.}
To assess whether harness choice during training affects the trained model's ability to generalize at evaluation time, we run a controlled comparison: using 12K resolved trajectories on SWE-rebench distilled from MiniMax-M2.5, we vary only the collection harness and train two SFT models with otherwise identical recipes.
We then evaluate each model under both harnesses on SWE-bench Verified, yielding results in Table~\ref{tab:swe-harness}.

\begin{table}[t]
\centering
\caption{Cross-harness generalization on SWE-bench Verified. Rows are the harness used to collect training trajectories, columns are the harness used at evaluation. All four cells use identical training data (12K resolved trajectories on SWE-rebench, MiniMax-M2.5 teacher) and the same SFT recipe; only the harness pairing differs. Diagonal entries (matched train/eval harness) are bolded.}
\label{tab:swe-harness}
\small
\setlength{\tabcolsep}{8pt}
\begin{tabular}{@{}lcc@{}}
\toprule
 & \multicolumn{2}{c}{\textbf{Evaluation harness}} \\
\cmidrule(lr){2-3}
\textbf{Training harness} & mini-swe-agent & OpenHands \\
\midrule
mini-swe-agent & \textbf{57.9} & 19.0 \\
OpenHands      & 28.0          & \textbf{53.5} \\
\bottomrule
\end{tabular}
\end{table}

The cross-harness matrix reveals a sharp diagonal--off-diagonal gap.
Models evaluated on the same harness used during training reach 53.5--57.9\% resolve rate, but performance collapses to 19.0--28.0\% under the mismatched harness.
This asymmetry suggests that OpenHands trajectories, which expose richer tool semantics and more structured observations, transfer slightly better to the simpler mini-swe-agent setting than the reverse.
The dominant effect, however, is that the model has not learned harness-agnostic SWE skills: tool-call format, observation structure, and turn-level conventions are tightly coupled to the harness seen during training.
This finding aligns with concurrent observations on harness coupling reported in the Qwen3-Coder-Next report~\citep{cao2026qwen3}.
The implication is that no single-harness training corpus can produce an agent that generalizes well across the harness ecosystem; multi-harness training is necessary.

\paragraph{Effect of credit-assignment SFT.}
We isolate the contribution of credit-assignment SFT through a controlled, scale-matched comparison.
Starting from the full resolved pool, we sub-sample 32K resolved trajectories so that the resolved baseline matches the 32{,}536 unresolved-trajectory rise segments in size, and train two SFT models with otherwise identical recipes:
\emph{(i)}~resolved-only (32K trajectories), and
\emph{(ii)}~resolved + unresolved with credit-assignment SFT (32K resolved + 32K rise-segment trajectories).
On SWE-bench Verified, the resolved-only baseline reaches 59.3\%, while adding credit-assignment SFT improves the resolve rate to 61.2\%---a gain of $+1.9$ points.
This gain validates that credit-assignment SFT extracts useful supervision from otherwise-discarded unresolved trajectories rather than fitting noise.
In the full Orchard-SWE recipe, the same signal compounds with the larger 74.6K-resolved corpus, contributing to the headline 64.3\% on SWE-bench Verified.


\paragraph{Effect of reinforcement learning.}
A natural question is how RL's benefit depends on the strength of the SFT checkpoint it builds on---particularly for \emph{out-of-distribution} generalization, where heavy SFT may leave less surface for RL to preserve cross-distribution capability. We compare RL initialized from two SFT checkpoints that differ by roughly two orders of magnitude in supervision: a \emph{moderate} checkpoint (the Composite/$N{=}512$ cell of Table~\ref{tab:swe-scaling}, 48.1\% on SWE-bench Verified) and a \emph{heavy} checkpoint (the full 107K-trajectory recipe, 64.3\%). We evaluate on SWE-bench Verified (in-distribution) and SWE-bench Multilingual (OOD) under mini-swe-agent.
The two initial models respond to RL very differently. From the moderate init, RL improves both axes, with the OOD gain \emph{larger} than the in-distribution gain: Verified $48.1\% \rightarrow 50.1\%$ ($+2.0$ pt) and Multilingual $22.0\% \rightarrow 28.7\%$ ($+6.7$ pt). From the heavy init, RL still improves Verified ($64.3\% \rightarrow 67.5\%$, $+3.2$ pt) but Multilingual slightly regresses. We read this as a specialization effect: heavy SFT places the policy on a sharper mode of the training distribution, so on-policy refinement sharpens in-distribution behavior at the cost of OOD transfer; a moderate base retains more behavioral diversity, so the same RL signal acts as broad-coverage refinement rather than narrow optimization.


\section{Orchard-GUI}
\label{sec:gui}
This section presents Orchard-GUI, our instantiation of the Orchard training recipe for multi-modal Graphical User Interface (GUI) agents.
We describe the problem setting, trajectory collection pipeline, two-stage training recipe, main results on evaluation benchmarks.

\subsection{Problem Setting}
\label{sec:gui:setup}
We adopt the standard task formulation for browser-use agents: each task is defined by a starting URL and a natural-language user intent (\eg, \ ``Find a dog bed on Amazon that is washable and has a length of at least 30 inches'').
The agent must navigate from the provided \texttt{start\_url} within a browser interface, interact with live web pages, and complete the task by producing a natural-language final answer (or executing the requested action). Success is evaluated using an LLM-as-a-judge, which scores the trajectory against the user intent based on the final response and the sequence of screenshots.
We evaluate on three benchmarks: i) \textit{WebVoyager}~\citep{he2024webvoyager}; ii) \textit{Online-Mind2Web}~\citep{deng2023mind2web}; and iii) \textit{DeepShop}~\citep{lyu2025deepshop}. We use the same evaluation protocol as FARA~\citep{fara7b2025} and Molmo-Web~\citep{gupta2026molmoweb} for fair comparison.

\subsection{Generic Tool-Calling Agent Harness}
Rather than adopting a bespoke browser-agent harness such as Browser-Use\footnote{\scriptsize\url{https://github.com/browser-use/browser-use}}, we intentionally employ a generic multi-turn ReAct-style loop~\citep{yao2023react}. This design choice avoids conflating harness-specific effects with differences in data or training recipes, and more importantly, enables a unified paradigm for agentic learning that can generalize across domains and tasks beyond GUI navigation.

Specifically, 
each episode begins with a system prompt that specifies the agent’s role, high-level operating guidelines, and an action schema defined in the standard OpenAI \texttt{tools} format.
Following standard GUI-agent practice, we define a fixed action space of 13 atomic tools using the OpenAI tool-calling interface: \texttt{click}, \texttt{write}, \texttt{press\_keys}, \texttt{scroll}, \texttt{wait}, \texttt{drag}, \texttt{hover}, \texttt{goto\_url}, \texttt{go\_back}, \texttt{new\_tab}, \texttt{switch\_tab}, \texttt{close\_tab}, and the terminal \texttt{done(response)}. The \texttt{done(response)} action is the only mechanism for terminating an episode and serves as the sole carrier of the final user-facing output. Table~\ref{tab:browser-tools} provides one-line summaries of each tool, with full argument specifications deferred to Appendix~\ref{app:browser:tools}. 

\begin{table}[t]
    \centering
    \caption{Browser action space: 13 atomic tools grouped by family. Full argument signatures are listed in Appendix~\ref{app:browser:tools}.}
    \label{tab:browser-tools}
    \small
    \begin{tabular}{@{}llp{0.65\columnwidth}@{}}
    \toprule
    \textbf{Category} & \textbf{Tool} & \textbf{Description} \\
    \midrule
    \multirow{3}{*}{Pointer Mgmt.}
                     & \texttt{click}        & Mouse-click at a screen pixel; supports single/double click and left/right/middle button. \\
                     & \texttt{hover}        & Move the cursor to a pixel to reveal tooltips or open dropdowns. \\
                     & \texttt{drag}         & Drag-and-drop from a start pixel to an end pixel. \\
    \midrule
    \multirow{2}{*}{Keyboard Mgmt.}
                     & \texttt{write}        & Clear the focused input and type a string. \\
                     & \texttt{press\_keys}  & Press one or more keys, sequentially or as a hotkey combo. \\
    \midrule
    \multirow{4}{*}{Page Nav.}
                     & \texttt{scroll}       & Scroll the page or a sub-element by a fraction of the viewport. \\
                     & \texttt{goto\_url}    & Navigate the current tab to a given URL. \\
                     & \texttt{go\_back}     & Navigate back in the browser history. \\
                     & \texttt{wait}         & Pause for $N$ seconds to allow the page to settle. \\
    \midrule
    \multirow{3}{*}{Tab Mgmt.}
                     & \texttt{new\_tab}     & Open a new blank browser tab. \\
                     & \texttt{switch\_tab}  & Switch to the tab with the given 0-based index. \\
                     & \texttt{close\_tab}   & Close the current tab. \\
    \midrule
    Termination      & \texttt{done}         & End the episode and emit the final user-facing answer. \\
    \bottomrule
    \end{tabular}
\end{table}

The first user turn provides the task intent along with the initial browser observation, comprising the latest screenshot, viewport dimensions, and a tab summary (URL and title for each open tab) obtained from the task’s \texttt{start\_url}.
At each subsequent step, the model produces a reasoning trace inside \texttt{<think>...</think>} followed by one or more \texttt{<tool\_call>} blocks.
Each call is parsed and executed in the Orchard Env sandbox. The resulting tool response (e.g., \texttt{Succeed: click on <button> "Continue shopping"}) is combined with the updated observation and appended to the context as the next user turn, framed as feedback to the preceding action.
This loop repeats until the model emits \texttt{done} or a predefined step budget is exhausted.

As a single screenshot can expand to thousands of vision tokens, naively concatenating the full screenshot history quickly inflates the context window beyond any reasonable training-time length: a 30-step rollout would saturate even a 64k-token context.
Empirically, the actionable information from earlier screenshots is already distilled into the agent's prior reasoning traces, which themselves remain in context across turns.
We therefore retain only the last $k$ screenshots verbatim to reduce context length substantially.
Appendix~\ref{app:browser:trajectory} shows one example input to VLM for the last turn with $k=1$ context image and the corresponding VLM response.

All interactions are executed within the Orchard Env sandbox: each task runs in an isolated Playwright-controlled Chromium instance (1 vCPU, 8 GiB memory) with task-specific configurations. This isolation not only helps reproducibility for scenarios involving authentication, region-locked content, and rate-limited APIs, but also enables scalable parallel execution, significantly improving training and evaluation throughput.

\subsection{Trajectory Collection and Curation}
\label{sec:swe:data}
We construct the Orchard-GUI dataset through a three-stage pipeline: \emph{(i)} sourcing and filtering raw task intents into a clean seed pool, \emph{(ii)} sampling teacher trajectories on those tasks within the Orchard Env, and \emph{(iii)} judge-based filtering and quality curation to produce the final SFT/RL splits.
Table~\ref{tab:browser-data} summarizes the composition of the collected dataset and the subsets used for SFT/RL.

\begin{figure}[h]
    \centering
    \includegraphics[width=\linewidth]{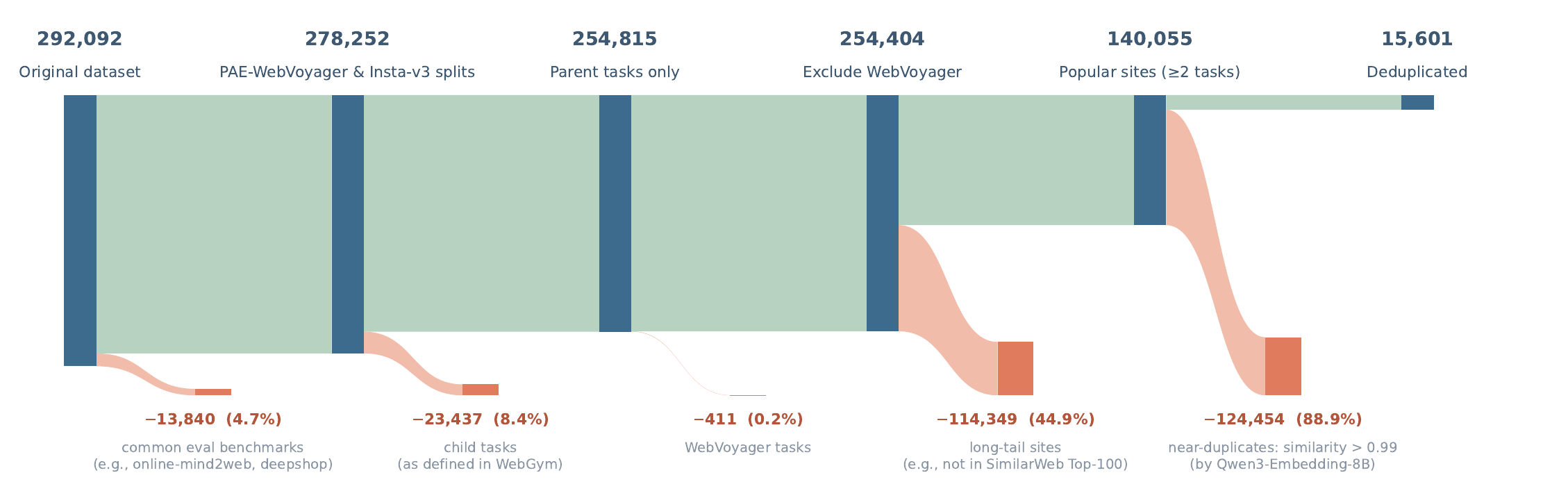}
    \caption{\textbf{Task-filtering pipeline.} Starting from $292{,}092$ raw tasks, we sequentially remove evaluation-benchmark overlap, child tasks, WebVoyager intents, long-tail sites, and near-duplicate intents (semantic similarity $\geq 0.99$ under \texttt{Qwen3-Embedding-8B}), yielding a final pool of $15{,}601$ deduplicated seed tasks on popular websites.}
    \label{fig:gui:task_source}
\end{figure}

\begin{figure}[h]
    \centering
    \begin{minipage}[c]{0.495\linewidth}
        \centering
        \includegraphics[width=\linewidth,keepaspectratio]{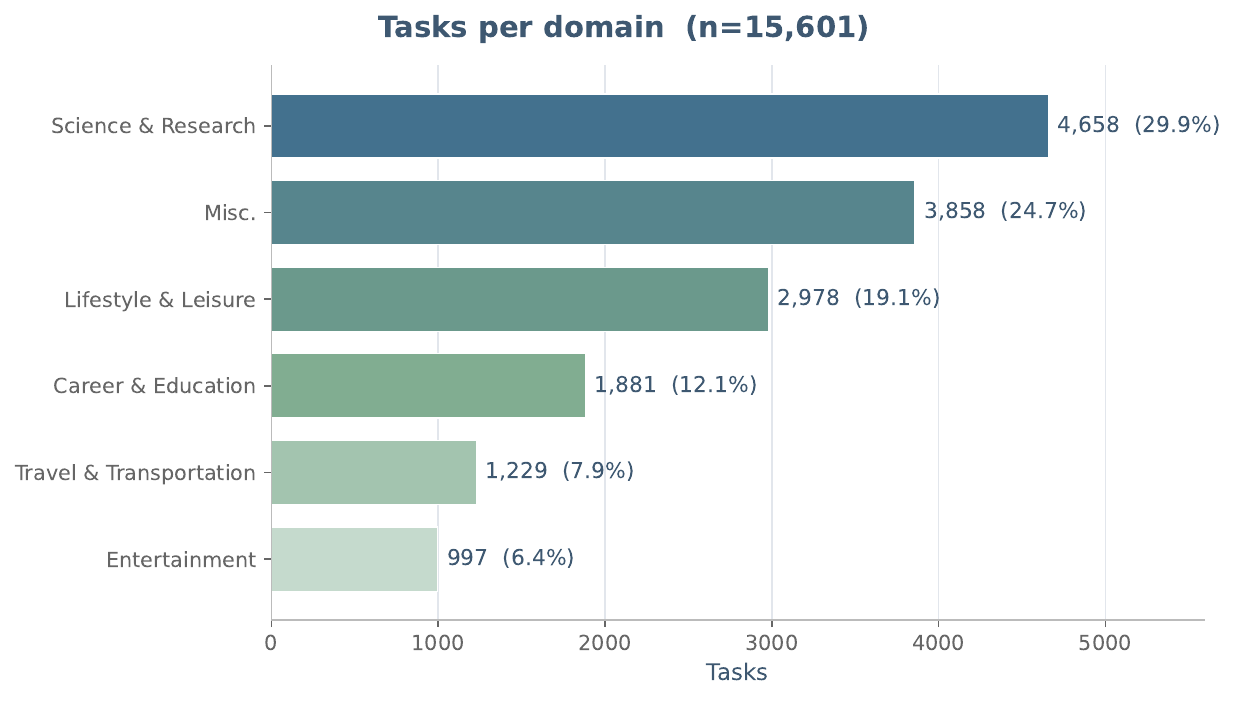}
    \end{minipage}\hfill
    \begin{minipage}[c]{0.495\linewidth}
        \centering
        \includegraphics[width=\linewidth,keepaspectratio]{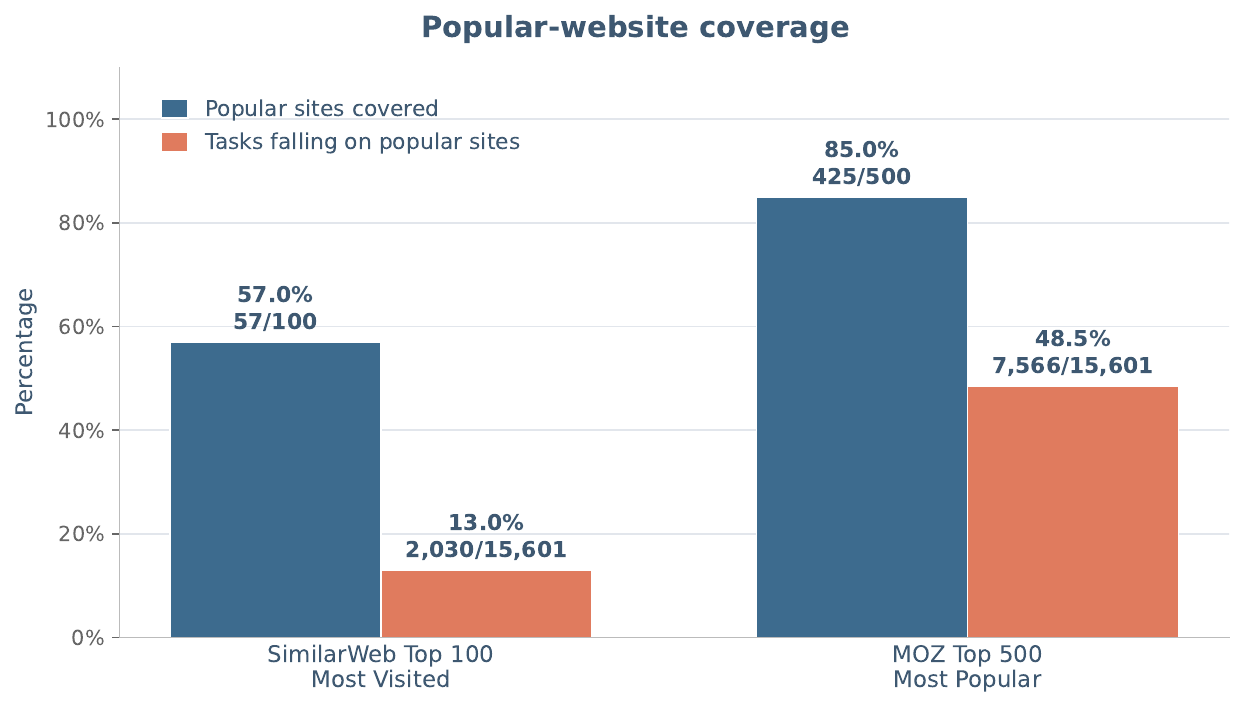}
    \end{minipage}
    \caption{\textbf{Composition of the filtered seed task pool.}
    \emph{Left:} task share by top-level domain ($6$ categories spanning $15{,}601$ tasks).
    \emph{Right:} the seed task pool covers $57.0\%$ of SimilarWeb Top-$100$ Most Visited and $85.0\%$ of MOZ Top-$500$ Most Popular websites, with $13.0\%$ and $48.5\%$ of tasks landing on those respective lists.}
    \label{fig:webgym-rollouts}
\end{figure}

\paragraph{Task sources.}
We draw task instances from the task set provided by WebGym~\citep{bai2026webgym}, which contains 292,092 raw instances in total. To produce a clean, evaluation-safe, and diverse pool of training prompts, we apply a five-step filtering pipeline (Figure~\ref{fig:gui:task_source}). The final filtered pool consists of 15,601 unique task intents, which serve as the seed set for sampling teacher trajectories used in SFT and RL. Of these, 2,537 come from PAE-WebVoyager~\citep{zhou2025pae}, and 13,064 come from InSTA~\citep{Trabucco2025InSTA}.
These tasks span 13,063 unique hosts across six broad domain categories (Figure~\ref{fig:webgym-rollouts}, left). They cover 425/500 (85.0\%) of the MOZ Top-500 websites and 57/100 (57.0\%) of the SimilarWeb Top-100 websites. Correspondingly, 48.5\% (7,566) of tasks fall on a MOZ Top-500 host, and 13.0\% (2,030) on a SimilarWeb Top-100 host (Figure~\ref{fig:webgym-rollouts}, right). More detailed information is provided in Appendix~\ref{app:gui:task_filtering}.
Note that the tasks used in the RL stage are drawn from the same task pool and processed using the same filtering pipeline, but with a more restrictive similarity-based deduplication threshold of 0.95, yielding a task set of 2,198 tasks (Table~\ref{tab:browser-data}).

\paragraph{Trajectory generation.}

\begin{figure}[t]
    \centering
    \begin{minipage}[c]{0.36\linewidth}
        \centering
        \includegraphics[width=\linewidth,keepaspectratio]{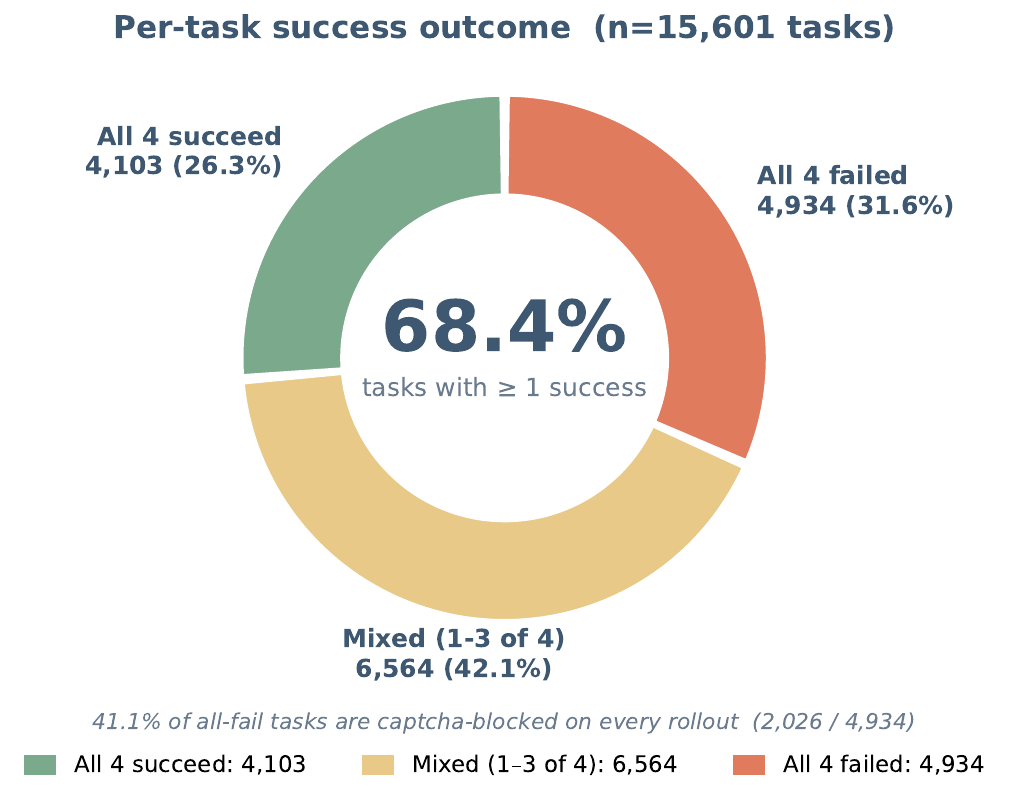}
    \end{minipage}\hfill
    \begin{minipage}[c]{0.63\linewidth}
        \centering
        \includegraphics[width=\linewidth,keepaspectratio]{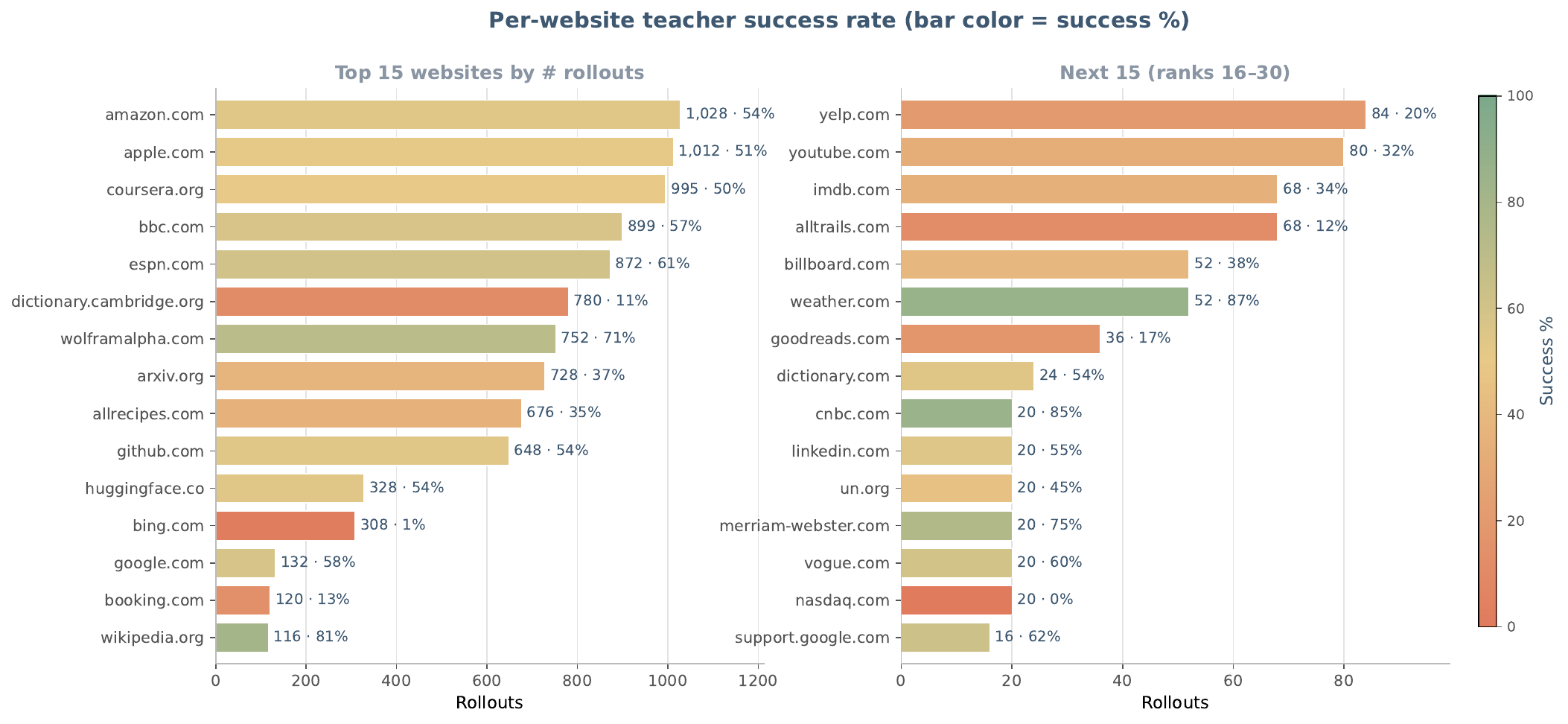}
    \end{minipage}
    \caption{\textbf{Per-website and per-task success outcomes.}
    \emph{Left:} share of tasks by outcome across the $4$ rollouts -- $26.3\%$ all-pass, $42.1\%$ mixed ($1$--$3$ of $4$), and $31.6\%$ all-fail. Of the all-fail tasks, $41.1\%$ ($2{,}026$ tasks; $13.0\%$ of the full $15{,}601$-task pool) are captcha-blocked on every one of their four rollouts.
    \emph{Right:} per-website teacher success rate for the top-$30$ websites by rollout count, split into two columns; bar length = number of rollouts, color = success rate.}
    \label{fig:webgym-site-success}
\end{figure}

We use \texttt{Qwen3-VL-235B-A22B-Thinking}~\citep{qwen2025qwen3vl} as the sole teacher for trajectory distillation.
For each of the $15{,}601$ filtered seed tasks, we sample $4$ independent rollouts through the Orchard Env under the same Tool-Calling Agent Harness described in Section~\ref{sec:gui:setup}, yielding a raw pool of $62{,}395$ teacher rollouts (a small fraction of attempts abort due to environment or rollout engine errors).
GPT-4.1 serves as the judge during data collection: its verdict on the final \texttt{done(response)} and the agent interaction history, together with the screenshot trail is used as a binary reward.
Under this judge, 
$68.4\%$ of tasks have at least one passing rollout, $26.3\%$ pass on all four rollouts, and the remaining $31.6\%$ fail on every rollout (Figure~\ref{fig:webgym-site-success}, left).
A non-trivial share of these failures is environmental rather than agentic: of the $4{,}934$ tasks that fail on every rollout, $41.1\%$ ($2{,}026$ tasks; $13.0\%$ of the full pool) are captcha-blocked on all four attempts, leaving roughly $18.6\%$ of the pool that the teacher genuinely cannot solve.
Per-website success rates also vary widely (Figure~\ref{fig:webgym-site-success}, right), with anti-bot--prone hosts (e.g., \texttt{dictionary.cambridge.org}, \texttt{bing.com}) clustered at the low end.
Sampling four rollouts per task is intentional: the redundancy provides \emph{(i)} pass-rate--based difficulty estimates and trajectory diversity for downstream curation, and \emph{(ii)} a much larger candidate pool than we ultimately train on, letting us study data efficiency by training on small, carefully curated subsets rather than the full collection.

\paragraph{Filtering and curation.}
We first retain only rollouts whose final \texttt{done(response)} is judged \textsc{success} by GPT-4.1 (reward $=1.0$), and split the survivors by source benchmark, yielding 4{,}826 successful PAE-WebVoyager rollouts and 26{,}154 successful InSTA-v3 rollouts.
For SFT we deliberately avoid using the full successful pool: oversaturating the student on imitation data before RL tends to drive it into a narrow imitation regime that on-policy gradients struggle to escape, so we instead select a small, carefully curated subset.
We further restrict the SFT pool to PAE-WebVoyager: although PAE-WebVoyager contributes only $16\%$ of the seed tasks ($2{,}537$ of $15{,}601$), $38.8\%$ of its tasks land on a SimilarWeb Top-100 site, versus just $3.6\%$ for InSTA-v3 (Table~\ref{tab:browser-data})---a $\sim 10\times$ density advantage on popular hosts that more closely reflect everyday user browsing habits.

Within the PAE-WebVoyager success pool we then apply two reductions to balance quality and diversity. \emph{(i) Within-task quality}: for each task we keep a single rollout, namely the \emph{shortest} successful trajectory (fewest turns, with ties broken by total response length), since shorter teacher trajectories tend to be cleaner and contain less recovery noise. \emph{(ii) Across-website diversity}: we cap each website at $K = 20$ tasks, preventing high-volume hosts (e.g.\ \texttt{amazon.com}, \texttt{coursera.org}) from dominating the SFT mix.
The resulting SFT corpus comprises 412 unique tasks spanning 70 websites; per-source breakdowns and outcome statistics for this subset and the RL pool are reported in Table~\ref{tab:browser-data}.

\begin{table}[t]
\centering
\caption{Composition of the Orchard-GUI training dataset. The \emph{Full Set} is the pool of seed tasks for which we collected rollouts. \emph{SFT Trajectories} are the judge-passing rollouts used for supervised fine-tuning, and \emph{RL Tasks} are seed prompts used to bootstrap rollout-based optimization. Tasks are classified by whether \emph{all} rollouts succeeded (All Succ.), \emph{all} failed (All Failed), or the outcome was mixed. The last column reports the number of tasks (restricted to those with at least one successful rollout) whose start URL falls on a SimilarWeb Top-100 site.}
\label{tab:browser-data}
\small
\setlength{\tabcolsep}{6pt}
\begin{tabular}{@{}llrrrrr@{}}
\toprule
\textbf{Subset} & \textbf{Source} & \textbf{\# Tasks} & \textbf{All Succ.} & \textbf{All Failed} & \textbf{Mixed} & \textbf{Tasks on Top-100}\\
\midrule
\multirow{3}{*}{Full Set}
 & PAE-WebVoyager & $2{,}537$  & $587$     & $800$     & $1{,}150$ & $985$ \\
 & InSTA-v3       & $13{,}064$ & $3{,}516$ & $4{,}134$ & $5{,}414$ & $473$ \\
 & \textit{Total} & $\mathbf{15{,}601}$ & $\mathbf{4{,}103}$ & $\mathbf{4{,}934}$ & $\mathbf{6{,}564}$ & $\mathbf{1{,}458}$ \\
\midrule
SFT Traj.\
 & PAE-WebVoyager & $\mathbf{412}$ & $\mathbf{128}$ & $\mathbf{0}$ & $\mathbf{284}$ & $\mathbf{196}$ \\
\midrule
\multirow{3}{*}{RL Tasks}
 & PAE-WebVoyager & $734$     & $189$ & $226$ & $319$ & $290$ \\
 & InSTA-v3       & $1{,}464$ & $500$ & $461$ & $503$ & $469$ \\
 & \textit{Total} & $\mathbf{2{,}198}$ & $\mathbf{689}$ & $\mathbf{687}$ & $\mathbf{822}$ & $\mathbf{759}$ \\
\bottomrule
\end{tabular}
\end{table}
\subsection{Training Recipe}
\label{sec:gui:recipe}

Our training recipe follows a two-stage pipeline: supervised fine-tuning (SFT) on teacher-distilled trajectories, followed by reinforcement learning (RL) with judge-based rewards.
Both stages use the Orchard Env as the execution backend.

\paragraph{Stage 1: Supervised fine-tuning.}
We initialize from \texttt{Qwen3-VL-4B-Thinking}~\citep{qwen2025qwen3vl} and fine-tune on the curated teacher trajectories.
From each teacher rollout we generate one training example per assistant turn: the $t$-th example carries the chat-template-serialized prefix up through turn $t$ and supervises only that turn's assistant response.
The serialized prefix follows the Qwen chat template, with a system turn carrying the agent role and the OpenAI-format tool schema; an initial user turn with the task intent and the \texttt{start\_url} observation; and the subsequent alternation of assistant turns (a \texttt{<think>...</think>} reasoning trace followed by one or more \texttt{<tool\_call>} blocks) and user turns (the tool response wrapping an updated browser observation that includes the latest screenshot).
Following standard practice for long-horizon agent training, the loss is computed only on the final (target) assistant turn; the system prompt, the earlier assistant turns retained as in-context history, and every environment observation are masked out.
The vision encoder and multi-modal projector are kept frozen and only the language-model weights are updated, which preserves the backbone's screenshot-grounding capability and concentrates SFT capacity on agent-specific reasoning and action prediction.
We train for $3$ epochs with peak learning rate $10^{-5}$ under a cosine schedule with a $10\%$ linear warmup. 
Each optimizer step uses a per-device batch of $2$ with $8$-step gradient accumulation, giving a per-worker effective batch of $16$ and a global batch of 128 across 8 data-parallel workers.

\paragraph{Stage 2: Reinforcement learning.}
Starting from the SFT checkpoint, we apply RL to improve the model's ability to recover from errors and explore alternative paths under partial observability.
We optimize a multi-turn variant of GRPO~\citep{shao2024deepseekmathpushinglimitsmathematical}: for each task we sample a group of $G$ trajectories from parallel browser instances, compute a group-relative advantage from the trajectory-level reward, and broadcast it to every assistant-response token across all turns; observation and environment-feedback tokens are masked out of the loss.
The reward combines a deterministic format check with a binary judge: a trajectory receives $+1$ when every assistant turn parses as a valid \texttt{<think>}+tool-call \emph{and} the final \texttt{done(response)} is judged \textsc{success} by GPT-4.1 against the screenshot trail and user intent, $-1$ when the rollout terminates from repeated format failures, and $0$ otherwise.
We use asymmetric PPO clipping ($\epsilon_{\mathrm{low}}{=}0.2$, $\epsilon_{\mathrm{high}}{=}0.28$) without KL or entropy regularization, and intentionally omit the per-trajectory $1/T_i$ loss normalization so that longer, harder tasks are not down-weighted.
To remove uninformative updates we apply DAPO-style trajectory-level dynamic sampling~\citep{yu2025dapo}---dropping groups whose rewards are all $0$ or all $+1$---and additionally zero the loss mask for judge API failures and captcha-aborted runs via the \texttt{remove\_sample} mechanism, so that infrastructure noise does not leak into the policy update.
We further adopt a step-budget curriculum within RL: we first run RL with the per-episode step budget capped at 15 until performance saturates, 
then continue training from that checkpoint with the budget raised to 30. 
The short-horizon phase produces dense reward signal cheaply on tasks the policy can already solve within 15 steps, while the long-horizon phase extends the policy to harder tasks that genuinely require more interaction. 

\begin{figure}[t]
    \centering
    \includegraphics[width=0.8\linewidth,keepaspectratio]{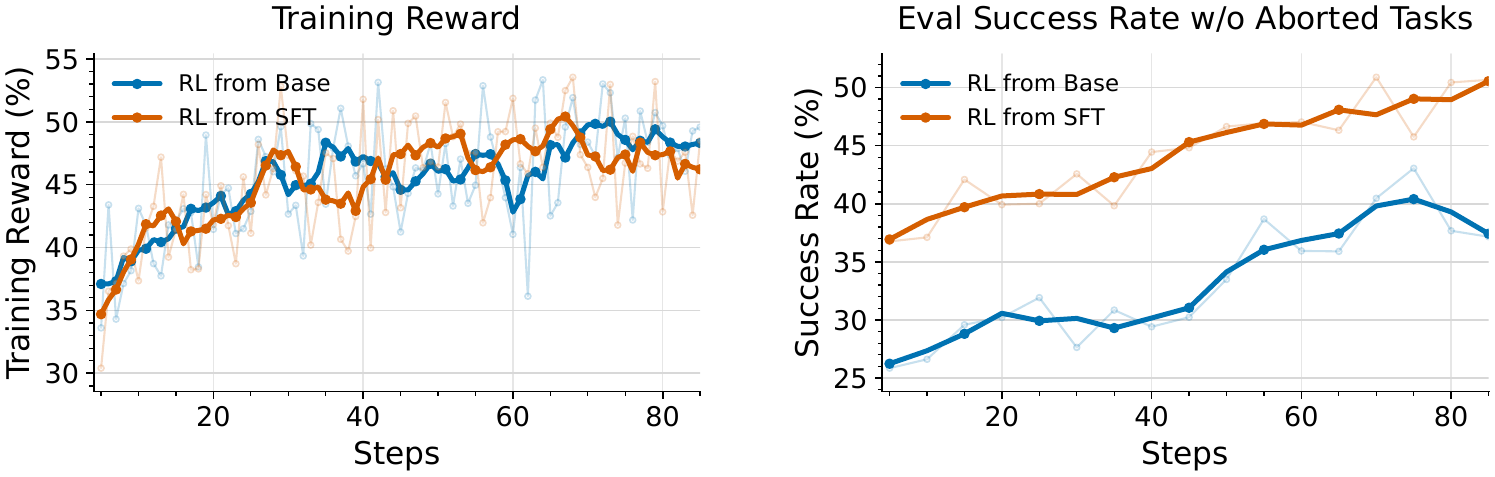}
    \caption{\textbf{Orchard-GUI RL Training and Evaluation Curve.} The red curves denote RL training starting from our SFT checkpoint, while the blue curves denote RL training initialized from the base model. Compared with the base model initialization, the SFT-initialized model achieves consistently higher evaluation success rates and more stable reward improvements throughout training.}
    \label{fig:gui:rl_curvef}
\end{figure}

\subsection{Main Results}
\label{sec:swe:results}
Table~\ref{tab:browser-results} compares Orchard-GUI with proprietary VLMs, prior open-source GUI agents, and same-scale baselines on WebVoyager, Online-Mind2Web, and DeepShop.
After two-stage training, Orchard-GUI reaches 74.1\% / 67.0\% / 64.0\% on WebVoyager / Online-Mind2Web / DeepShop, for a 68.4\% average---the strongest open-source result by a wide margin and competitive with the best proprietary system (Gemini computer-use-preview, 69.3\% avg) despite a 4B backbone and only 2.6k training tasks. RL contributes most of this gain, lifting the SFT checkpoint by +13.9 / +20.0 / +15.3 absolute points across the three benchmarks (52.0\% $\rightarrow$ 68.4\% average).

Four findings stand out.
First, on WebVoyager Orchard-GUI is on par with the strongest open-source baselines (74.1\% vs.\ MolmoWeb-4B's 75.2\% and MolmoWeb-8B's 78.2\%) while consuming roughly two orders of magnitude fewer training tasks (2.6k vs.\ $>$278.5k). WebVoyager covers only 15 popular sites with relatively short horizons, leaving little room to separate from baselines that have been heavily distilled on this exact distribution.

Second, on Online-Mind2Web and DeepShop Orchard-GUI substantially outperforms every previous open-source model---by +31.7 and +21.7 absolute points over MolmoWeb-8B, the strongest prior open baseline---and also surpasses its own 235B Qwen3-VL teacher by +3.3 / +7.3, demonstrating that environment-grounded RL extracts capability the teacher itself does not exhibit.

Third, the training dynamics in Figure~\ref{fig:gui:rl_curvef} show that RL initialized from the SFT checkpoint consistently achieves higher evaluation success rates and more stable optimization behavior than RL initialized directly from the base model. While both settings obtain comparable training rewards, the SFT-initialized policy converges to substantially stronger generalization performance, ultimately reaching over 50\% success on the evaluation set compared with below 40\% for base-model initialization. This gap indicates that supervised initialization provides a crucial behavioral prior that stabilizes exploration and enables RL to more effectively translate reward optimization into downstream task success.

\begin{table*}[t]
\centering
\caption{GUI-agent success rates (\%) across three open-web benchmarks. * marks numbers reported in FARA~\citep{fara7b2025}; $^{\dagger}$ marks numbers reported in MolmoWeb~\citep{gupta2026molmoweb}.}
\label{tab:browser-results}
\small
\setlength{\tabcolsep}{2pt}
\definecolor{avgblue}{RGB}{220,235,255}
\begin{tabular}{lcccccc}
\toprule
\textbf{System} & \textbf{\# Steps} & \textbf{\# Tasks} & \textbf{WebVoyager} & \textbf{Online-M2W} & \textbf{DeepShop} & \cellcolor{avgblue}\textbf{Avg} \\
\midrule
\midrule
\multicolumn{7}{l}{\textit{Proprietary Models}} \\
\midrule
GPT-5 (Axtree)$^{\dagger}$              & 30   & -- & 70.6 & 41.9 & 40.7 & \cellcolor{avgblue}51.1 \\
Gemini-3-flash (Axtree)$^{\dagger}$     & 30   & -- & 74.4 & 34.8 & 45.1 & \cellcolor{avgblue}51.4 \\
Gemini-3-flash (Axtree)$^{\dagger}$    & 100  & -- & 85.6 & 44.8 & 55.3 & \cellcolor{avgblue}61.9 \\
GPT-4o (SoM)*               & 100  & -- & 65.1 & 34.6 & 16.0 & \cellcolor{avgblue}38.6 \\
o3 (SoM)*                   & 100  & -- & 79.3 & 55.4 & 49.7 & \cellcolor{avgblue}61.5 \\
GPT-5 (SoM)*                & 100  & -- & 90.6 & 57.7 & 49.1 & \cellcolor{avgblue}65.8 \\
OpenAI computer-use-preview*      & 100  & -- & 70.9 & 58.3 & 24.7 & \cellcolor{avgblue}51.3 \\
Gemini computer-use-preview$^{\dagger}$       & 100  & -- & 88.6 & 57.3 & 62.0 & \cellcolor{avgblue}69.3 \\
\midrule
\midrule
\multicolumn{7}{l}{\textit{Open-Source Models}} \\
\midrule
Holo1-7B$^{\dagger}$                     & 30   & >15.6k & 55.4 & --  & --  & \cellcolor{avgblue}-- \\
UI-TARS-1.5-7B*              & 100  & -- & 66.4 & 31.3 & 11.6 & \cellcolor{avgblue}36.4 \\
GLM-4.1V-9B-Thinking*        & 100  & -- & 66.8 & 33.9 & 32.0 & \cellcolor{avgblue}44.2 \\
Fara-7B*                     & 100  & >123.2k & 73.5 & 34.1 & 26.2 & \cellcolor{avgblue}44.6 \\
MolmoWeb-4B$^{\dagger}$                  & 100  &  >278.5k & 75.2 & 31.3 & 35.6 & \cellcolor{avgblue}47.4 \\
MolmoWeb-8B$^{\dagger}$                  & 100  &  >278.5K & \textbf{78.2} & 35.3 & 42.3 & \cellcolor{avgblue}51.9 \\
Qwen3-VL-4B-Thinking         & 30  & -- & 49.0 & 32.0 & 33.3 & \cellcolor{avgblue}38.1 \\
Qwen3-VL-235B-A22B-Thinking  & 30  & -- & 63.1 & 63.7 & 56.7 & \cellcolor{avgblue}61.2 \\
\midrule
\textbf{Orchard-GUI-4B-SFT}      & 30  & 0.4k & 60.2 & 47.0          & 48.7          & \cellcolor{avgblue}52.0\\
\textbf{Orchard-GUI-4B (SFT + RL)} & 30  & 2.6k & 74.1 & \textbf{67.0} & \textbf{64.0} & \cellcolor{avgblue}\textbf{68.4} \\
\bottomrule
\end{tabular}
\end{table*}

Finally, the largest gains appear on Online-Mind2Web, which spans a substantially broader and more diverse website distribution than either WebVoyager (15 fixed sites) or DeepShop (a single shopping vertical). Success on this benchmark therefore requires generalization to previously unseen interfaces rather than adaptation to a narrow site set. The fact that Orchard-GUI improves most strongly in this regime suggests that judge-grounded RL over a relatively small but diverse task pool can generalize across the open web more effectively than large-scale teacher distillation on narrow distributions, which is ultimately the practically relevant setting for deployable browser agents.

\section{Orchard-Claw}
\label{sec:claw}
This section presents Orchard-Claw, our instantiation of the Orchard training recipe for claw-based agents.
We describe the problem setting, trajectory collection methods, two-stage training recipe, main results on Claw-Eval \citep{ye2026clawevaltrustworthyevaluationautonomous}, and ablations that isolate key design choices that impacts performance.

\subsection{Problem Setting}
\label{sec:claw:setup}

\paragraph{Task and evaluation} We target multi-step daily workflow tasks formulated by Claw-Eval \citep{ye2026clawevaltrustworthyevaluationautonomous}. Given a task instruction such as ``Sort my inbox — which emails need a reply, which are notifications, and which are spam?'', the agent need to interact with a diverse set of daily tools such as ``gmail\_list\_messages'', ``gmail\_get\_message'', etc., to complete the task while being safe and robust.
Specifically, after the agent completes the task, the evaluation audits the entire agent trajectory to measure the completion, safety, and robustness of the agent using a combination of automated scripts and LLM-as-a-judge \citep{zheng2023judgingllmasajudgemtbenchchatbot,xiong2026multicritbenchmarkingmultimodaljudges}.
These three dimensions are aggregated into a single task score, $\text{task\_score} = \text{safety} \times (0.8 \cdot \text{completion} + 0.2 \cdot \text{robustness})$, and a task is counted as a pass if $\text{task\_score} \geq 0.75$.
We use \textbf{Claw-Eval} as our primary evaluation benchmark.

\paragraph{Agent harness and tool interface}
We collect trajectories using two different agent harnesses: a ReAct-style harness defined by the Claw-Eval benchmark, and the ZeroClaw \citep{zeroclaw2026} harness -- a faster, more lightweight Rust version of the popular OpenClaw \citep{openclaw2026}.
All environment and harnesses are implemented in a docker runtime routed through the Orchard Env service: each task run (environment and harness) runs in an isolated sandbox (2 vCPU, 2 GiB memory), provisioned from a python-based image with ClawEval and ZeroClaw pre-installed.
During both SFT and RL, we train Orchard-Claw on both harnesses and study whether such end-to-end training helps the model better leverage advanced harnesses such as ZeroClaw and reach higher performance.

\subsection{Trajectory Collection and Curation}
\label{sec:claw:data}
As claw-based agents are relatively new, we conduct a preliminary study using Claude Opus 4.6 \citep{anthropic2026opus46} to synthesize claw-agent tasks as our training set.

\paragraph{Task sources.} We draw seed tasks from two sources: (1) tasks from Claw-Eval, and (2) workflows from popular skills on ClawHub\footnote{We use the official OpenClaw CLI to access skills on \url{https://clawhub.ai/}}. From these seeds, we prompt Opus 4.6 via claude-agent-sdk to synthesize new tasks in a four-step loop: (1) propose and filter task ideas; (2) generate the environment, files, tool server, and test script; (3) run MiniMax-M2.5 \citep{minimax2026m25} as the solver to produce rollouts; (4) refine the task based on the rollouts to ensure feasibility and instruction clarity. Each task costs 4.9 USD to synthesize on average, yielding 192 tasks in total shared across the Claw-Eval and ZeroClaw harnesses.

\paragraph{Trajectory generation.} For simplicity, we distill SFT data from a single teacher model. We choose MiniMax-M2.5 for its strong performance. For each synthesized task, we sample five rollouts from MiniMax-M2.5 through Orchard Env under the corresponding harness (ReAct-style or ZeroClaw), and keep only the trajectories that complete the task.
To record training samples from complex harnesses such as ZeroClaw, we implement a proxy LLM server that records every LLM call (input and output) during the rollout.
Appendix~\ref{app:claw:trajectory} shows an example of a recorded input-output pair from the ZeroClaw harness.
Once a rollout finishes, each recorded (input and output) pair is grouped back as a trajectory for training (and also for reward computation during RL).
This yields 561 trajectories with 4537 training pairs in total for SFT.

\subsection{Training Recipe}
\label{sec:claw:training}
Our training recipe follows a two-stage pipeline: supervised fine-tuning (SFT) on teacher-distilled trajectories, followed by reinforcement learning (RL).
Both stages use Orchard Env as the execution backend.
We use Qwen3-30B-A3B-Thinking-2507 \citep{qwen3technicalreport} as the backbone model for our training.

\paragraph{Stage 1: Supervised fine-tuning.} We initialize from the base backbone and fine-tune on the curated teacher trajectories. Each training sample is an (input prompt, LLM response) pair logged by our proxy LLM server, and we mask the input and train only on the response.
We run SFT for 1 epoch with a global batch size of 16 and a 64k context window, using a cosine learning rate decayed from $10^{-5}$ to $10^{-6}$, and apply left truncation to sequences exceeding the context window.
At inference time, we extend the context to the model's maximum of 256k.

\begin{figure}[t]
    \centering
    \begin{minipage}[c]{0.495\linewidth}
        \centering
        \includegraphics[width=\linewidth,keepaspectratio]{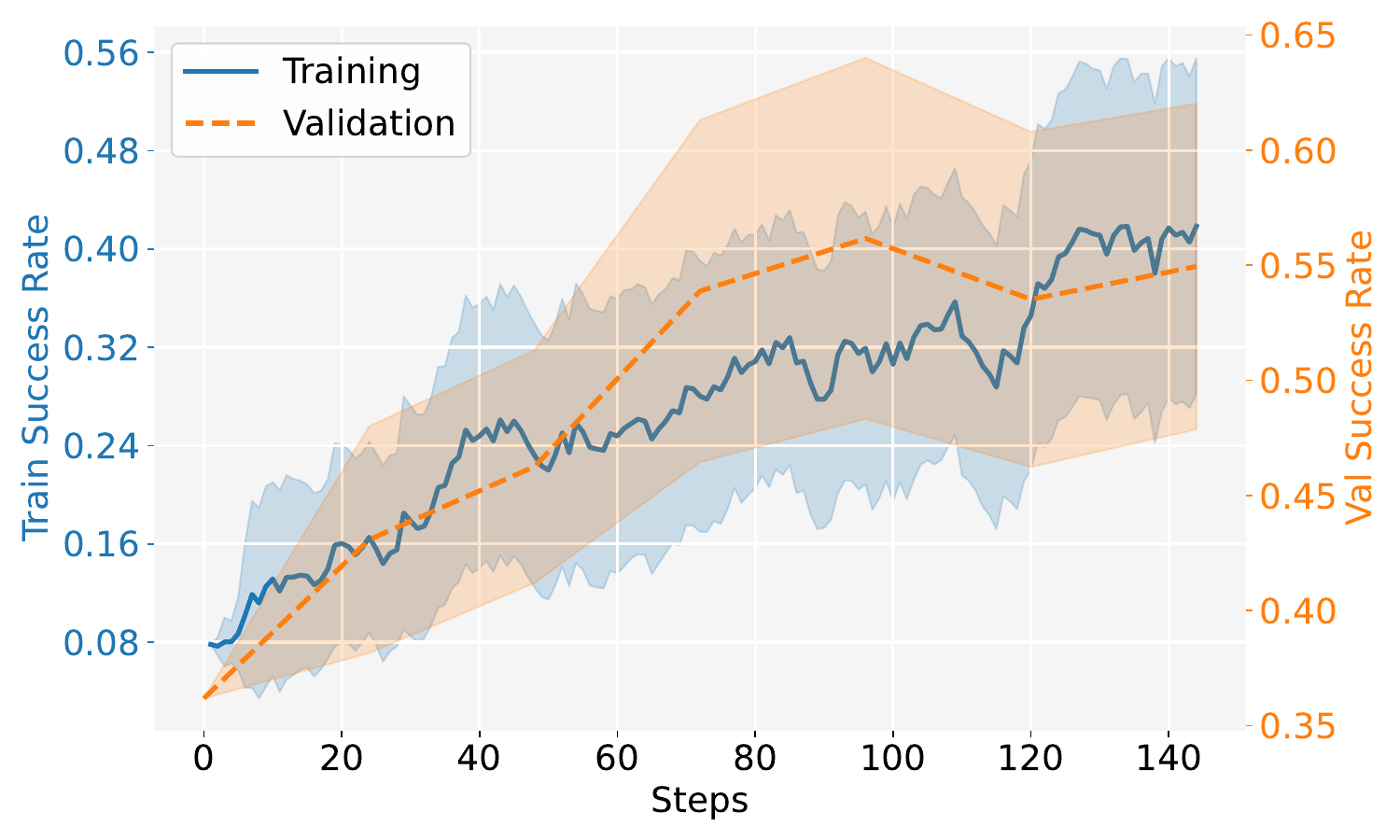}
    \end{minipage}\hfill
    \begin{minipage}[c]{0.495\linewidth}
        \centering
        \includegraphics[width=\linewidth,keepaspectratio]{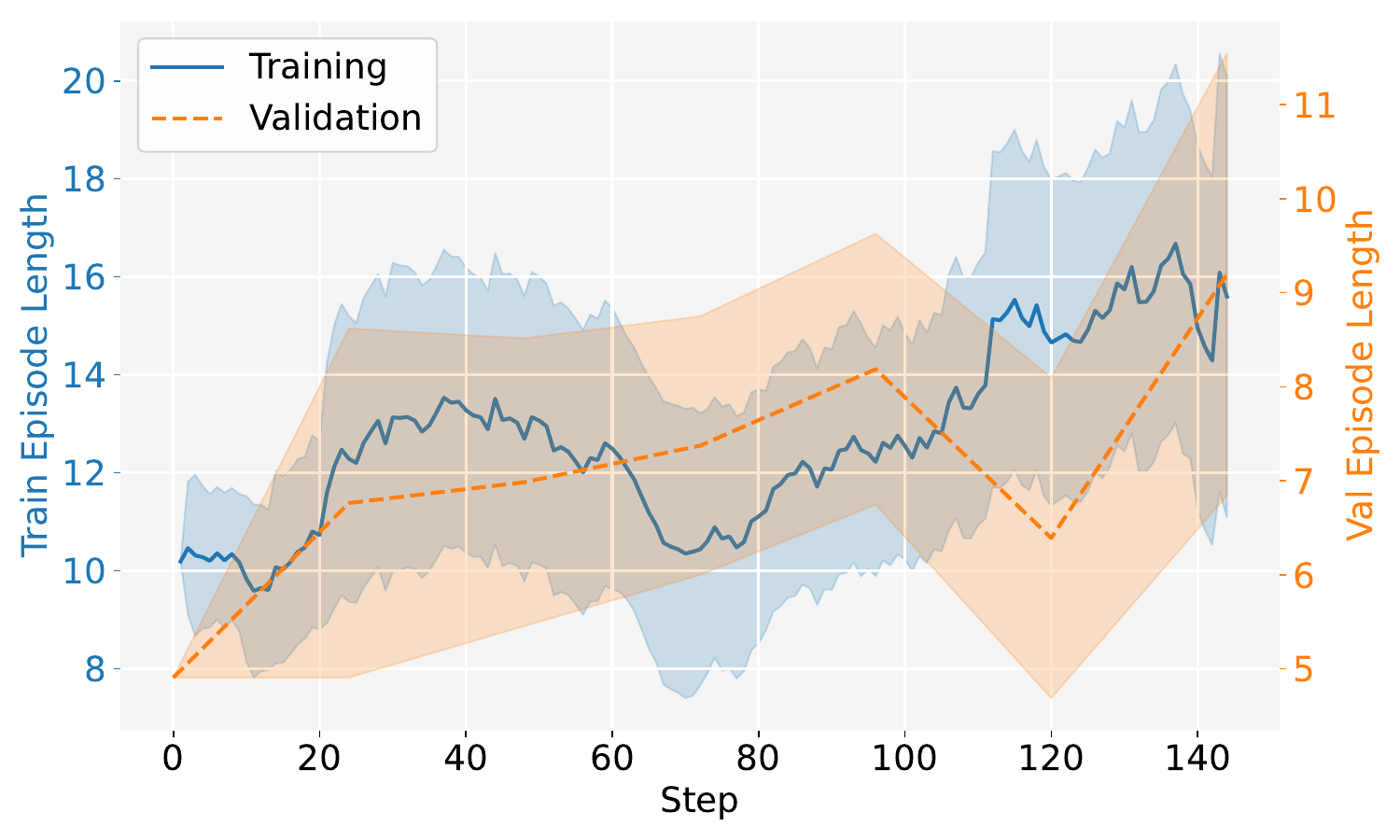}
    \end{minipage}
    \caption{\textbf{Orchard-Claw RL training curves.}
    \emph{Left:} train and validation success rate over RL steps.
    \emph{Right:} train and validation episode length (number of agent turns per rollout).
    Validation tasks are sampled from the ClawEval benchmark.
    Both metrics rise steadily over the course of training, indicating that the agent learns to solve more tasks while also engaging in longer multi-turn interactions.}
    \label{fig:claw-training-curves}
\end{figure}

\paragraph{Stage 2: Reinforcement learning.} Starting from the SFT checkpoint, we apply RL to teach the model to recover from errors and explore alternative paths to task completion.
The reward is binary and environment-grounded: if a rollout passes all test scripts, every (input, output) pair in the rollout receives $+1$; otherwise, every (input, output) pair receives $-1$.
We optimize with standard GRPO \citep{shao2024deepseekmathpushinglimitsmathematical,Guo_2025} using a batch size of 8 and group size of 8 over 150 training steps.
Orchard's sandbox parallelization is critical at this stage, allowing us to easily run 64 asynchronous rollout sandboxes per step, which substantially improves training throughput.
Additionally, during rollout we do not set a maximum step limit but rather a 10-minute wall-clock budget for each task.
We find this better accommodates the differing per-step latencies of different harnesses and different tool calls, and better matches real-world usage.
Rollouts that exceed the budget are aborted, and all of their turns are excluded from training.
In Figure~\ref{fig:claw-training-curves} we plot the training and validation success rate and trajectory length over the course of RL training.

\begin{table*}[t]
\centering
\caption{Claw-agent performance on Claw-Eval. We use the general domain (0408) for evaluation. * marks numbers reported by \citet{ye2026clawevaltrustworthyevaluationautonomous}.}
\label{tab:claw-results}
\small
\setlength{\tabcolsep}{6pt} 
\begin{tabular}{lcccc}
\toprule
\textbf{System} & \textbf{\#Tasks} & \textbf{ClawEval}{($pass^3$)} & \textbf{ClawEval}{($pass@3$)}\\
\midrule
\multicolumn{4}{l}{\textit{SOTA Large Language Models}} \\
\midrule
Claude Opus 4.6*   & -- & 70.8 & 80.8\\
GPT 5.4*          & -- & 60.2 & 75.8\\
Gemini 3.1 Pro*              & -- & 55.9 & 80.8\\
Qwen3.5-397B-A17B*             & -- & 57.8 & 70.8\\
GLM 5 Turbo*      & -- & 52.8 & 73.3\\
MiniMax M2.7*      & -- & 49.7 & 72.0 \\
MiniMax M2.5                 & -- & 47.2 & 65.2\\
Kimi K2.5*                   & -- & 36.6 & 67.1\\
\midrule
\multicolumn{4}{@{}l}{\textit{Similar-size baselines and our model (30B-A3B; \textasciitilde 3B active)}} \\
\midrule
Nemotron-3-nano-30b-a3b        & -- & 26.1 & 57.8 \\
Qwen3-30B-A3B-Thinking        & -- & 14.3 & 39.8 \\
Qwen3-Coder-30B-A3B-Instruct& -- & 30.4 & 49.7 \\
\textbf{Orchard-Claw (SFT)}      & 0.2k & 22.4 & 50.3 \\
\textbf{Orchard-Claw (SFT + RL)} & 0.2k & \textbf{31.7} & \textbf{59.6} \\
\bottomrule
\end{tabular}
\end{table*}

\subsection{Main Results}
\label{sec:claw:results}
Table~\ref{tab:claw-results} compares Orchard-Claw against large proprietary models and open models of similar size on Claw-Eval \citep{ye2026clawevaltrustworthyevaluationautonomous}, using its native ReAct-style harness.
After our two-stage training, Orchard-Claw reaches 31.7\% $pass^3$ and 59.6\% $pass@3$, substantially outperforming its backbone and also surpassing code- and tool-call specialized models such as Qwen3-Coder-30B-A3B-Instruct \citep{cao2026qwen3} and Nemotron-3-nano-30b-a3b \citep{nvidia_nemotron_nano_v3_2025}, despite being trained on only 0.2k synthetic tasks.
RL contributes most of this gain, adding 9.3 absolute points on $pass^3$ and 9.3 absolute points on $pass@3$ over the SFT checkpoint. This suggests that even with limited synthetic data, RL is highly effective at refining agent behavior beyond what teacher distillation alone can offer.


\begin{table}[t]
\centering
\caption{Cross-harness evaluation. ReAct* is the ReAct-style loop from the ClawEval benchmark. ZeroClaw is a lightweight, Rust version of the popular OpenClaw Harness.}
\label{tab:claw-harness}
\small
\setlength{\tabcolsep}{8pt}
\begin{tabular}{@{}lcccc@{}}
\toprule
 & \multicolumn{2}{c}{\textbf{ClawEval($pass^3$)}} & \multicolumn{2}{c}{\textbf{ClawEval($pass@3$)}} \\
\cmidrule(lr){2-3}\cmidrule(lr){4-5}
\textbf{Model} & ReAct* & ZeroClaw & ReAct* & ZeroClaw \\
\midrule
Qwen3-30B-A3B-Thinking
  & 14.3 & 20.5 \tiny{(+6.2)}
  & 39.8 & 44.7 \tiny{(+4.9)} \\

Qwen3-Coder-30B-A3B-Instruct
  & 30.4 & 29.8 \tiny{(-0.6)}
  & 49.7 & 54.7 \tiny{(+5.0)} \\

\textbf{Orchard-Claw (SFT)}
  & 22.4 & 25.5 \tiny{(+3.1)}
  & 50.3 & 62.1 \tiny{(+11.8)}\\

\textbf{Orchard-Claw (SFT+RL)}
    & \textbf{31.7} & \textbf{41.0 \tiny{(+9.3)}}
    & \textbf{59.6} & \textbf{73.9 \tiny{(+14.3)}} \\
\bottomrule
\end{tabular}
\end{table}


In Table~\ref{tab:claw-harness} we further evaluate Orchard-Claw under both its native ReAct-style harness and the more advanced ZeroClaw harness. Pairing Orchard-Claw (SFT+RL) with ZeroClaw lifts performance to 41.0\% $pass^3$ and 73.9\% $pass@3$, a +9.3 and +14.3 absolute improvement over the same model run under the ReAct-style harness. This gain is also the largest among all models in the comparison, including baselines such as Qwen3-Coder-30B-A3B-Instruct that benefit much less or even regress when switching to ZeroClaw.
We attribute this to our end-to-end training using the target harnesses during rollout, enabled by Orchard Env.
By exposing the agent to the harnesses during training, the agent learns to take advantage of the features --- including but not limited to subagents, auto-compact, and more --- that stronger harnesses offer at inference time.
\section{Related Work}
\label{sec:related_work}
\paragraph{Interactive environment orchestration for agentic training.}
Unlike traditional model training, the cornerstone of agentic training is interactive environment orchestration. It requires agents to execute actions, process feedback, and iterate through multi-turn trajectories within isolated sandboxes.
To address the specialized demand on the underlying infrastructure layer, there are two distinct design paradigms emerged: integrated training stacks and decoupled environment services.

In the integrated paradigm, the execution environment is a sub-component embedded within a larger training or orchestration system. 
This allows the co-design of the environment layer with specific training frameworks or agent harnesses, tailoring the infrastructure for a particular task pipeline.
MegaFlow \citep{megaflow2026} decomposes agentic training into three co-designed services (Model, Agent, Environment), having coordinated tens of thousands of concurrent agent tasks. 
While it recognizes that the environment service should be independently scalable, the three services are co-designed for the Qwen training pipeline and are not designed to be composed with arbitrary external trainers or third-party harnesses.
ProRL Agent \citep{prorlagent2026} achieves an important partial step toward decoupling: it separates rollout generation from the trainer via an HTTP service. However, its environment layer remains bound to agent scaffolding through AgentHandler plugins, so the harnesses cannot be swapped without modifying the environment configuration.
In contrast, Orchard Env's REST API is explicitly decoupled from the training loop, the specific task and the agent harness. This modularity makes it uniquely amenable for open-source development and heterogeneous research environments.
By abstracting the environment into a standalone service, our environment orchestration supports the entire agentic development life cycles: trajectory distillation, on-policy rollouts and evaluation.

The second paradigm is decoupled environment services, where the execution environment is exposed as a thin, independent service with a minimal API surface, reusable across different training frameworks, agent scaffolds, and task domains.
Commercial platforms such as E2B \citep{e2b2024} Daytona \citep{daytona2024} and Modal \citep{modal2025} exemplify this approach for developer-facing use cases: they expose REST APIs or SDKs for sandbox lifecycle and code execution. 
However, these platforms generally lack the fine-grained environment controls required for scalable RL training, e.g., tunable resource limits, heartbeat-based lifecycle management tied to training state, per-sandbox network isolation policies. Furthermore, the operational costs of these proprietary services are typically significantly higher than those of Orchard Env, which utilizes a Kubernetes-native design to maximize resource efficiency and minimize overhead.
Consequently, Orchard Env serves as a high-performance, cost-effective foundation specifically engineered for open-source research and agentic development.

\paragraph{Software engineering agents.}
Automated software engineering has converged on a canonical task formulation: given a real GitHub issue and repository snapshot, produce a patch that passes the associated test suite. SWE-bench \citep{jimenez2024swebench} and its human-validated SWE-bench Verified subset \citep{openai2024swebenchverified} operationalize this formulation and serve as the primary evaluation benchmark for Orchard-SWE.
Within this landscape, work has pursued two complementary directions: designing better agent scaffolds and scaling training data. On the scaffold side, SWE-agent \citep{yang2024sweagent} introduces a specialized agent-computer interface enabling structured file viewing, editing, and codebase search.
Its lightweight derivative, mini-swe-agent, serves as one of Orchard-SWE's two training harnesses.
Additionally, OpenHands \citep{wang2025openhands} provides a full-featured multi-agent platform that is used as the second harness for Orchard-SWE.
On the data-scaling side, SWE-smith \citep{yang2025swesmith} generates new training instances from arbitrary repositories via automated task synthesis, scaling task diversity, BugPilot \citep{sonwane2025bugpilot} generates ``unintentional'' bugs by instructing agents to implement new features in a repository.
Orchard-SWE is orthogonal: rather than synthesizing new tasks, we scale trajectory quality through multi-teacher distillation from frontier models and partial-credit supervision on failed traces, drawing from real Github issues \citep{badertdinov2025swe,badertdinov2026swe,zhao2026immersion} as task sources. 

\paragraph{GUI and browser navigation agents.}
GUI agent research is organized around a set of complementary benchmarks that together span the structural diversity of real-world web and desktop tasks.
Mind2Web \citep{deng2023mind2web} introduces the first large-scale dataset of human-annotated cross-website tasks across 137 sites, establishing the dominant web-navigation evaluation suite. OSWorld \citep{xie2024osworld} extends evaluation to full desktop environments with 369 real computer tasks across multi-application workflows, requiring agents to operate over screenshots with no DOM access.
WebVoyager \citep{he2024webvoyager} establishes an end-to-end web task benchmark using live websites with GPT-4V, serving as both a benchmark and a prompting-only baseline.
More recently, Online-Mind2Web revisits the Mind2Web task space in a fully live setting, removing the static snapshot shortcut, and DeepShop \citep{lyu2025deepshop} introduces a transactional e-commerce benchmark requiring multi-step reasoning under shopping constraints — both serving as held-out evaluation targets for Orchard-GUI. We evaluate Orchard-GUI on WebVoyager, Online-Mind2Web, and DeepShop — selected specifically because they represent structurally distinct task types, use live environments rather than static snapshots, and have no overlap in their action spaces or reward signals, making them a demanding testbed for a single unified model without benchmark-specific tuning.

The methodological progression in this field moves from these prompting-only baselines toward sophisticated training-heavy paradigms.
Early systems like WebVoyager \citep{he2024webvoyager} establish strong prompting baselines but leave substantial headroom for trained models.
The subsequent dominant paradigm focuses on supervised fine-tuning (SFT) on human or model-generated demonstrations.
Fara \citep{fara7b2025} introduces FaraGen, a scalable pipeline that proposes multi-step web tasks and filters successes via automatic verifiers to produce low-cost SFT data, yielding a screenshot-only 7B agent competitive with frontier models.
More recently, MolmoWeb \citep{gupta2026molmoweb} assembles MolmoWebMix—a large curated blend of synthetic trajectories, human demonstrations, and atomic web-skill data—to train a fully open agent that achieves state-of-the-art among open-weight models across WebVoyager, Online-Mind2Web, and DeepShop.
A newer wave of research integrates reinforcement learning (RL) to enhance reasoning and out-of-distribution (OOD) performance. UI-TARS \citep{qin2025uitarspioneeringautomatedgui} pioneers an iterative RL data flywheel for GUI agents.
Recent open-source methods \citep{luo2025guir1generalistr1style,lu2025uir1enhancingefficientaction} focus more on improving grounding and reasoning quality through RL.
Despite these advances, cross-benchmark generalization across both online and offline environments under a unified training setup remains rare.
Orchard-GUI trained with Orchard Env as a harness-agnostic execution backend demonstrates that a single SFT+RL recipe applied on a 4B model achieves cross-domain generalization. This provides concrete evidence for the scalability and reusability of the proposed open development system.

\paragraph{Generalist long-running autonomous agents (Claw-agent).}
Claw-style agents represent a shift from episodic domain tools (SWE/GUI) toward persistent, general-purpose partners. While domain agents optimize for specific environments with resetting memory, Claw-agents maintain persistent state and identity via structured artifacts. They leverage dynamic skill libraries (ClawHub), execute multi-step workflows across heterogeneous APIs, and use proactive heartbeats to sustain an ambient presence. This architectural divergence targets conversational alignment over open-ended horizons, making cross-harness generalization a central challenge as the agent's tool surface continually expands.

Several recent benchmarks \citep{ye2026clawevaltrustworthyevaluationautonomous,li2026clawevalliveliveagentbenchmark,bai2026clawgymscalableframeworkbuilding} establish the evaluation standards for Claws-type agents.
Claw-Eval \citep{ye2026clawevaltrustworthyevaluationautonomous} provides high-quality, human-curated scenarios that rigorously assess long-term planning and tool-calling stability. In contrast, ClawGym \citep{bai2026clawgymscalableframeworkbuilding} relies on automated data synthesis over mock workspaces to create a scalable data pipeline for training and evaluation. 
Recent training innovations focus on efficiency and rapid adaptation.
MetaClaw \citep{xia2026metaclawjusttalk} enables continual skill synthesis from failure trajectories and idle-period RL updates, showing gains on Claw benchmarks.
Furthermore, OpenClaw-RL \citep{openclaw2026} treats live deployment signals, such as user feedback, as a continuous training source via Hindsight-Guided On-Policy Distillation. 
Using Orchard-Env, we instantiate an end-to-end training pipeline, Orchard-Claw, to achieve continual improvement on Claw-Eval through a unified, harness-agnostic execution backend.

\section{Conclusion}
\label{sec:conclusion}

This paper presented \textbf{Orchard}, an open-source framework for scalable agentic modeling built around a thin, Kubernetes-native, harness-agnostic environment service. By decoupling sandbox management from agent harnesses, trainers, and task domains, Orchard Env makes trajectory collection, SFT, RL rollouts, and evaluation more reusable, reproducible, and cost-effective.
Across software engineering, GUI navigation, and personal-assistant workflows, Orchard demonstrates that a shared environment layer can support diverse agents and training recipes. Orchard-SWE, Orchard-GUI, and Orchard-Claw achieve strong results while showing improved transfer across harnesses, domains, and pipeline stages.
Overall, Orchard shows that scalable agentic progress depends on both infrastructure and training design. By making the environment service, training recipes, and data collection reusable across domains and harnesses, Orchard lowers the barrier to open, reproducible, and capability-focused research in agentic AI.

\bibliography{dlg-agenticmodeling}
\bibliographystyle{dlg-agenticmodeling}

\clearpage
\appendix
\section{Orchard Env Design Details}
\label{sec:appendix:env_details}

The design of Orchard Env is guided by a central principle: the environment layer should be \emph{thin enough to be reusable} across training recipes, agent harnesses, and model backends, while providing the isolation and lifecycle management needed for large-scale agentic training.
We highlight five design choices that realize this principle and, together, satisfy requirements R1--R3 stated in the section opening.
Two are \emph{distinguishing technical choices} that set Orchard Env apart from existing environment services: \emph{agent injection} addresses image heterogeneity at near-zero adaptation cost (R2), and \emph{direct Pod-IP communication} keeps the service thin and removes the Kubernetes control plane from the hot path (R1).
The remaining three---\emph{network isolation}, \emph{asynchronous lifecycle with heartbeat-based cleanup}, and \emph{watch-based readiness}---are \emph{operational properties} that make the service production-grade at the concurrency and reliability levels required by large-scale agentic training; together with the Kubernetes-native deployment they support R3 (quantified in \S\ref{sec:orchard:comparison}).
The five paragraphs below present the distinguishing choices first, then the operational properties.

\paragraph{Agent Injection via Init Containers.}
A central challenge in building environment services for agentic training is \emph{image heterogeneity}: different tasks require different base images (e.g., specific Python versions, system libraries, or language toolchains), and modifying each image to include an execution agent is impractical at scale.
Orchard Env addresses this through a Kubernetes init container that copies a self-contained Python runtime and agent server into a shared \texttt{emptyDir} volume before the main container starts.
The main container then launches the agent from the shared volume via \texttt{/opt/sandbox-agent/start.sh}.
This design avoids baking Python or the agent into each task image; in practice, Orchard Env targets Linux container images and, by default, launches the injected agent through \texttt{sh -c}.

\paragraph{Direct Pod-IP Communication.}
After a sandbox is provisioned, all execution and file operation requests are routed directly to the pod IP, bypassing the Kubernetes API server entirely.
This avoids the control-plane mediation and WebSocket setup overhead of the Kubernetes \texttt{exec} API.
Direct communication reduces per-command round-trip overhead and removes the API server as a throughput bottleneck under high concurrency.

\paragraph{Network Isolation.}
Orchard Env enforces network isolation through Kubernetes NetworkPolicy resources.
A namespace-wide default-deny egress policy prevents sandbox containers from initiating outbound connections.
When a sandbox requires network access (e.g., for package installation), the orchestrator creates a per-sandbox NetworkPolicy that selectively allows egress, which is cleaned up with the sandbox.
This provides defense-in-depth: even if a user-supplied command attempts to exfiltrate data, it is blocked at the network layer.

\paragraph{Asynchronous Lifecycle with Heartbeat-Based Cleanup.}
Sandbox creation is asynchronous: the API returns immediately after pod creation, and clients poll or block on a \texttt{/wait} endpoint until readiness.
This decouples API responsiveness from Kubernetes scheduling latency.
Long-running sandboxes can be kept alive by periodic heartbeat messages from the client SDK.
A background cleanup loop in the orchestrator detects sandboxes whose heartbeat has expired and deletes them, preventing resource leakage from crashed or abandoned clients.

\paragraph{Watch-Based Readiness.}
Rather than polling the Kubernetes API for pod status, the orchestrator maintains a persistent \texttt{LIST+WATCH} stream that tracks all sandbox pod state transitions in real time.
State changes are cached in memory and waiters are notified via \texttt{asyncio.Event}, avoiding repeated polling of the Kubernetes API.
\section{Cost Analysis Details}
\label{sec:appendix:cost}

This appendix provides the full methodology and discussion for the cost comparison in Table~\ref{tab:cost-comparison}.

\paragraph{Scenario.}
We estimate the cost of running 128 parallel sandbox environments for 240 hours, with each sandbox configured at 2~vCPUs and 8~GiB RAM---matching a typical SWE-bench task environment.

\paragraph{Orchard setup.}
Orchard is deployed on 17 Azure Standard\_D16ads\_v5 instances (16~vCPU, 64~GiB RAM each): 16 nodes host 8 sandboxes each (128 total), and 1 node runs the orchestrator.
Sandbox nodes use \emph{spot instances} (preemptible VMs at ${\sim}$80\% discount, ${\sim}\$0.165$/hr vs.\ $\$0.824$/hr on-demand), which are well-suited for ephemeral sandbox workloads that can tolerate occasional preemption.
The orchestrator node uses standard pay-as-you-go pricing for stability.

\paragraph{Managed service pricing.}
For managed services, we use their official per-second or per-hour pricing for a 2-vCPU, 8-GiB sandbox, including both compute and memory charges:
\begin{itemize}
    \item \textbf{E2B:} vCPU charge of \$0.000014/vCPU/s $\times$ 2 = \$0.000028/s, plus RAM charge of \$0.0000045/GiB/s $\times$ 8 = \$0.000036/s.
    Total: \$0.000064/s = \$0.2304/hr per sandbox.
    \item \textbf{Daytona:} vCPU charge of \$0.0504/vCPU/hr $\times$ 2 = \$0.1008/hr, plus RAM charge of \$0.0162/GiB/hr $\times$ 8 = \$0.1296/hr.
    Total: \$0.2304/hr per sandbox.
    \item \textbf{Modal:} CPU charge of \$0.00003942/physical-core/s $\times$ 1 core (= 2 vCPU) = \$0.00003942/s, plus RAM charge of \$0.00000672/GiB/s $\times$ 8 = \$0.00005376/s.
    Total: \$0.00009318/s = \$0.3354/hr per sandbox.
    Modal Sandbox pricing is non-preemptible by default.
    \item \textbf{MegaFlow:} Estimated based on Alibaba Cloud ecs.c8a.2xlarge instances (8~vCPU, 16~GiB, ${\sim}\$0.15$/hr), one task per instance as described in the original paper. The per-sandbox resource allocation exceeds our 2-vCPU target.
\end{itemize}

\paragraph{Key observations.}
\begin{itemize}
    \item \textbf{Spot-instance economics.}
    Orchard's self-hosted design enables the use of cloud spot instances---preemptible VMs offered at steep discounts (${\sim}\$0.165$/hr vs.\ $\$0.824$/hr on-demand for D16ads\_v5) in exchange for the possibility of short-notice eviction.
    Because sandbox containers are ephemeral and can be recreated on eviction, spot pricing is a natural fit for the sandbox node pool.
    Managed services cannot pass through spot pricing because they control the underlying infrastructure.
    \item \textbf{VM-level multiplexing.}
    Because Orchard packs multiple sandboxes onto each VM (8 per D16ads\_v5 node), the per-sandbox cost benefits from shared overhead.
    MegaFlow's one-task-per-instance model, by contrast, allocates a full VM to each sandbox, leading to higher per-sandbox cost even with comparable cloud pricing.
    \item \textbf{On-demand comparison.}
    Even without spot instances, Orchard's on-demand cost (\$3,362) is roughly half the cost of E2B and Daytona (\$7,078 each) and one-third of Modal (\$10,305).
    The key advantage is that Orchard gives researchers full control over the cluster---they can tune node pools, autoscaling, network policies, and resource limits without depending on a vendor's control plane.
\end{itemize}

These cost differences compound over the course of a research project.
Generating 160K rollout trajectories, running ablation studies, and iterating on training recipes can easily require thousands of hours of environment interaction.
Orchard's self-hosted, spot-friendly design makes such workloads practical for academic research budgets.

\section{Orchard RL Implementation Details}
\label{sec:appendix:swe_rl_details}

Orchard Env's fast execution latency (0.28\,s per command; Section~\ref{sec:orchard:eval}) is critical at this stage, as each RL rollout requires dozens of environment interactions, and training throughput scales directly with sandbox responsiveness.

\paragraph{System Design and Component Orchestration.}
Our RL system builds on slime~\citep{slime_github} post-training framework, extending its Ray-based, Megatron-LM--backed training and SGLang-based inference architecture for asynchronous agentic RL. The system is organized as four loosely coupled services that communicate through Ray actor handles and HTTP endpoints, so that each component can be scaled, replaced, or restarted independently:

\begin{itemize}
\item \textbf{Policy Trainer.} A Megatron-LM-based distributed trainer, sharded with tensor, pipeline, expert, and context parallelism, that owns the trainable parameters and performs the optimization step using advantage-weighted policy gradients.

\item \textbf{Rollout Inference Service.} An SGLang-based inference service, fronted by a request router, that serves the latest policy snapshot. It supports KV-cache reuse, deterministic sampling seeds, and per-token log-probability extraction.

\item \textbf{Sandboxed Execution Service.} A sandbox runtime initialized from Orchard Env. Each agent trajectory is bound to an isolated sandbox in which bash commands and unit-test suites can be executed safely through Orchard Env.

\item \textbf{Agentic Loop Driver.} A per-sample asynchronous coroutine that orchestrates the tool-calling interaction between the inference service and the execution sandbox. At each step, it tokenizes the running message history with the chat template and registered \texttt{bash} tool schema, queries the inference service, parses the structured tool call from the assistant message, executes the call inside the sandbox, and appends the resulting observation as a \texttt{tool} message. The loop terminates when the agent submits a patch, exceeds a step, wall-clock, or token budget, or is aborted due to rare sandbox failures.
\end{itemize}

The orchestration is asynchronous and pipelined. While the trainer is updating weights for rollout $k$, the rollout manager has already dispatched generation for rollout $k{+}1$. A central rollout manager also implements robust handling of partially-failed trajectories so a single sandbox failure cannot crash the optimizer.

Our Agentic Loop Driver is hardened with a multi-layer timeout and retry hierarchy (sandbox creation, LLM inference, observation execution, sandbox shutdown, total reward evaluation) that bounds tail latency without sacrificing end-to-end reliability. When a sandbox crash is attributed to resource exhaustion, CPU and memory allocations are automatically escalated on retry, and small random jitter is injected before sandbox creation to prevent thundering-herd effects when hundreds of concurrent trajectories spin up simultaneously. Loss masking is applied in token space so that gradients flow only through assistant-generated tokens; tool-result tokens are explicitly masked out, which makes multi-turn agentic RL well-defined under a standard LM cross-entropy objective.

\section{Orchard-SWE Prompt List}
\label{app:swe:prompts}

\begin{toolfamily}[listing options={basicstyle=\ttfamily\scriptsize,breaklines=true,breakatwhitespace=true,columns=fullflexible,keepspaces=true,showstringspaces=false}]{Value-Estimation Prompt}
<|im_start|>system
You are an expert software engineer performing credit assignment on a coding agent's trajectory.

This trajectory FAILED -- the agent's final patch did not pass the required tests. Your task is to estimate P(resolve), the probability the trajectory will ultimately succeed, at each step checkpoint. This is retrospective temporal value estimation.

Reason BACKWARD from the known failure:
1. Read the test outcome to understand exactly what went wrong.
2. Identify the critical mistake -- the step(s) where the agent went wrong.
3. Assign probabilities that REFLECT this knowledge. P should rise during correct exploration but MUST DROP at or before the critical mistake and stay low through submission.

Calibration rules:
- Start at a base rate of 0.3-0.5.
- P MUST DROP when the agent makes the critical error.
- P at the final step should be BELOW 0.2 (we know it failed).
- Do NOT give monotonically increasing P.

Output format: a JSON array, one entry per step. Nothing else.
[
  {"step": 0, "p_resolve": 0.40, "reasoning": "initial exploration, base rate"},
  {"step": 1, "p_resolve": 0.50, "reasoning": "found relevant source file"},
  ...
  {"step": K, "p_resolve": 0.25, "reasoning": "edit is incomplete, missed X"},
  ...
  {"step": N, "p_resolve": 0.10, "reasoning": "submitted without fixing the core issue"}
]
<|im_end|>

<|im_start|>user
## Issue Description
[issue text from the original GitHub bug report]

## Outcome (this trajectory FAILED)
Tests that should have been fixed but WERE NOT:
  - FAILED: tests/test_module.py::test_specific_case
  ... and 3 more
Tests that were successfully fixed: 2
Previously passing tests that REGRESSED:
  - REGRESSED: tests/test_other.py::test_unrelated

## Agent Trajectory

[STEP 0]
Thinking: I need to first explore the repository structure...
Action: bash(ls -la)
Result: [directory listing, truncated to 5K chars]

[STEP 1]
Thinking: Let me find the file that defines the buggy class...
Action: bash(grep -rn "class Foo" src/)
Result: src/module.py:42:class Foo: ...

[... additional steps elided ...]

[STEP N-1]
Thinking: I'll submit this patch now.
Action: submit()
Result: Patch submitted.

## Total steps: N
Provide P(resolve) estimates for steps 0 through N-1.
<|im_end|>
\end{toolfamily}

\begin{toolfamily}[listing options={basicstyle=\ttfamily\scriptsize,breaklines=true,breakatwhitespace=true,columns=fullflexible,keepspaces=true,showstringspaces=false}]{Process-Reward Grading Prompt}
<|im_start|>system
You are a senior software engineer acting as a strict, calibrated grader. You will evaluate an AI coding agent's ENTIRE trajectory on a SWE-bench style bug-fix task. The agent runs bash commands in a sandbox to localize and fix a bug, then (ideally) submits a patch.
 
Grade PROCESS QUALITY along seven dimensions, each on a 10-level scale (one of 0.0, 0.1, 0.2, ..., 1.0) -- as much as possible INDEPENDENT of whether tests ultimately passed. R1 scores the patch itself; R2-R7 judge the agent's SELF-VERIFICATION quality (the tests/repros/checks it wrote or ran to convince ITSELF the fix is correct). The dimensions are combined by a weighted average, so score each dimension independently on its own merits, and ALWAYS cite concrete step evidence (commands / outputs). Score ONLY from the trajectory.
 
CORE PRINCIPLE: penalize based on the model's RESPONSE to evidence (ignoring an error, mislabeling test output, submitting anyway), NOT merely because an error string appears in a tool output. This data is outcome-resolved, so most error/traceback strings are benign noise from normal exploration -- do NOT lower a score just because one appears. Only penalize when the agent's own reasoning/action mishandles the evidence.
 
REFERENCE-GUIDED: if a [REFERENCE] block (held-out official tests + gold patch) is provided, use it ONLY to ground self-test relevance / coverage / discriminativeness (R2 / R3 / R5). Do NOT grade exact-match of the edit, and do NOT penalize the agent merely for not running those hidden official tests -- judge the self-tests it ACTUALLY wrote/ran.
 
R1 -- Fix Quality [0.0-1.0]
   Is the final patch minimal, root-cause, and in files the agent actually read? Penalize sprawling/unrelated changes, band-aids, or editing unread files.
   1.0 focused, minimal, root-cause fix; touches the right area.
   0.5 plausible but over-broad or partially off-target.
   0.0 sprawling/malformed/unrelated, or no usable edit.
 
--- SELF-VERIFICATION (R2-R7): judge ONLY the tests / repro scripts / checks the agent wrote or ran to verify its OWN work. You are NOT judging whether the patch is ultimately correct. Be skeptical and evidence-based; quote concrete commands/outputs. ---
 
R2 -- Self-test Relevance [0.0-1.0]
   Did the self-tests / repros reproduce and target the issue's required behavior? A genuine repro that exhibits the bug BEFORE the fix and re-checks it after is the strongest signal ("collected 0 items" / "no tests ran" does NOT count).
   0.0 off-topic or absent; no repro and does not exercise the changed behavior.
   0.5 partially on-target; touches the area but misses the core asserted behavior, or only a weak / post-hoc repro.
   1.0 directly reproduces and checks the exact behavior in the issue (and, if a [REFERENCE] block is given, the behavior its official FAIL_TO_PASS assert).
 
R3 -- Coverage [0.0-1.0]
   Breadth of scenarios / edge cases the self-tests exercise.
   0.0 single happy-path case only.
   0.5 a couple of cases but misses obvious edge cases (empty/None/unicode/error paths).
   1.0 main case plus relevant edge cases and failure modes.
   When [REFERENCE] is given: judge coverage relative to the behaviors the official
   FAIL_TO_PASS / gold test_patch exercise -- i.e. how many of those required scenarios the agent's self-tests actually hit -- not just generic edge cases.
 
R4 -- Rigor [0.0-1.0]
   Strength of the checks.
   0.0 "print-only" success: prints "passed"/"works" with NO assertion, or a script whose exit code cannot fail.
   0.5 some real assertions but mixed with print-only checks or loose comparisons.
   1.0 consistent, precise assertions (exact values / types / structures), like a real test.
 
R5 -- Discriminativeness [0.0-1.0]
   Would a WRONG or incomplete patch fail these self-tests?
   0.0 vacuous; would pass regardless of the patch.
   0.5 would catch gross errors but not subtle / incomplete fixes.
   1.0 tight enough that a wrong / partial patch would clearly fail.
   When [REFERENCE] is given: use the official FAIL_TO_PASS as the bar -- would the agent's self-tests reject a patch that those official tests would reject? Score higher the more the self-tests overlap that discriminating behavior.
 
R6 -- Regression guard [0.0-1.0]
   Did the agent check existing behavior wasn't broken?
   0.0 no attempt to run pre-existing tests / check for regressions.
   0.5 ran some existing tests but narrowly or incidentally.
   1.0 deliberately ran the relevant existing test suite(s) to guard against regressions.
 
R7 -- Verification integrity [0.0-1.0]
   Was the verification real and respected? (Decoupled from R2-R6 test-design quality.)
   0.0 dishonest / broken: a test command silently no-op'd, OR a real failure (FAILED / AssertionError) was IGNORED and the agent submitted anyway, OR submitted with no verification at all.
   0.5 verification ran but with caveats (failures partially addressed, or ambiguous).
   1.0 tests genuinely ran; failures were acted upon; submitted only after green checks.
 
Think first, then score. Output ONLY a JSON object, no other text, with keys in
THIS order:
{
  "reasoning": "<=3 short sentences citing concrete evidence (with step indices)",
  "scores": {"R1": <v>, "R2": <v>, "R3": <v>, "R4": <v>, "R5": <v>, "R6": <v>, "R7": <v>}
}
<|im_end|>
<|im_start|>user
[TASK / ISSUE]
[issue text from the original GitHub bug report]
 
[REFERENCE -- held-out official tests + gold patch, for grounding only]
Use this ONLY to ground self-test RELEVANCE / COVERAGE / DISCRIMINATIVENESS (R2 / R3 / R5). Do NOT grade exact-match of the agent's edit, and do NOT penalize it for not running these hidden tests.
official_fail_to_pass: tests/test_module.py::test_specific_case, ...
gold_changed_files: src/module.py
gold_test_files: tests/test_module.py
gold_patch:
[gold source-code diff, truncated]
 
[TRAJECTORY] exit_status=Submitted, steps=N
--- STEP 0 ---
reasoning: ...
command: ...
observation (returncode=0): [collapsed to a size marker]
--- STEP 1 ---
reasoning: ...
command: ...
observation (returncode=0): [collapsed to a size marker]
[... additional steps, observations compressed head-and-tail ...]
--- STEP N-1 ---
reasoning: ...
command: submit()
observation (returncode=0): Patch submitted.
 
[FINAL SUBMITTED PATCH]
[final unified diff]
 
Grade THIS trajectory now. Output only the JSON object.
<|im_end|>
\end{toolfamily}
\section{Orchard-SWE RPR Rubrics and Curves}

\begin{table}[h]
\centering
\scriptsize
\caption{Process reward rubric. R1 grades the patch; R2--R7 grade
self-verification quality (reference-guided). Weights sum to $1$, with $0.85$ of the
mass placed on self-verification (R2--R7).}
\label{tab:rpr-rubric}
\begin{tabular}{@{}llp{9.5cm}c@{}}
\toprule
ID & Dimension & What it measures & $w_d$ \\
\midrule
R1 & Fix Quality & Is the final patch minimal, root-cause, and confined to files the agent actually read (rather than a sprawling or band-aid edit)? & 0.15 \\[2pt]
R2 & Self-test Relevance & Do the self-tests / reproductions exhibit the bug before the fix and re-check the issue behavior after? & 0.18 \\[2pt]
R3 & Coverage & Breadth of scenarios and edge cases the self-tests exercise, beyond a single happy path. & 0.12 \\[2pt]
R4 & Rigor & Strength of the checks: precise assertions on values, types, and structures versus ``print-only'' success. & 0.15 \\[2pt]
R5 & Discriminativeness & Would a \emph{wrong} or incomplete patch actually fail these self-tests, or are they vacuous? & 0.14 \\[2pt]
R6 & Regression guard & Did the agent run pre-existing tests to confirm that existing behavior was not broken? & 0.12 \\[2pt]
R7 & Verification integrity & Was verification real and respected---failures acted upon rather than masked or ignored before submission? & 0.14 \\
\midrule
\multicolumn{3}{@{}l}{Total} & 1.00 \\
\bottomrule
\end{tabular}
\end{table}

\label{app:swe:prm_curves}

\begin{figure*}[h]
  \centering
  \includegraphics[width=\textwidth]{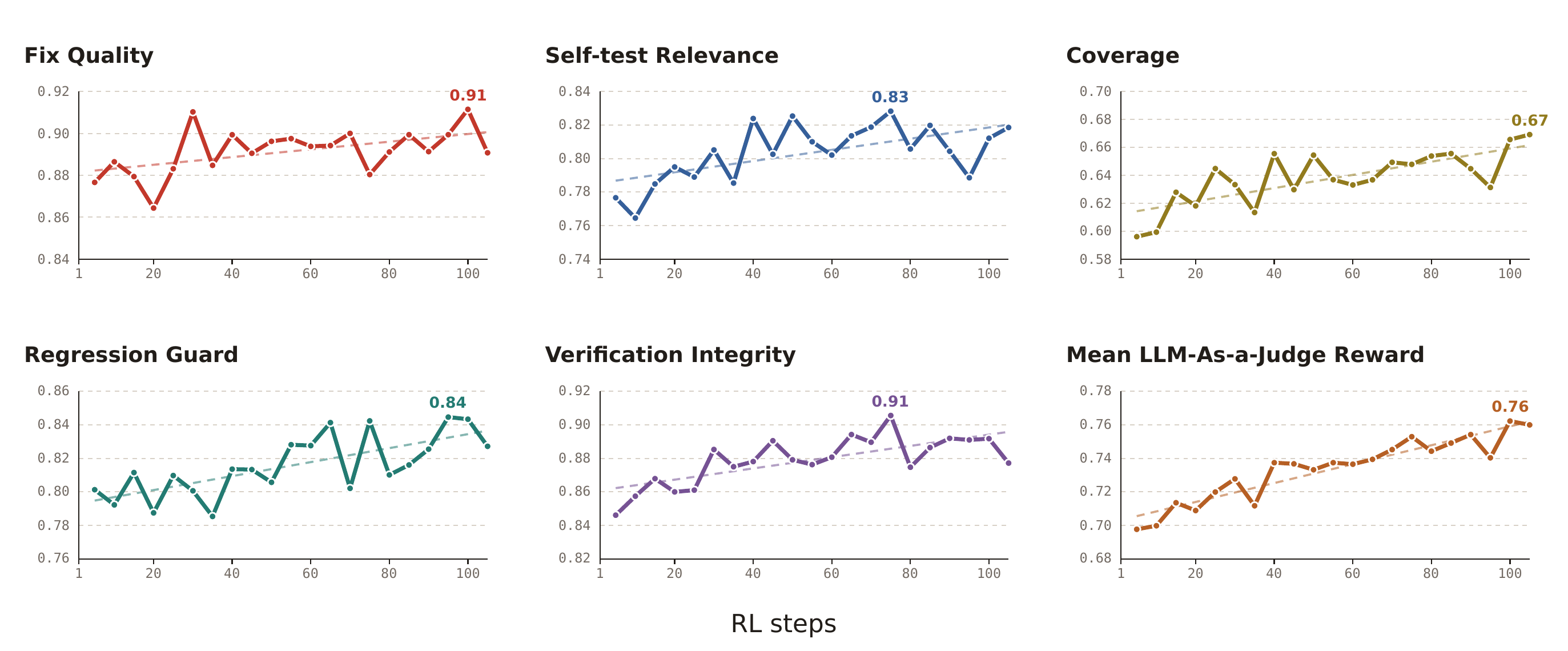}
  \caption{LLM-as-a-Judge process rewards over RL training: fix quality, self-test relevance, coverage, regression guard,
  verification integrity, and their mean. All metrics trend upward with
  training.}
  \label{fig:llm-judge}
\end{figure*}
\section{Orchard-GUI Tool List}
\label{app:browser:tools}

This appendix reproduces the full OpenAI tool-call \emph{JSON Schema} for all 13 atomic tools used by Orchard-GUI.
At each step the agent emits one or multiple tool calls with each inside a \texttt{<tool\_call>}\ldots\texttt{</tool\_call>} block\.
For readability, the schemas are grouped by family, one styled box per family.


\begin{toolfamily}[listing options={basicstyle=\ttfamily\scriptsize,breaklines=true,breakatwhitespace=true,columns=fullflexible,keepspaces=true,showstringspaces=false}]{Pointer Actions: click / hover / drag}
{
    "type": "function",
    "function": {
        "name": "click",
        "description": "Performs a mouse click at specific screen coordinates.",
        "parameters": {
            "type": "object",
            "properties": {
                "point_2d": {
                    "type": "array",
                    "items": {"type": "number"},
                    "minItems": 2, "maxItems": 2,
                    "description": "The [x, y] pixel coordinates relative to the viewport's top-left corner."
                },
                "clicks": {
                    "type": "integer",
                    "description": "The number of consecutive clicks to perform (e.g., 1 for a single click, 2 for a double click).",
                    "default": 1
                },
                "button": {
                    "type": "string",
                    "enum": ["left", "right", "middle"],
                    "description": "The mouse button to trigger.",
                    "default": "left"
                }
            },
            "required": ["point_2d"]
        }
    }
},
{
    "type": "function",
    "function": {
        "name": "hover",
        "description": "Move the mouse cursor to a specific coordinate. Use this to trigger hover states, reveal tooltips, or open dropdown menus.",
        "parameters": {
            "type": "object",
            "properties": {
                "point_2d": {
                    "type": "array",
                    "items": {"type": "integer"},
                    "minItems": 2, "maxItems": 2,
                    "description": "The [x, y] pixel coordinates to move the cursor to."
                }
            },
            "required": ["point_2d"]
        }
    }
},
{
    "type": "function",
    "function": {
        "name": "drag",
        "description": "Perform a drag-and-drop operation from a starting screen coordinate to an ending coordinate.",
        "parameters": {
            "type": "object",
            "properties": {
                "start_point_2d": {
                    "type": "array",
                    "items": {"type": "number"},
                    "minItems": 2, "maxItems": 2,
                    "description": "The [x, y] coordinates to start dragging from."
                },
                "end_point_2d": {
                    "type": "array",
                    "items": {"type": "number"},
                    "minItems": 2, "maxItems": 2,
                    "description": "The [x, y] coordinates to drop at."
                },
                "steps": {
                    "type": "integer",
                    "description": "The number of intermediate mouse steps to take. Higher values make the drag slower and more human-like. Use 10-20 for standard drags.",
                    "default": 10
                }
            },
            "required": ["start_point_2d", "end_point_2d"]
        }
    }
}
\end{toolfamily}

\begin{toolfamily}[listing options={basicstyle=\ttfamily\scriptsize,breaklines=true,breakatwhitespace=true,columns=fullflexible,keepspaces=true,showstringspaces=false}]{Keyboard Actions: write / press\_keys}
{
    "type": "function",
    "function": {
        "name": "write",
        "description": "Type text using the keyboard. This will first clear any existing content in the focused input field (using Ctrl+A and Backspace), then type the new text. Make sure to click on an input field before using this action.",
        "parameters": {
            "type": "object",
            "properties": {
                "message": {
                    "type": "string",
                    "description": "The text content to type into the focused input field"
                }
            },
            "required": ["message"]
        }
    }
},
{
    "type": "function",
    "function": {
        "name": "press_keys",
        "description": "Press specific keys or key combinations on the keyboard. Use this for navigation, shortcuts, or discrete key presses.",
        "parameters": {
            "type": "object",
            "properties": {
                "keys": {
                    "type": "array",
                    "items": {"type": "string"},
                    "description": "The list of specific keys to press.\nACCEPTED KEYS:\n- Modifiers: \"Alt\", \"Control\", \"Meta\", \"Shift\"\n- Actions: \"Enter\", \"Tab\", \"Space\", \"Backspace\", \"Delete\", \"Escape\"\n- Navigation: \"ArrowUp\", \"ArrowDown\", \"ArrowLeft\", \"ArrowRight\", \"Home\", \"End\", \"PageUp\", \"PageDown\"\n- Standard: \"a\"-\"z\", \"0\"-\"9\", \"F1\"-\"F12\""
                },
                "is_hotkey": {
                    "type": "boolean",
                    "description": "If true, all keys in the list are pressed simultaneously (e.g., [\"Control\", \"c\"]). If false, keys are pressed one by one in sequence.",
                    "default": false
                },
                "presses": {
                    "type": "integer",
                    "description": "Number of times to repeat the press action.",
                    "default": 1
                }
            },
            "required": ["keys"]
        }
    }
}
\end{toolfamily}

\begin{toolfamily}[listing options={basicstyle=\ttfamily\scriptsize,breaklines=true,breakatwhitespace=true,columns=fullflexible,keepspaces=true,showstringspaces=false}]{Page-Navigation Actions: scroll / goto\_url / go\_back / wait}
{
    "type": "function",
    "function": {
        "name": "scroll",
        "description": "Scroll the page or a specific scrollable element in the given direction. If point_2d is provided, the scroll occurs on the element at that coordinate (useful for scrollable sub-containers like sidebars, modals, or chat panels). Otherwise, scrolls the main page.",
        "parameters": {
            "type": "object",
            "properties": {
                "direction": {
                    "type": "string",
                    "enum": ["up", "down", "left", "right"],
                    "description": "Direction to scroll.",
                    "default": "down"
                },
                "amount": {
                    "type": "number",
                    "minimum": 0.0, "maximum": 1.0,
                    "description": "Scroll amount as a percentage of the viewport height (0.0 to 1.0). For example, 0.5 means scroll by half a page.",
                    "default": 0.5
                },
                "point_2d": {
                    "type": "array",
                    "items": {"type": "number"},
                    "minItems": 2, "maxItems": 2,
                    "description": "Optional [x, y] pixel coordinates of the element to scroll within. If omitted, scrolls the main page."
                }
            }
        }
    }
},
{
    "type": "function",
    "function": {
        "name": "goto_url",
        "description": "Navigate the current open tab to a specific URL. Use this to open a new webpage directly in the current tab instead of clicking through links or typing in the address bar.",
        "parameters": {
            "type": "object",
            "properties": {
                "url": {
                    "type": "string",
                    "description": "The full URL to navigate to (e.g., \"https://www.example.com\")."
                }
            },
            "required": ["url"]
        }
    }
},
{
    "type": "function",
    "function": {
        "name": "go_back",
        "description": "Navigate back to the previous page in the browser history. Equivalent to clicking the browser's back button.",
        "parameters": {"type": "object", "properties": {}}
    }
},
{
    "type": "function",
    "function": {
        "name": "wait",
        "description": "Wait for a specified number of seconds before continuing. Use this when you need to allow time for page content to load or for animations to complete.",
        "parameters": {
            "type": "object",
            "properties": {
                "seconds": {
                    "type": "integer",
                    "description": "Number of seconds to wait (integer value)",
                    "default": 3
                }
            }
        }
    }
}
\end{toolfamily}

\begin{toolfamily}[listing options={basicstyle=\ttfamily\scriptsize,breaklines=true,breakatwhitespace=true,columns=fullflexible,keepspaces=true,showstringspaces=false}]{Tab-Management Actions: new\_tab / switch\_tab / close\_tab}
{
    "type": "function",
    "function": {
        "name": "new_tab",
        "description": "Open a new blank browser tab. Use goto_url after switching to the new tab to navigate to a specific URL.",
        "parameters": {"type": "object", "properties": {}}
    }
},
{
    "type": "function",
    "function": {
        "name": "switch_tab",
        "description": "Switch to a different browser tab by its index (0-based). Use this to navigate between multiple open tabs.",
        "parameters": {
            "type": "object",
            "properties": {
                "tab_index": {
                    "type": "integer",
                    "description": "The 0-based index of the tab to switch to."
                }
            },
            "required": ["tab_index"]
        }
    }
},
{
    "type": "function",
    "function": {
        "name": "close_tab",
        "description": "Close the current browser tab. If there are multiple tabs open, the browser will switch to the nearest remaining tab.",
        "parameters": {"type": "object", "properties": {}}
    }
}
\end{toolfamily}

\begin{toolfamily}[listing options={basicstyle=\ttfamily\scriptsize,breaklines=true,breakatwhitespace=true,columns=fullflexible,keepspaces=true,showstringspaces=false}]{Termination Action: done}
{
    "type": "function",
    "function": {
        "name": "done",
        "description": "Signal task completion and provide the final answer. Call this function when you have successfully completed the task, when you have determined that the task cannot be completed, or when you encounter a CAPTCHA or anti-bot verification page that blocks further progress. Provide a clear and concise response summarizing what was accomplished or explaining why the task could not be completed (e.g., blocked by CAPTCHA).",
        "parameters": {
            "type": "object",
            "properties": {
                "response": {
                    "type": "string",
                    "description": "The final answer, result, summary of the task, or reason for task failure. Be specific and include relevant details such as: the information found, actions completed, or reason for task failure."
                }
            },
            "required": ["response"]
        }
    }
}
\end{toolfamily}

\section{Example GUI Agent Trajectory}
\label{app:browser:trajectory}

This appendix shows a representative multi-turn rollout produced by Orchard-GUI on a WebVoyager-style task.
The first box gives the prompt fed to the model at the final step (system prompt $+$ accumulated trajectory through prior steps).
The second box shows the model's response at that step (its final reasoning followed by the terminating \texttt{done} call).
The full trajectory of seven \emph{think $\rightarrow$ tool\_call $\rightarrow$ tool\_response} cycles is shown verbatim. Only the JSON tool schema inside the system prompt is abbreviated with \texttt{[\ldots]} (the full schema appears in Appendix~\ref{app:browser:tools}).

\begin{toolfamily}[listing options={basicstyle=\ttfamily\scriptsize,breaklines=true,breakatwhitespace=true,columns=fullflexible,keepspaces=true,showstringspaces=false}]{Inputs to LLM -- full trajectory  (last turn, 1 context image)}
<|im_start|>system
You are a GUI agent designed to operate in an iterative loop to automate browser tasks.

# GUI Agent Policy

As an autonomous GUI agent operating on the **Web Browser** platform, your primary function is to analyze screen captures and perform appropriate UI actions to complete assigned tasks.

## Core Responsibilities

You can perform web browser interactions including:
- **Mouse interactions** - click, double-click, right-click, hover, drag, and scroll (page or element)
- **Keyboard interactions** - type text, press keys, and execute keyboard shortcuts
- **Navigation** - go to a URL, go back in browser history
- **Tab management** - open, switch, and close browser tabs
- **Task completion** - provide responses to queries and terminate tasks with status
- **Waiting** - allow time for UI changes to occur

## Input Information

At each step, you will receive the following information:

1. **Action History**: Your interaction history showing all previous actions taken to accomplish the current task. This helps you track progress and avoid repeating actions.
2. **User Request**: The primary objective that clearly specifies the task you need to complete. This is your main goal.
3. **Observation**: Current state information about the web page, including:
   - **Tab Info**: The currently active tab index and a list of all open tabs with their index, URL, and page title
   - **Screenshot**: Visual representation of the current page state
   - **A11y Tree** *(optional)*: Accessibility tree containing interactive elements with their IDs, types, labels, and positions

## Output Requirements

- Your output must include two tags: one `<think>` and one or more `<tool_call>` blocks.
- Always start your response with `<think>`.

## Guidelines

- **Reasoning process**: In your `<think>` block, you should analyze the current state (e.g., what do you see on the screen), reflect on your previous actions (e.g., did they produce the expected result, or did something go wrong), assess progress toward the goal, plan your next steps, and validate that your planned actions are safe and correct. If you notice you've been repeating the same action without progress, consider an alternative approach.
- **Valid and executable tool calls only**: All tool calls must exist within the defined tool set, and must be valid, executable tool calls.
- **Use multiple tool calls when appropriate**: If a task step naturally involves a short chain of actions on the current page (e.g., "click -> write -> press Enter" or "new_tab -> goto_url"), emit them all in one response with multiple `<tool_call>` blocks -- one per action, each on its own line.
- **Sequential execution**: When using multiple `<tool_call>` blocks, they are executed in order from top to bottom. Ensure the sequence is logically correct -- later actions may depend on earlier ones completing successfully.

# Tools

You may call one or more functions to assist with the user query.

You are provided with function signatures within <tools></tools> XML tags:
<tools>
{"type": "function", "function": {"name": "click", "description": "Performs a mouse click at specific screen coordinates.", "parameters": {...}}}
{"type": "function", "function": {"name": "write", "description": "Type text using the keyboard. ...", "parameters": {...}}}

[... 11 additional tool schemas omitted; see Appendix~\ref{app:browser:tools} for the full list ...]

</tools>

For each function call, return a json object with function name and arguments within <tool_call></tool_call> XML tags:
<tool_call>
{"name": <function-name>, "arguments": <args-json-object>}
</tool_call>
<|im_end|>

<|im_start|>user
<user_request>
Find a dog bed on Amazon that is washable and has a length of at least 30 inches.
</user_request>
<observation>
screen size: 1280 x 1000
Current tab: 0
Available tabs:
- Tab 0 (active): https://www.amazon.com/ - Amazon.com
</observation>
<|im_end|>

<|im_start|>assistant
<think>
The current screen shows an Amazon verification page that requires clicking a button to continue shopping. This is a standard anti-bot measure. I need to click the "Continue shopping" button to proceed to the main Amazon page where I can search for dog beds.

My plan:
1. Click the "Continue shopping" button to get past this verification page
2. Once on the main Amazon page, search for "dog bed washable 30 inches"
3. Apply filters for washable beds with length >= 30 inches
4. Review results and find a suitable product

The next logical action is to click the "Continue shopping" button to proceed with the task. This is necessary before I can access the search functionality of Amazon.
</think>

<tool_call>
{"name": "click", "arguments": {"point_2d": [501, 213]}}
</tool_call>
<|im_end|>

<|im_start|>user
<tool_response>
Succeed: `click` on <button> type=submit "Continue shopping" at (641, 213) executed. Note: no visible navigation or new tab detected.
<observation>
screen size: 1280 x 1000
Current tab: 0
Available tabs:
- Tab 0 (active): https://www.amazon.com/ - Amazon.com. Spend less. Smile more.
</observation>
</tool_response>
<|im_end|>

<|im_start|>assistant
<think>
The previous action successfully got me past the Amazon verification page, and now I'm on the main Amazon homepage. I can see the search bar at the top of the page which I need to use to search for dog beds.

My plan remains:
1. Search for "dog bed washable 30 inches" in the Amazon search bar
2. Apply filters for washable beds with length >= 30 inches
3. Review results and find a suitable product

The next logical step is to click on the search bar so I can type my search query. The search bar is clearly visible at the top of the page with the placeholder text "Search Amazon". Once I click on it, I'll be able to type my search terms for dog beds that are washable and at least 30 inches long.

I should click on the search bar to focus it before typing my search query.
</think>

<tool_call>
{"name": "click", "arguments": {"point_2d": [385, 31]}}
</tool_call>
<|im_end|>

<|im_start|>user
<tool_response>
Succeed: `click` on <input> type=text role=searchbox "Search Amazon" at (493, 31) executed. Note: no visible navigation or new tab detected.
<observation>
screen size: 1280 x 1000
Current tab: 0
Available tabs:
- Tab 0 (active): https://www.amazon.com/ - Amazon.com. Spend less. Smile more.
</observation>
</tool_response>
<|im_end|>

<|im_start|>assistant
<think>
The previous action successfully clicked on the Amazon search bar, and I can see it's now focused and ready for text input. The cursor is blinking in the search field, indicating it's ready to receive my search query.

My plan is still:
1. Search for "dog bed washable 30 inches" in the Amazon search bar
2. Apply filters for washable beds with length >= 30 inches
3. Review results and find a suitable product

Now that the search bar is focused, the next logical step is to type my search query. I should enter "dog bed washable 30 inches" to find products that match all the criteria specified in the user request. This search term includes all the key requirements:
- It's for a dog bed
- It specifies "washable" as a required feature
- It includes "30 inches" to help find beds of the appropriate size

After typing this search query, I'll need to press Enter to execute the search, which will take me to the search results page where I can further refine the results using filters.
</think>

<tool_call>
{"name": "write", "arguments": {"message": "dog bed washable 30 inches"}}
</tool_call>
<|im_end|>

<|im_start|>user
<tool_response>
Succeed: `write` typed "dog bed washable 30 inches" into <input> type=text role=searchbox "Search Amazon".
<observation>
screen size: 1280 x 1000
Current tab: 0
Available tabs:
- Tab 0 (active): https://www.amazon.com/ - Amazon.com. Spend less. Smile more.
</observation>
</tool_response>
<|im_end|>

<|im_start|>assistant
<think>
The previous action successfully typed "dog bed washable 30 inches" into the Amazon search bar, and I can see the search suggestions dropdown has appeared with various related search terms. The search query is ready to be submitted.

My plan is still:
1. Execute the search for "dog bed washable 30 inches"
2. Apply filters for washable beds with length >= 30 inches
3. Review results and find a suitable product

Now that I've entered the search query, the next logical step is to submit the search. I can do this by clicking the search button (the magnifying glass icon) next to the search bar. This will take me to the search results page where I can see all the dog beds that match my criteria.

The search button is visible as an orange button with a magnifying glass icon at the right end of the search bar. Clicking this button will execute the search and show me the results for dog beds that are washable and at least 30 inches long.

After the search results load, I'll need to check if the results match the criteria and potentially apply additional filters to narrow down the options to find the best match for the user's requirements.
</think>

<tool_call>
{"name": "click", "arguments": {"point_2d": [649, 31]}}
</tool_call>
<|im_end|>

<|im_start|>user
<tool_response>
Succeed: `click` on <input> type=submit at (831, 31) executed. Page navigated to https://www.amazon.com/s?k=dog+bed+washable+30+inches&crid=3M6EI7976XXX6&sprefix=dog+bed+...
<observation>
screen size: 1280 x 1000
Current tab: 0
Available tabs:
- Tab 0 (active): https://www.amazon.com/s?k=dog+bed+washable+30+inches&crid=3M6EI7976XXX6&sprefix=dog+bed+... - Amazon.com : dog bed washable 30 inches
</observation>
</tool_response>
<|im_end|>

<|im_start|>assistant
<think>
The search has been successfully executed and I'm now on the Amazon search results page for "dog bed washable 30 inches". The page shows over 10,000 results, which is quite a lot to manually review. I can see several dog beds in the results, and I need to find one that meets the specific criteria: washable and at least 30 inches in length.

Looking at the left sidebar, I can see there's a "Washable" filter option under "Special Features" that I should apply to narrow down the results. This will help ensure that all the results shown are for washable dog beds, which is one of the key requirements.

I also notice that the first product listed (Waterproof Dog Bed Medium Size Dog) mentions "30 inch" in its description, which appears to meet the length requirement. However, to be thorough, I should apply the washable filter first to ensure all results meet that criteria, and then I can look for beds that are at least 30 inches long.

My updated plan:
1. Apply the "Washable" filter by clicking the checkbox in the left sidebar
2. After filtering, review the results to confirm they meet the length requirement (at least 30 inches)
3. Select a suitable product that meets both criteria

The next logical action is to click the "Washable" checkbox under Special Features to filter the results to only show washable dog beds. This will help narrow down the options and make it easier to find products that meet the user's requirements.
</think>

<tool_call>
{"name": "click", "arguments": {"point_2d": [27, 509]}}
</tool_call>
<|im_end|>

<|im_start|>user
<tool_response>
Succeed: `click` on <i> at (35, 509) executed. Page navigated to https://www.amazon.com/s?k=dog+bed+washable+30+inches&rh=p_n_g-1003349998111...
<observation>
screen size: 1280 x 1000
Current tab: 0
Available tabs:
- Tab 0 (active): https://www.amazon.com/s?k=dog+bed+washable+30+inches&rh=p_n_g-1003349998111...
</observation>
</tool_response>
<|im_end|>

<|im_start|>assistant
<think>
The washable filter has been successfully applied, and now I'm seeing 464 results for dog beds that are washable. The page shows several options that appear to meet the criteria, but I need to verify that they have a length of at least 30 inches.

Looking at the first product, the FURTIME Washable Dog Bed, it mentions "Large Sized Dogs" but doesn't explicitly state the dimensions in the visible text. The second product, EHEYCIGA Washable Large Dog Bed, clearly states "35x23 Inch" which means it has a length of 35 inches - meeting the requirement of at least 30 inches.

I should check the product details to confirm the dimensions and washability. The EHEYCIGA bed is a good candidate since it explicitly states 35 inches in length, which exceeds the 30-inch requirement, and it's already filtered as washable.

My next step should be to click on this product to view more details and confirm it meets all the requirements. The EHEYCIGA Washable Large Dog Bed (35x23 Inch) appears to be a strong candidate as it:
1. Is washable (confirmed by the filter)
2. Has a length of 35 inches (exceeding the 30-inch requirement)

I'll click on this product to verify the details and see if it's a suitable match for the user's request.
</think>

<tool_call>
{"name": "click", "arguments": {"point_2d": [496, 514]}}
</tool_call>
<|im_end|>

<|im_start|>user
<tool_response>
Succeed: `click` on <span> "EHEYCIGA Washable Large Dog Be..." at (635, 514) executed. Page navigated to https://www.amazon.com/EHEYCIGA-Washable-Reversible-Sleeping-Mattress/dp/B0FG7XNSYH...
<observation>
screen size: 1280 x 1000
Current tab: 0
Available tabs:
- Tab 0 (active): https://www.amazon.com/EHEYCIGA-Washable-Reversible-Sleeping-Mattress/dp/B0FG7XNSYH/... - Amazon.com : EHEYCIGA Washable Large Dog Bed for C
screenshot:
<|vision_start|><|image_pad|><|vision_end|>
</observation>
</tool_response>
<|im_end|>

<|im_start|>assistant
<think>
\end{toolfamily}

\begin{toolfamily}[listing options={basicstyle=\ttfamily\scriptsize,breaklines=true,breakatwhitespace=true,columns=fullflexible,keepspaces=true,showstringspaces=false}]{Example response from LLM at the last turn}
I've successfully navigated to the product page for the EHEYCIGA Washable Large Dog Bed. This product meets all the requirements specified in the user request:

1. It's washable - confirmed by the product name and description ("Washable Large Dog Bed")
2. It has a length of 35 inches - explicitly stated in the product title as "35x23 Inch" and confirmed in the size specifications at the bottom of the page showing "35" x 23" x 3.5""

The product details confirm it's designed for dogs up to 45 lbs and is labeled as "Amazon's Choice" with good reviews (4.7 stars from 381 reviews). It's also marked as having sustainability features.

I've found a dog bed that satisfies both key requirements:
- Washable (as required)
- At least 30 inches in length (it's 35 inches, which exceeds the minimum requirement)

The task has been successfully completed. I've found a suitable product that meets all the criteria specified in the user request. There's no need for further actions as I've identified a product that matches the requirements.

I should now signal that the task is complete and provide the details of the found product as the answer.
</think>

<tool_call>
{"name": "done", "arguments": {"response": "I found a suitable dog bed on Amazon that meets your requirements: The EHEYCIGA Washable Large Dog Bed (35\" x 23\" x 3.5\") is washable and has a length of 35 inches, which exceeds your minimum requirement of 30 inches. It's priced at \$17.99, has 4.7 stars from 381 reviews, and is labeled as 'Amazon's Choice'. This bed is designed for dogs up to 45 lbs and is currently in stock."}}
</tool_call>
<|im_end|>
\end{toolfamily}

\section{Orchard-GUI Task Filtering Pipeline}
\label{app:gui:task_filtering}

We draw task instances from the task set organized by WebGym~\citep{bai2026webgym}, which consists of \textbf{292{,}092} raw task instances in total. To produce a clean, evaluation-safe, and diverse pool of training prompts, we apply a five-stage filtering pipeline:

\begin{enumerate}
    \item \emph{Remove common evaluation benchmarks.} We strip out splits that overlap with our held-out benchmarks (e.g., Online-Mind2Web~\citep{deng2023mind2web} and DeepShop~\citep{lyu2025deepshop}) to prevent train/test contamination, retaining only the two complementary PAE-WebVoyager~\citep{zhou2025pae} and InSTA-v3~\citep{Trabucco2025InSTA} splits (-13{,}840, $4.7\%$ $\rightarrow$ \textbf{278{,}252}). The former consists of automatically proposed web-navigation tasks generated by a context-aware task proposer, grounded in the websites covered by the WebVoyager~\citep{he2024webvoyager} benchmark, while the later contains tasks automatically synthesized by an LLM over a large and diverse set of websites. Each task is grounded in a specific domain and phrased as a realistic user goal (e.g., finding information, retrieving attributes, or completing simple workflows), with an emphasis on feasibility and safety.
    \item \emph{Keep parent tasks only.} WebGym additionally provides child tasks decomposed from each parent intent. Since child tasks share substantial structure with their parents, we retain only the parents to avoid intra-family redundancy (-23{,}437, $8.4\%$ $\rightarrow$ \textbf{254{,}815}). 
    \item \emph{Exclude WebVoyager tasks.} We further drop any task whose intent appears in the original WebVoyager benchmark, eliminating residual contamination at the prompt level (-411, $0.2\%$ $\rightarrow$ \textbf{254{,}404}).
    \item \emph{Restrict to popular websites.} Long-tail websites are noisier (more captchas, anti-bot blocks, broken pages) and less representative of realistic browsing. We keep only tasks whose target site falls within the SimilarWeb Top-100 list and the MOZ Top 500 Most Popular Websites, and where the same site has at least two tasks, ensuring sufficient per-site coverage for the downstream agent (-114{,}349, $44.9\%$ $\rightarrow$ \textbf{140{,}055}).
    \item \emph{Semantic deduplication.} The remaining pool is dominated by near-duplicate intents (e.g., paraphrases of the same shopping or search query across thousands of products). We embed each task intent with \texttt{Qwen/Qwen3-Embedding-8B} and greedily remove tasks whose cosine similarity to a previously kept task exceeds $0.99$ (-124{,}454, $88.9\%$ $\rightarrow$ \textbf{15{,}601}).
\end{enumerate}

The final filtered pool of \textbf{15{,}601} unique task intents serves as the seed set from which we sample teacher trajectories for SFT and RL prompts.
\section{Example Claw Agent Trajectory}
\label{app:claw:trajectory}

\begin{toolfamily}[listing options={basicstyle=\ttfamily\scriptsize,breaklines=true,breakatwhitespace=true,columns=fullflexible,keepspaces=true,showstringspaces=false}]{Example Model Input in the ZeroClaw Harness}
<|im_start|>system
You are ZeroClaw, an AI assistant with tool access. Use tools to accomplish tasks.
Be concise and helpful. Execute tools directly when needed without excessive explanation.

# Tools

You may call one or more functions to assist with the user query.

You are provided with function signatures within <tools></tools> XML tags:
<tools>
{"type": "function", "function": {"name": "shell", "description": "Execute a shell command and return the output.\n\nArgs:\n    command: The shell command to execute\n\nReturns:\n    The command output (stdout and stderr combined)", "parameters": {"properties": {"command": {"type": "string"}}, "required": ["command"], "type": "object"}}}
[... truncated ...]

For each function call, return a json object with function name and arguments within <tool_call></tool_call> XML tags:
<tool_call>
{"name": <function-name>, "arguments": <args-json-object>}
</tool_call><|im_end|>
<|im_start|>user
Diana from operations emailed about the Q2 inventory rotation. Read her email and Tom's follow-up for full instructions. Check the inventory, identify critical and expiring items, email the relevant suppliers (look up their contacts first), set up weekly scheduler follow-ups for critical items, and write the rotation plan summary to /home/user/rotation_plan.txt.
<|im_end|>
<|im_start|>assistant
<think>
The user wants me to:
1. Read Diana's email about Q2 inventory rotation
2. Read Tom's follow-up email
3. Check inventory
4. Identify critical and expiring items
5. Email relevant suppliers (look up contacts first)
6. Set up weekly scheduler follow-ups for critical items
7. Write rotation plan summary to /home/user/rotation_plan.txt

Let me start by listing recent emails to find Diana's email and Tom's follow-up.
</think>

<tool_call>
{"name": "gmail_list_messages", "arguments": {"days_back": 14, "max_results": 20}}
</tool_call><|im_end|>
<|im_start|>user
<tool_response>
{"messages":[{"message_id":"msg_401","from":"d.kowalski@ourcompany.com","subject":"Q2 Inventory Rotation - Need Your Help","date":"2026-04-10T03:53:24Z",...
[... truncated ...]
</tool_response><|im_end|>
<|im_start|>assistant
<think>
I found Diana's email (msg_401) and Tom's follow-up (msg_402). Let me get the full content of both.
</think>

<tool_call>
{"name": "gmail_get_message", "arguments": {"message_id": "msg_401"}}
</tool_call>
<tool_call>
{"name": "gmail_get_message", "arguments": {"message_id": "msg_402"}}
</tool_call><|im_end|>
<|im_start|>user
<tool_response>
{"message_id":"msg_401","from":"d.kowalski@ourcompany.com","to":"assistant@ourcompany.com","subject":"Q2 Inventory Rotation - Need Your Help","body":"Hi,\n\nIt's time for our quarterly inventory rotation review. I need you to go through our current inventory and identify all items that are CRITICAL (stock below minimum threshold) or EXPIRING SOON ...
[... truncated ...]
</tool_response>
<tool_response>
{"message_id":"msg_402","from":"t.nguyen@ourcompany.com","to":"assistant@ourcompany.com","subject":"Re: Q2 Inventory Rotation - Budget Note","body":"Hey,\n\nJust a heads up for the inventory rotation - our Q2 procurement budget is tight. For any item where the restock cost would exceed $1000 (quantity needed to reach minimum x unit cost), flag it as HIGH PRIORITY in the rotation plan so I can fast-track the PO approval.\n\nAlso, once you've emailed the suppliers, send me a summary email at t.nguyen@ourcompany.com listing all restock requests you've sent out.\n\nTom Nguyen\nProcurement Lead","date":"2026-04-10T05:38:24Z","is_read":false,"labels":["inbox","operations"]}
</tool_response><|im_end|>
<|im_start|>assistant
<think>
\end{toolfamily}

\begin{toolfamily}[listing options={basicstyle=\ttfamily\scriptsize,breaklines=true,breakatwhitespace=true,columns=fullflexible,keepspaces=true,showstringspaces=false}]{Example Response from LLM}
Now I understand the task:
1. Check inventory for items that are:
   - CRITICAL: stock below minimum threshold
   - EXPIRING SOON: expiry date within 45 days from April 10, 2026 (so before May 25, 2026)
2. Email suppliers for restocking quotes - group by supplier
3. Set up weekly scheduler follow-ups for critical items
4. Write rotation plan summary to /home/user/rotation_plan.txt
5. Also need to flag items >$1000 as HIGH PRIORITY for Tom

Let me get the inventory list first.
</think>

<tool_call>
{"name": "inventory_list_products", "arguments": {}}
</tool_call>
<|im_end|>
\end{toolfamily}
\end{document}